\documentclass[table]{arxiv_v2/ai2style/ai2}

\usepackage{microtype}
\usepackage{hyperref}
\usepackage{url}
\usepackage{booktabs, tabularx} % tabularx for ian's decon table
\usepackage{graphicx}
\usepackage{lineno}
\usepackage{enumitem}
\usepackage{listings} % for code listing
\usepackage{svg}

\definecolor{ darkblue}{rgb}{0, 0, 0.5}
\hypersetup{colorlinks=true, citecolor=darkblue, linkcolor=darkblue, urlcolor=darkblue}

\usepackage{amssymb}
\usepackage{multirow}
\usepackage{bigdelim}
\usepackage{todonotes}
\usepackage{longtable}
\usepackage{tabularray}
\usepackage{wrapfig}
\usepackage[most]{tcolorbox}
\usepackage{url}
\usepackage{xspace}
\usepackage{svg}
\usepackage[absolute]{textpos} % Add this to your preamble
\usepackage[utf8]{inputenc} % allow utf-8 input
\usepackage{url}            % simple URL typesetting
\usepackage{booktabs}       % professional-quality tables
\usepackage{amsfonts}       % blackboard math symbols
\usepackage{nicefrac}       % compact symbols for 1/2, etc.
\usepackage{microtype}      % microtypography
\usepackage[table,xcdraw]{xcolor}         % colors. Pass options into class too, to avoid option conflicts at arxiv time % Ian is using xcdraw in decon perf drop table
\usepackage{amsmath}
\usepackage[most]{tcolorbox}
\usepackage{csquotes}
\usepackage{tablefootnote}
\usepackage{ragged2e}
\usepackage{graphicx}
\usepackage{arydshln}
\usepackage{wrapfig}
\usepackage{enumitem}

\usepackage{bm,placeins,hyphenat}
\definecolor{ourcolor}{RGB}{230,240,245}
\definecolor{baselinecolor}{gray}{.9}

\usepackage{multirow}
\usepackage{graphicx}
\usepackage{listings}
\lstdefinelanguage{Prompt}{
  morestring=[b]",
}
\definecolor{codebg}{RGB}{245,245,245}
\usepackage{arxiv_v2/algorithm}
\usepackage{arxiv_v2/algorithmic}
\definecolor{codekw}{rgb}{0.00,0.00,0.60}
\definecolor{codestr}{rgb}{0.58,0.00,0.13}
\definecolor{codecmt}{rgb}{0.30,0.50,0.30}
\definecolor{codenum}{rgb}{0.55,0.55,0.55}
\lstdefinestyle{pystyle}{
  language=Python,
  backgroundcolor=\color{codebg},
  basicstyle=\ttfamily\footnotesize,
  keywordstyle=\color{codekw}\bfseries,
  commentstyle=\color{codecmt}\itshape,
  stringstyle=\color{codestr},
  numberstyle=\tiny\color{codenum},
  showstringspaces=false,
  breaklines=true,
  breakatwhitespace=true,
  numbers=left,
  numbersep=6pt,
  frame=none,
  framexleftmargin=4pt,
  xleftmargin=12pt,
  columns=fullflexible,
  keepspaces=true,
  belowskip=4pt, aboveskip=6pt,
  emph={self,True,False,None},
  emphstyle=\color{codekw},
}

\usepackage{multirow}
\usepackage{xspace}
\usepackage{adjustbox}
\usepackage{pifont}
\usepackage{caption}
\usepackage{makecell}
\usepackage{subcaption}
\usepackage{bold-extra}
\usepackage{url}

\usepackage{pgf-pie}

\usepackage{hyperref}
\definecolor{linkcolor}{RGB}{0, 0, 128}
\hypersetup{
     colorlinks   = true,
     citecolor    = linkcolor,
     linkcolor    = linkcolor,
     urlcolor     = linkcolor,
}
\usepackage{pifont}% http://ctan.org/pkg/pifont
\usepackage{listings}
\setlist[itemize]{leftmargin=*,itemsep=0em,parsep=0.3em,topsep=0.3em}

\ifdefined\DeclareUnicodeCharacter
\DeclareUnicodeCharacter{2212}{\ensuremath{-}}
\fi

\definecolor{maroon}{HTML}{F26035}
\definecolor{yellow}{HTML}{FDBC42}
\definecolor{lavender}{HTML}{734f96}
\definecolor{darkergrey}{HTML}{444444}
\definecolor{midgrey}{HTML}{e6eded}
\definecolor{ai2pink}{HTML}{f0529c}%{105257}
\definecolor{ai2midpink}{HTML}{fad3e5}
\definecolor{ai2lightpink}{HTML}{fbecf3}
\definecolor{ai2midwhite}{HTML}{f2e5d9}
\definecolor{ai2offwhite}{HTML}{fbf4ee}
\definecolor{ai2green}{HTML}{0fcb8c}
\definecolor{ai2lightgreen}{HTML}{e7f9f3}
\definecolor{ai2darkgreen}{HTML}{105257}
\definecolor{ai2purple}{HTML}{B932EB}
\definecolor{ai2lightpurple}{HTML}{f7e8fc}
\definecolor{neutralEight}{HTML}{343434}
\definecolor{neutralFive}{HTML}{838383}
\definecolor{neutralThree}{HTML}{bebebe}
\definecolor{neutralOne}{HTML}{dedede}
\definecolor{lightgrey}{HTML}{fafcfc}
\definecolor{plum}{rgb}{0.56,0.27,0.52}

\usepackage{tikz}

\definecolor{maroon}{HTML}{F26035}
\definecolor{yellow}{HTML}{FDBC42}
\definecolor{darkred}{RGB}{156, 39, 33}
\definecolor{darkblue}{RGB}{31, 90, 153}
\definecolor{forestgreen}{rgb}{0.13, 0.55, 0.13}
\definecolor{brickred}{rgb}{0.8, 0.25, 0.33}
\definecolor{olmoDarkBlue}{HTML}{012e59}
\definecolor{olmoBlue}{HTML}{265ed4}
\definecolor{olmoLightBlue}{HTML}{012e59}
\definecolor{olmoTeal}{HTML}{00d5ff}
\definecolor{olmoYellow}{HTML}{ffbb00}
\definecolor{olmoOrange}{HTML}{ff9100}

\newcommand{\app}{\raise.17ex\hbox{$\scriptstyle\sim$}}
\renewcommand{\paragraph}[1]{\vspace{0.5mm}\noindent\textbf{#1}}

\definecolor{molmocolor}{RGB}{240, 82, 156}
\definecolor{tablegray}{RGB}{223, 242, 252}
\definecolor{tablegreen}{RGB}{15, 203, 150}
\definecolor{tableyellow}{RGB}{250, 242, 233}
\definecolor{tableblue}{RGB}{240, 82, 156}
\definecolor{darkpink}{RGB}{139, 14, 98}

\usepackage{setspace}

\newcolumntype{L}[1]{>{\raggedright\let\newline\\\arraybackslash\hspace{0pt}}m{#1}}
\newcolumntype{C}[1]{>{\centering\let\newline\\\arraybackslash\hspace{0pt}}m{#1}}
\newcolumntype{R}[1]{>{\raggedleft\let\newline\\\arraybackslash\hspace{0pt}}m{#1}}
\newcolumntype{P}[1]{>{\centering\let\newline\\\arraybackslash\columncolor{ai2lightpink}}m{#1}}
\newcommand{\github}{\raisebox{-1.5pt}{\includegraphics[height=1.05em]{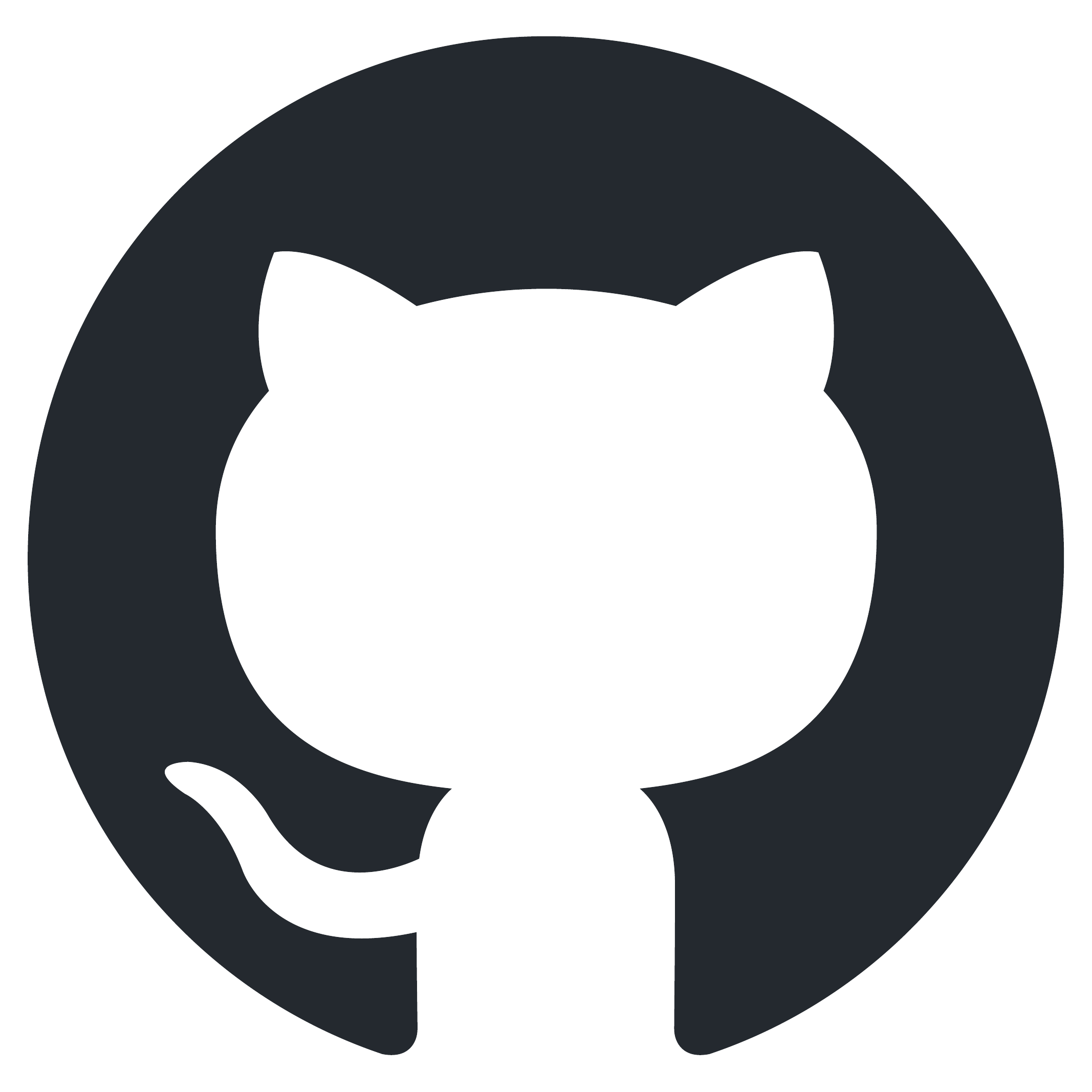}}\xspace}

\makeatletter
\DeclareRobustCommand\onedot{\futurelet\@let@token\@onedot}
\def\@onedot{\ifx\@let@token.\else.\null\fi\xspace}

\makeatother

\definecolor{titlegA}{HTML}{3FA838}  % deep green   (top-left)
\definecolor{titlegB}{HTML}{6FB833}  % yellow-green
\definecolor{titlegC}{HTML}{9FC72E}  % chartreuse
\definecolor{titlegD}{HTML}{CFD629}  % lime-yellow
\definecolor{titlegE}{HTML}{F0CB1F}  % warm yellow  (bottom-right)

\title{{\fontsize{16pt}{12pt}\selectfont%
    \mbox{\raisebox{-0.28\height}{\includegraphics[height=1.6em]{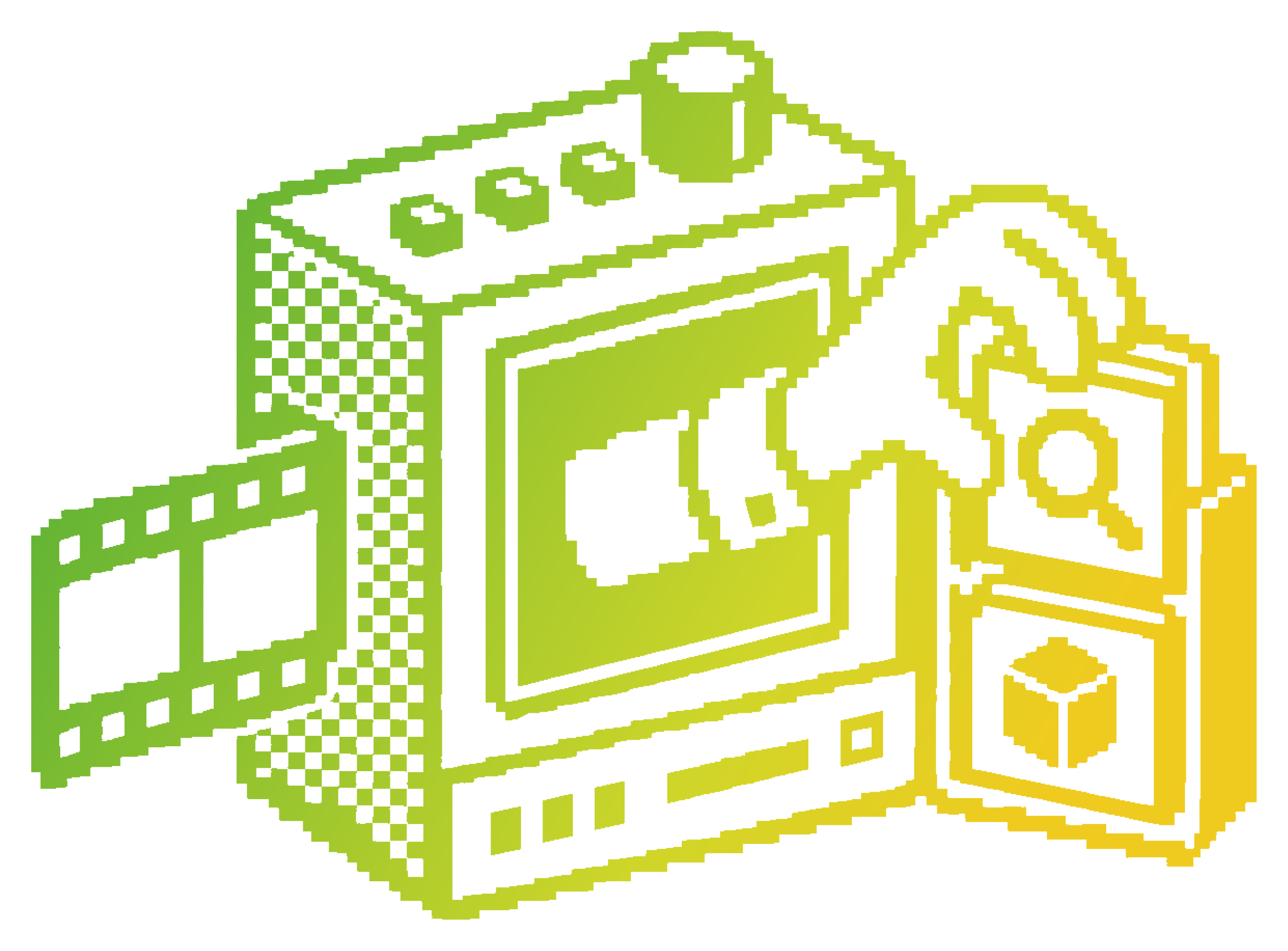}}\,%
    \textcolor{titlegA}{VideoGen-Agent:}\,\textcolor{titlegB}{Reinforcing}\,%
    \textcolor{titlegC}{Video}\,\textcolor{titlegD}{Generation}\,%
    \textcolor{titlegE}{Agents}}%
  }}

\authorOne[1]{Binxu Li}
\authorOne[1]{Haoyi Duan}
\authorOne[2]{Yaohui Zhang}
\authorOne[2]{Yuhui Zhang}
\authorOne[1]{Hui Yuan}
\authorTwo[3]{Zihao Lin}
\authorTwo[4]{Kaituo Feng}
\authorTwo[1]{Suozhi Huang}
\authorTwo[5]{Xiangyi Li}
\authorTwo[6]{Yu Li}
\authorThree[]{Chunyuan Li}
\authorThree[1]{Mengdi Wang}
\authorThree[1]{Shilong Liu}

\affiliation[1]{Princeton University}
\affiliation[2]{Stanford University}
\affiliation[3]{UC Davis}
\affiliation[4]{MMLab, CUHK}
\affiliation[5]{BenchFlow}
\affiliation[6]{GWU}
\newcommand{\institutionlogos}{%
  \makebox[\textwidth]{%
    \raisebox{-0.15\height}{\includegraphics[height=0.9cm]{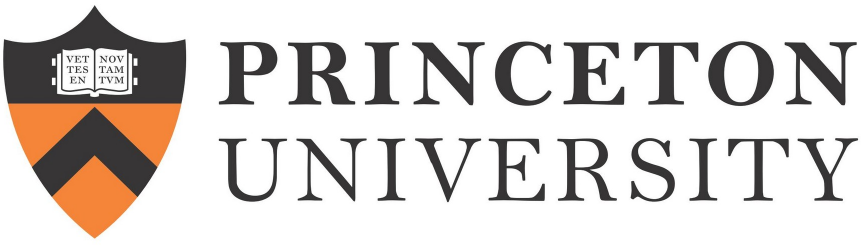}}%
    \hfill
    \raisebox{-0.15\height}{\includegraphics[height=0.9cm]{arxiv_v2/logos/stanford.pdf}}%
  }%
}
\newlength{\headrulegap}
\renewcommand{\headrule}{%
  \vskip\headrulegap%
  \hrule height\headrulewidth width\headwidth%
  \vskip-\headrulewidth%
}
\fancypagestyle{firstpage}{%
  \fancyhf{}%
  \fancyhead[C]{\institutionlogos}%
  \renewcommand{\headrulewidth}{1.4pt}%
  \renewcommand{\footrulewidth}{0pt}%
}
\fancypagestyle{withtitle}{%
  \fancyhf{}%
  \fancyhead[C]{\small\sffamily VideoGen-Agent: Reinforcing Video Generation Agents}%
  \fancyfoot[C]{\small\thepage}%
  \renewcommand{\headrulewidth}{1.2pt}%
  \renewcommand{\footrulewidth}{0pt}%
}
\newcommand{\method}{VideoGen-Agent}
\newcommand{\backbone}{Qwen3-VL-8B-Instruct}
\newcommand{\ourbench}{VABench}

\newcommand{\taskoneab}{PK}
\newcommand{\tasktwoab}{SI}
\newcommand{\taskthreeab}{MI}
\newcommand{\taskfourab}{PS}
\newcommand{\taskfiveab}{CS}
\newcommand{\tasksixab}{MS}

\newcommand{\taskone}{Procedural Knowledge}
\newcommand{\tasktwo}{Single-Entity Identity}
\newcommand{\taskthree}{Multi-Entity Identity}
\newcommand{\taskfour}{Physics Simulation}
\newcommand{\taskfive}{Compositional Scene}
\newcommand{\tasksix}{Multi-Shot}
\newcommand{\taskcombo}{Procedural Identity}
\newcommand{\taskcomboab}{PI}

\definecolor{baselinecolor}{gray}{.9}

\usepackage[utf8]{inputenc}
\usepackage[most]{tcolorbox}
\tcbuselibrary{listings,breakable}

\lstdefinestyle{promptstyle}{
  basicstyle=\ttfamily\scriptsize,
  breaklines=true,
  breakatwhitespace=false,
  columns=fullflexible,
  keepspaces=true,
  showstringspaces=false,
  xleftmargin=0pt,
  xrightmargin=0pt,
  breakindent=0pt,
  breakautoindent=false,
  resetmargins=true,
  postbreak=\mbox{},
  linewidth=\linewidth,
  literate=
    {∈}{{$\in$}}1
    {≥}{{$\ge$}}1
    {≤}{{$\le$}}1
    {×}{{$\times$}}1
    {→}{{$\to$}}1
    {—}{{-}}1
    {–}{{-}}1
    {“}{{"}}1
    {”}{{"}}1
    {‘}{{'}}1
    {’}{{'}}1
    {²}{{$^2$}}1
}

\newtcblisting{promptbox}[1][]{
  listing only,
  breakable,
  colback=gray!5,
  colframe=gray!60,
  boxrule=0.5pt,
  arc=2pt,
  left=0pt,
  right=0pt,
  top=6pt,
  bottom=6pt,
  title={#1},
  listing options={style=promptstyle}
}

\definecolor{molmocolor}{RGB}{240, 82, 156}
\definecolor{tablegray}{RGB}{223, 242, 252}
\definecolor{tablegreen}{RGB}{15, 203, 150}
\definecolor{tableyellow}{RGB}{250, 242, 233}
\definecolor{tableblue}{RGB}{240, 82, 156}
\definecolor{darkpink}{RGB}{139, 14, 98}
\definecolor{baselinecolor}{gray}{.9}

\newlength{\teaserurlgap}
\newlength{\teaserbottomgap}
\renewcommand{\teaserfigure}{%
  \begingroup
    \centering
    \includegraphics[width=1\textwidth]{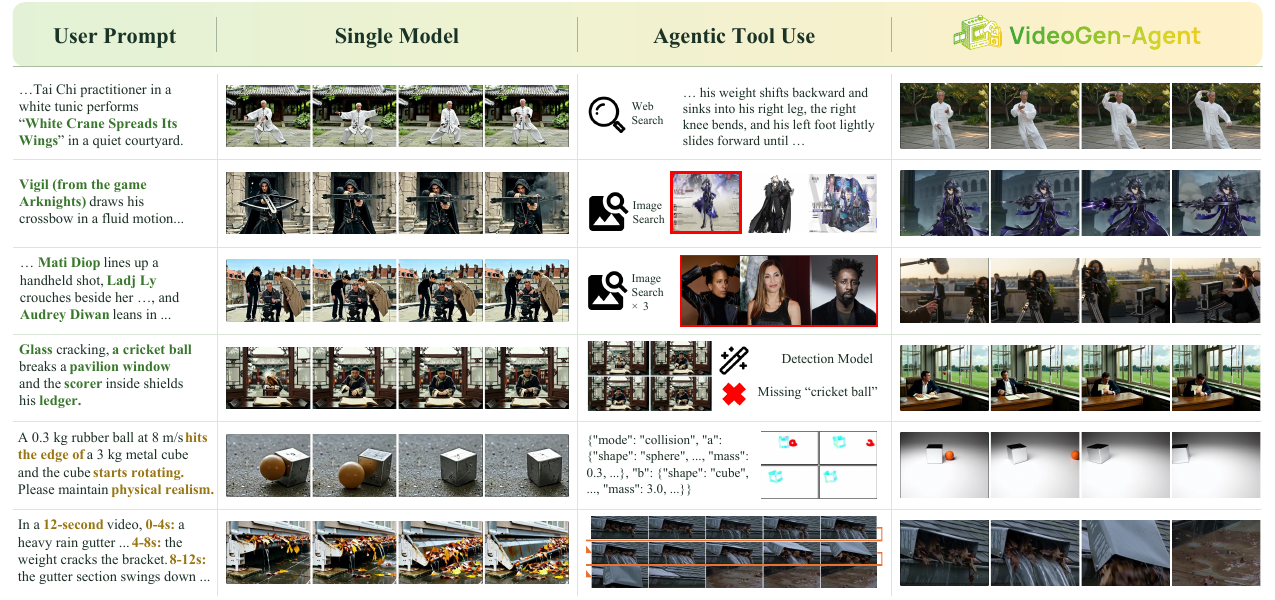}\par
  \endgroup
  \vspace*{\teaserbottomgap}%
  \begin{center}
  \begin{tcolorbox}[enhanced, frame hidden,
                    width=0.9\textwidth,
                    left=0.5cm, right=0.5cm, top=0.5cm, bottom=0.5cm,
                    arc=3.5pt, colback=ai2background, before skip=0pt]
    {\Large\sffamily\bfseries Abstract\par}
    \vspace{0.3cm}
    Recent advances in video generative models have enabled high-fidelity, temporally coherent video generation. However, these models often struggle to satisfy prompts requiring specialized knowledge, specific identities, physical consistency, or ordered events. In this paper, we present \textbf{\method{}}, a multimodal agent trained through \textbf{multitask agentic reinforcement learning} to use external tools for video generation. The agent coordinates augmentation, generation, and verification tools through multi-turn interactions, using the prompt and intermediate observations to guide its decisions. We train a shared policy on a category-balanced dataset spanning six tasks. Supervised fine-tuning on teacher-generated trajectories establishes tool-use behavior, which is then refined through reinforcement learning. A category-aware hybrid reward evaluates tool-call validity, task-appropriate tool use, and generated video quality. We further introduce \textbf{\ourbench{}}, a held-out benchmark of 600 prompts designed to evaluate video generation agents across six capability categories: procedural knowledge, single- and multi-entity identity preservation, physical consistency, scene composition, and multi-shot temporal structure. On \ourbench{}, \method{} improves over its base video generator by $19.1$ points, from $56.5$ to $75.6$. Upgrading the generation tools further raises the score to $86.1$ without additional agent training. Human raters prefer the upgraded configuration over the strongest standalone baseline in $84.3\%$ of comparisons. These results support learning tool use across video-generation tasks and show that the trained agent can benefit from subsequent advances in generation tools.

\vspace{2pt}
\begingroup
    \raggedright
    \noindent
    {\small \github Project homepage:\;
      \href{https://andyca111.github.io/VideoGen_Agent/}%
           {\nolinkurl{andyca111.github.io/VideoGen_Agent}}}\par
\endgroup

  \end{tcolorbox}
  \end{center}
}

\lstdefinestyle{datapromptstyle}{
 style=promptstyle,
 literate=
 {×}{{$\times$}}1 {â}{{\^a}}1 {é}{{\'e}}1 {í}{{\'i}}1
 {–}{{-}}1 {—}{{---}}3 {’}{{'}}1 {“}{{"}}1 {”}{{"}}1
 {…}{{...}}3 {→}{{$\rightarrow$}}2 {↔}{{$\leftrightarrow$}}2
 {★}{{*}}1 {⛔}{{[EXCLUDE]}}9 {✅}{{[PASS]}}6 {❌}{{[REJECT]}}8
 {王者荣耀}{{Wangzhe Rongyao}}{14}
 {📂}{{[CATEGORIES]}}{12} {📐}{{[FORMAT]}}8 {🔑}{{[KEY]}}5
 {🚨}{{[IMPORTANT]}}{11}
}
\newtcblisting{datapromptbox}[1][]{
 listing only, breakable,
 colback=gray!3, colframe=gray!60, boxrule=0.4pt, arc=1pt,
 left=3pt, right=3pt, top=4pt, bottom=4pt,
 title={#1}, listing options={style=datapromptstyle}
}

\begin{document}

\maketitle
\thispagestyle{firstpage}  % page 1 gets the logos header (pages 2+ inherit \pagestyle{withtitle})

\section{Introduction}
\label{sec:intro}

Video generation has advanced rapidly, bringing synthetic videos closer to the visual quality of recorded footage~\citep{brooks2024sora, seedance2026seedance, gao2025seedance}. Diffusion-based models can now generate photorealistic, temporally coherent videos from free-form text prompts~\citep{ho2022video,blattmann2023stable,wan2025wan}. They can depict complex camera movements and varied lighting conditions across a range of visual styles~\citep{wan2025wan}. However, this visual quality does not ensure that a generated video accurately captures the content specified in a prompt~\citep{t2vcompbench}. 

Current video generators rely on knowledge acquired during pretraining, limiting their ability to depict entities or events that emerged afterward. Even for familiar subjects, they can misrepresent a named technique or fail to preserve a person's visual identity throughout a clip~\citep{yuan2025identity}. Generated motion may also violate basic physical laws, producing implausible behavior under gravity or during collisions~\citep{guo2025t2vphysbench}. Prompts that specify an ordered sequence of actions pose a further challenge, as models may omit phases or depict them out of order~\citep{11339268}. These limitations affect practical uses ranging from product demonstrations and educational content to sports analysis and film pre-visualization~\citep{Long_2026_CVPR}. These challenges motivate the use of external knowledge and tools to guide video generation.

Agentic video generation supplements video models with external tools and feedback, but existing approaches often target specific objectives~\citep{wu2025automated,Long_2026_CVPR,feng2026newton}. Learning tool use across heterogeneous tasks requires an agent to select appropriate support, such as visual references for identity preservation or simulation for physical motion. It must also use intermediate observations to guide subsequent actions. This motivates multitask agentic reinforcement learning to optimize these decisions within a single agent using task-dependent feedback.

In this paper, we propose \textbf{\method{}}, a multimodal agent trained through \textbf{multitask agentic reinforcement learning} to use external tools for video generation. It addresses six tasks: \taskone{}, \tasktwo{}, \taskthree{}, \taskfour{}, \taskfive{}, and \tasksix{}. Across these tasks, the agent coordinates \emph{augmentation}, \emph{generation}, and \emph{verification} tools based on the prompt and intermediate observations. Retrieval and simulation guide generation, while object detection~\citep{dino} and depth estimation~\citep{dpany} provide feedback for refinement. We train the agent across all six categories through supervised fine-tuning on teacher-distilled trajectories, followed by multitask agentic reinforcement learning. A hybrid reward evaluates tool use and generated video quality, while task advantage normalization balances learning signals across categories. A common tool interface also allows compatible tools to be upgraded without redesigning the agent's workflows.

\begin{figure}[!t]
    \centering
    \includegraphics[width=\linewidth]{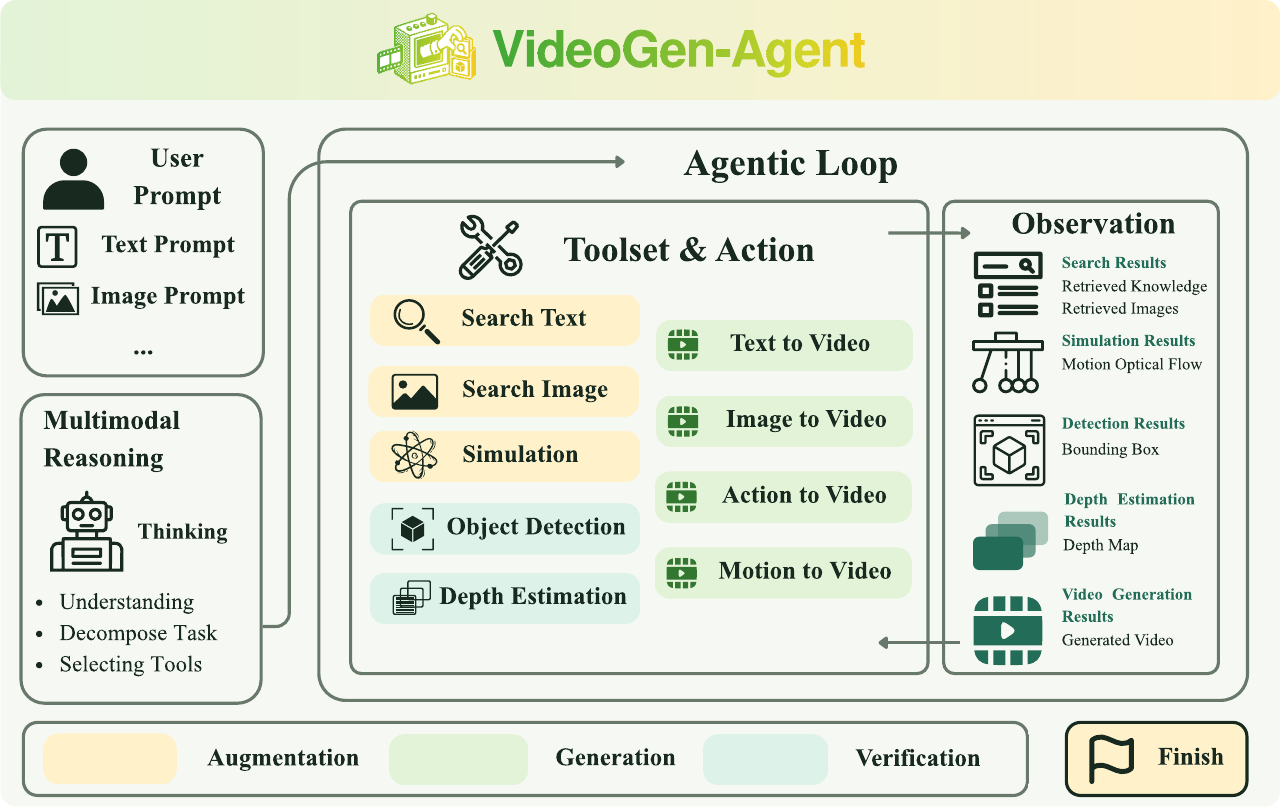}
    \caption{\textbf{Overview of \method{}.} Given a user prompt, the multimodal agent reasons about the generation requirements and selects tools within a multi-turn reasoning--action--observation loop. Tool outputs inform subsequent decisions, allowing the agent to gather information, generate video candidates, and refine them. The toolbox contains three functional groups: \emph{augmentation}, \emph{generation}, and \emph{verification}. Augmentation tools retrieve textual knowledge and visual references or simulate motion to guide generation. Generation tools support text-to-video, image-to-video, and motion-to-video generation; Appendix~\ref{app:robotics} explores an additional action-to-video tool chain. Here, image-to-video includes both single-image-to-video (I2V) and multiple-reference images-to-video (R2V). Verification tools use object detection and depth estimation to assess subject presence and spatial structure, providing feedback for refinement. The agent returns the generated video when it finishes the interaction.}
    \vspace{-15pt}
    \label{fig:main}
\end{figure}

To evaluate \textbf{\method{}}, we introduce \textbf{\ourbench{}}, a held-out benchmark comprising 600 complex prompts for video generation agents, spanning the six capability categories described above. Each prompt emphasizes its target capability, enabling targeted evaluation of distinct challenges in video generation. We score generated videos using category-specific VLM rubrics and assess the judge's agreement with human preferences.

On \ourbench{}, \method{} improves its base T2V generator by $19.1$ points, from $56.5$ to $75.6$. Upgrading the generation tools further raises the score to $86.1$ without additional agent training, achieving the highest scores across all six categories. Human raters prefer the upgraded configuration over the strongest standalone baseline in $84.3\%$ of comparisons. Ablations confirm the contributions of both training stages and reward components, supporting the benefits of learned tool use alongside stronger generation tools.

Our contributions are as follows:
\begin{itemize}
    \item We introduce \textbf{\method{}}, a multimodal agent that learns to use external tools for video generation through \textbf{multitask agentic reinforcement learning}. SFT followed by multitask RL trains a shared policy to coordinate augmentation, generation, and verification tools.

    \item We construct a tool-use trajectory dataset spanning six task categories and introduce \textbf{\ourbench{}} for held-out evaluation. The benchmark covers six capability categories that current single video generation models struggle with, including \taskone{}, \tasktwo{}, \taskthree{}, \taskfour{}, \taskfive{} and \tasksix{}.

    \item \method{} improves the base T2V generator by approximately $19$ points on \ourbench{}. Human evaluation supports the upgraded configuration, while ablations demonstrate the contributions of SFT and RL with the generation tools held fixed.
\end{itemize}

\section{Method}
\vspace{-3pt}
\label{sec:method}

In this section, we present the multitask agentic workflow and training procedure of \method{}.
We first describe how the agent interacts with tools across six video-generation tasks, then detail supervised fine-tuning and reinforcement learning.

\subsection{Multitask Agentic Workflow}

Given a text prompt $x$, \method{} uses a shared multimodal policy to generate a video through multi-turn tool interactions. As illustrated in Figure~\ref{fig:main}, each interaction follows a reasoning--action--observation loop. At step $t$, the policy $\pi_\theta$ reasons over the history $H_t$, including the prompt, previous actions, and tool observations. It then selects a tool and specifies its arguments, incorporating the returned observation into the history for the next decision. This loop supports different generation workflows, allowing the agent to gather information, generate candidates, and verify or refine them as needed.
The complete rollout procedure is provided in Algorithm~\ref{alg:agent} in Appendix~\ref{app:agent_rollout}.

\subsubsection{Tasks and Tools}
\label{sec:task_tools}

Table~\ref{tab:workflow} lists the tools most relevant to each of the six task categories together with a default workflow. These workflows are reference patterns rather than fixed pipelines: the shared policy selects tools under the system prompt's tool-specific constraints and decides which tools to call, in what order, and whether to verify or regenerate, and incorporates returned information according to each prompt. \textbf{(a) \emph{\taskone{}} (\taskoneab{})} prompts describe specialized actions or processes whose accurate depiction requires detailed procedural knowledge. Text retrieval supplies relevant steps and motion details, which the agent incorporates into the generation prompt. \textbf{(b) \emph{\tasktwo{}} (\tasktwoab{})} and \textbf{(c) \emph{\taskthree{}} (\taskthreeab{})} require preserving the visual identity of one or more named subjects throughout a video. Subjects may include public landmarks, branded objects, game characters, celebrities, and so on. The agent searches visual references and passes selected images to an image-conditioned or multiple-reference-conditioned generator. \textbf{(d) \emph{\taskfour{}} (\taskfourab{})}
prompts specify physical motion under given initial conditions, restricted to collisions and falling/dropping in both training and VABench. The agent invokes simulation tools to obtain motion conditions and passes them to a motion-conditioned generator. \textbf{(e) \emph{\taskfive{}} (\taskfiveab{})}
tests multi-subject composition, inter-object interactions, and spatial relations. Prompts use visually canonical subjects to emphasize scene structure rather than identity. The agent generates a candidate, checks it with verification tools, and uses the feedback to guide regeneration when necessary. \textbf{(f) \emph{\tasksix{}} (\tasksixab{})}
tests whether a video depicts all requested phases in the correct temporal order. These prompts also use visually canonical subjects, allowing evaluation to focus on temporal structure. The agent generates the phases sequentially, using the final frame of each segment to condition the next and maintain visual continuity.

All tools are accessed through a common interface, allowing compatible backend implementations to be used within the same workflows. Table~\ref{tab:tool_impl} lists the concrete implementations in both toolsets; Appendix~\ref{app:tool_impl} provides additional tool details.

\begin{figure}[!t]
    \centering
    \includegraphics[width=\linewidth]{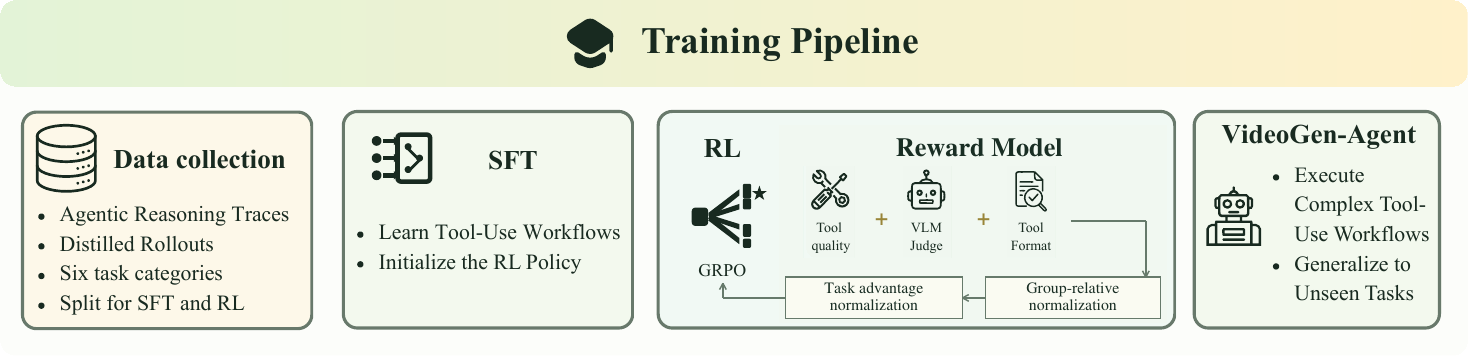}
    \vspace{-10pt}
    \caption{\textbf{Overview of the training pipeline for \method{}.} We construct prompts across six task categories and collect teacher-generated tool-use trajectories for supervised fine-tuning (SFT). Starting from the SFT checkpoint, the agent generates new rollouts on a separate prompt set for multitask reinforcement learning with GRPO. A hybrid reward evaluates tool-call validity, task-appropriate tool use, and generated video quality. Task advantage normalization follows group-relative normalization to balance learning signals across categories.}
    \vspace{-15pt}
    \label{fig:training}
\end{figure}

\subsection{Data Construction}
\label{sec:data}

High-quality training data is essential for an agent that must learn to compose specialized tools into a generation pipeline. However, no public dataset directly aligns prompts with the category labels, agentic trajectories, and reference assets needed for tool-grounded video generation. To address this, we carefully curate a challenging training dataset.

\paragraph{Prompt Construction}
We use \textbf{Claude Opus 4.7} to generate training prompts in batches of 100 for each of the six task categories. We supplement these prompts with examples drawn from public datasets, including grounded image-generation examples from Gen-Searcher~\citep{feng2026gensearcher} adapted into video prompts. Following the task definitions in Section~\ref{sec:task_tools}, we simplify non-target aspects of each prompt so that the intended capability remains the primary challenge. For \emph{\taskone{}} and \emph{\taskfour{}}, we use simple subjects and backgrounds to focus on procedural accuracy and physical dynamics, respectively. For \emph{\tasktwo{}} and \emph{\taskthree{}}, we use simple actions or explicitly describe signature skills, keeping the challenge focused on identity preservation. For \emph{\taskfive{}} and \emph{\tasksix{}}, both subjects and individual actions are simple, concentrating the challenge on spatial composition and temporal ordering, respectively. Actual \emph{\taskfour{}} prompts specify quantitative collisions or falling/dropping. Each \emph{\tasksix{}} prompt specifies two or three equal-duration shots, each at most five seconds, to support sequential generation.  Construction, quality filtering, and the training/evaluation split are detailed in Appendix~\ref{app:data_construction}; one initial data-generation prompt per task is provided in Appendix~\ref{app:data_system_prompts}.

\paragraph{Agentic Trajectory Generation}
For each constructed prompt, we randomly select \textbf{Gemini 3.1 Pro} or \textbf{Claude Opus 4.7} as the teacher agent to generate a tool-use trajectory. Both teachers use the same high-level tool interfaces as \method{} and follow the system prompt in Appendix~\ref{app:system_prompt}. Each teacher uses textual and visual observations to select relevant evidence and construct inputs for subsequent generation calls. We record the complete interaction history, including the teacher's reasoning, tool calls, returned observations, and final generated video. This process yields 24K agent trajectories, of which \textbf{16K} teacher trajectories are used for SFT. The remaining \textbf{8K} prompts are reserved for RL, during which the policy generates new trajectories for reward-based optimization.

\begin{table}[!t]
\centering
\caption{Training prompt counts, commonly used tools, and typical tool-use workflows across six task categories. The workflows illustrate common tool-use patterns; the agent selects tool calls under the system prompt's tool-specific constraints; these sequences are not enforced by code. The 24K prompts are split into 16K for SFT and 8K for RL. T2V, I2V, R2V, and M2V denote text-, image-, reference-, and motion-conditioned video generation. Code denotes code-based simulation, ELF extracts the last frame to condition the next segment, and Obj Det.\ and Depth Est.\ denote object detection and depth estimation. $N$ is the number of named subjects, and $K$ is the number of phases specified in the prompt. Regeneration after verification is performed when needed.}
\label{tab:workflow}
\resizebox{\columnwidth}{!}{
\begin{tabular}{@{}lcll@{}}
\toprule
Task & \# of Prompts & Tools & Typical Workflow \\
\midrule
\taskone{}   & 4K & Search-Text, T2V
& Search-Text $\rightarrow$ T2V \\
\tasktwo{}   & 4K & Search-Image, R2V, I2V
& Search-Image $\rightarrow$ R2V or I2V \\
\taskthree{} & 4K & Search-Image, R2V
& Search-Image$^{\times N}$ $\rightarrow$ R2V \\
\taskfour{}  & 4K & Code, M2V
& Code $\rightarrow$ M2V \\
\taskfive{}  & 4K & T2V, Obj Det., Depth Est.
& T2V $\rightarrow$ (Obj Det., Depth Est.) $\rightarrow$ T2V \\
\tasksix{}   & 4K & T2V, ELF, I2V
& T2V $\rightarrow$ (ELF $\rightarrow$ I2V)$^{\times(K-1)}$ \\
\bottomrule
\end{tabular}
}
\end{table}

\subsection{Stage 1: Supervised Fine-Tuning}
\label{sec:sft}

We fine-tune the shared agent policy $\pi_\theta$ on the 16K teacher trajectories described in Section~\ref{sec:data} using a standard next-token prediction objective. We use the same system prompt as the teacher agents, provided in Appendix~\ref{app:system_prompt}, which specifies the available tools and example tool-use patterns for common prompt types. Failed action spans are masked from the loss but retained in the interaction context together with their error observations. The remaining valid steps, including subsequent recovery actions, continue to provide supervision. We denote the resulting agent policy by $\pi_{\theta_0}$ and use it to initialize reinforcement learning.

\subsection{Stage 2: Agentic RL}
\label{sec:grpo}

Starting from $\pi_{\theta_0}$, we optimize the agent policy using Group Relative Policy Optimization (GRPO)~\citep{shao2024deepseekmath}. For each training prompt $x$ from category $c$, the behavior policy $\pi_{\theta_{\mathrm{old}}}$ samples $n$ rollout trajectories $\{\tau_i\}_{i=1}^{n}$.
Each trajectory contains multi-turn tool calls, returned observations, and a final generated video.
Only agent-emitted tokens contribute to the policy objective; tool-response tokens serve as context and are masked from the loss.
External tool errors are returned as textual observations to allow recovery attempts.
Trajectories that fail because of these errors are discarded and resampled until each prompt has $n=6$ usable rollouts; group statistics are computed on the replenished group.

\subsubsection{Multitask Hybrid Reward}
\label{sec:reward_design}

For a trajectory $\tau$ generated from prompt $x$ in category $c$, we define
\begin{equation}
R(\tau,x,c)
=
\lambda_{\mathrm{f}}R_{\mathrm{format}}(\tau)
+
\lambda_{\mathrm{v}}R_{\mathrm{vlm}}(\tau,x,c)
+
\lambda_{\mathrm{t}}R_{\mathrm{tool}}(\tau,c),
\label{eq:reward}
\end{equation}
with weights $(\lambda_{\mathrm{f}},\lambda_{\mathrm{v}},\lambda_{\mathrm{t}})=(0.1,0.5,0.4)$.
The format reward checks output and tool-call validity.
The VLM reward evaluates final video quality and prompt consistency using category-specific rubrics.
The tool-use reward assesses task-appropriate tool selection and utilization of returned outputs.
Component definitions and VLM rubrics are provided in Appendices~\ref{app:rl_details} and~\ref{app:reward_prompt}, respectively.

\subsubsection{Advantage Normalization}

For the $n$ trajectories sampled from the same prompt, let $R_i=R(\tau_i,x,c)$.
We first compute the group-relative advantage:
\begin{equation}
\hat{A}_i
=
\frac{R_i-\operatorname{mean}_{j=1}^{n}R_j}
{\operatorname{std}_{j=1}^{n}R_j+\epsilon}.
\label{eq:group_advantage}
\end{equation}
Each agent-emitted token in $\tau_i$ receives the advantage $\hat{A}_i$.
Following AgentRL~\citep{agenticrl}, we further normalize these advantages within each task category:
\begin{equation}
\widetilde{A}_{i,k}
=
\frac{\hat{A}_i-\mu_c}{\sigma_c+\epsilon},
\label{eq:task_advantage}
\end{equation}
where $\mu_c$ and $\sigma_c$ are computed over agent-token advantages from category $c$ in the global training batch.
The statistics are aggregated across all data-parallel workers.
This second normalization places token-level advantages from different categories on comparable scales.

\subsubsection{Token-Level GRPO Objective}

Let $z_{i,k}$ denote the $k$-th token in trajectory $\tau_i$, with preceding multimodal context $C_{i,k}$.
For agent-emitted tokens, the importance ratio is
\begin{equation}
\rho_{i,k}(\theta)
=
\frac{\pi_\theta(z_{i,k}\mid C_{i,k})}
{\pi_{\theta_{\mathrm{old}}}(z_{i,k}\mid C_{i,k})}.
\label{eq:token_ratio}
\end{equation}
We optimize the agent policy using
\begin{align}
\mathcal{L}_{\mathrm{GRPO}}(\theta)
=
\mathbb{E}\Bigg[
\frac{1}{Z}
\sum_i\sum_{k\in\mathcal{M}_i}
\Big[
&-\min\!\Big(
\rho_{i,k}\widetilde{A}_{i,k},
\operatorname{clip}(\rho_{i,k},1-\epsilon_{\mathrm{lo}},1+\epsilon_{\mathrm{hi}})
\widetilde{A}_{i,k}
\Big)
\nonumber\\
&+\beta_{\mathrm{KL}}
D_{\mathrm{KL}}\!\Big(
\pi_\theta(\cdot\mid C_{i,k})
\,\Vert\,
\pi_{\mathrm{ref}}(\cdot\mid C_{i,k})
\Big)
\Big]
\Bigg],
\label{eq:grpo}
\end{align}
where $\mathcal{M}_i$ contains the agent-emitted token positions and $Z=\sum_i|\mathcal{M}_i|$. The expectation is over training prompts and rollouts sampled by the behavior policy $\pi_{\theta_{\mathrm{old}}}$. The optional direct KL term uses the fixed SFT reference $\pi_{\mathrm{ref}}=\pi_{\theta_0}$. Additional implementation details are provided in Appendix~\ref{app:rl_details}.

\vspace{-5pt}
\section{Experiments}
\label{sec:experiments}
\vspace{-5pt}
\subsection{Implementation Details}
\label{sec:setup}

We use Qwen3-VL-8B-Instruct~\citep{bai2025qwen3} as the base policy. SFT uses 16K trajectories, AdamW~\citep{adamw} with learning rate $5\times10^{-5}$, and two epochs on eight H200 141GB GPUs with FSDP~\citep{10.14778/3611540.3611569}. RL runs for one epoch on 8K prompts with learning rate $10^{-6}$, 10 warmup steps, and category-stratified batches of 10. Each prompt has six usable rollouts, replenished after external tool failures before computing group statistics. Sampling uses temperature $1.0$ and top-$p=1.0$; clipping uses $(\epsilon_{\mathrm{lo}},\epsilon_{\mathrm{hi}})=(0.2,0.28)$. Only agent tokens enter the loss. RL uses 8 $\times$ H200 GPUs for the policy and eight for tools. Appendix~\ref{app:rl_config} gives further settings.

We use separate T2V, I2V, R2V, and M2V tools to support different conditioning inputs while balancing speed and quality. Toolset~1 is used throughout training. Toolset~2 upgrades the generation tools while retaining the same augmentation and verification tools and compatible interfaces. We evaluate the same trained policy with both toolsets, measuring the benefit of stronger generation tools without additional agent training. Table~\ref{tab:tool_impl} lists their implementations; Appendix~\ref{app:tool_impl} provides verification-tool details.

\begin{table}[H]
\centering
\caption{Concrete tool implementations in Toolset~1 and Toolset~2. The toolsets share augmentation and verification tools but differ in their generation tools.}
\label{tab:tool_impl}
\resizebox{\linewidth}{!}{
\small
\setlength{\tabcolsep}{5pt}
\begin{tabular}{@{}llll@{}}
\toprule
Tool group & Tool & Toolset~1 & Toolset~2 \\
\midrule
\multirow{3}{*}{Augmentation}
& Search-Text & Bing Search & Bing Search \\
& Search-Image & Bing Image Search & Bing Image Search \\
& Code & Custom physics simulator (App.~\ref{app:custom_fn}) & Custom physics simulator \\
\midrule
\multirow{4}{*}{Generation}
& T2V & Seedance 1.0 Pro Fast & Seedance 2.0 Fast \\
& I2V & Seedance 1.0 Pro Fast & Seedance 2.0 Fast \\
& R2V & Wan 2.1-VACE-1.3B references & Seedance 2.0 Fast \\
& M2V & Wan 2.1-VACE-1.3B flow & Wan 2.1-VACE-14B flow \\
\midrule
\multirow{2}{*}{Verification}
& Obj Det. & Grounding DINO 1.5 & Grounding DINO 1.5 \\
& Depth Est. & Depth Anything V2 (ViT-L) & Depth Anything V2 (ViT-L) \\
\bottomrule
\end{tabular}
}
\end{table}

\noindent\textbf{\ourbench{}.}
\ourbench{} contains 600 held-out prompts, with 100 per category, selected manually from 150 candidates after repetition checks and VLM filtering. Following Flow-GRPO and Gen-Searcher~\citep{liu2025flowgrpo,feng2026gensearcher}, we use the RL evaluator, Gemini 3.1 Pro, with the same category-specific rubrics and weights for benchmark evaluation. We separately assess judge--human agreement on 600 pairs and human preference with four raters. Failed operations are retried with at most three attempts (Appendix~\ref{app:evaluation_retries}); all 600 prompts contribute to category means on a $[0,100]$ scale. Appendix~\ref{app:data_construction} describes construction and splits; Appendices~\ref{app:reward_prompt} and~\ref{app:evaluation_references} give rubrics, fixed shared references, and assessment procedures.

\noindent\textbf{Baselines.}
We compare \method{} against open-source and proprietary text-to-video models used in a single-pass setting. Open-source baselines include CogVideoX-5B~\citep{yang2024cogvideox}, Mochi-1~\citep{genmo2024mochi}, HunyuanVideo-13B~\citep{kong2024hunyuanvideo}, and Wan2.1-T2V-14B~\citep{wan2025wan}. Proprietary baselines include Hailuo 2.0~\citep{minimax2025hailuo02}, Kling 3.0~\citep{kling2026video3}, Seedance 1.0 Pro Fast~\citep{gao2025seedance}, and Seedance 2.0 Fast~\citep{seedance2026seedance}. We evaluate \method{} with both toolsets against these standalone generators.

\subsection{Main Results}
\label{sec:main_results}
Table~\ref{tab:per_cat} compares \method{} with standalone video generators on \ourbench{} under the evaluation protocol in Section~\ref{sec:setup}. With Toolset~1, \method{} achieves an overall score of $75.6$, improving on its base T2V generator, Seedance 1.0, by $19.1$ points. It also exceeds Seedance 2.0, the strongest standalone baseline overall, by $2.4$ points, although it remains below this model on \taskthree{} and \taskfive{}. With Toolset~2, \method{} reaches $86.1$ overall and achieves the highest score in every category.

\begin{table}[!t]
\centering
\caption{
Performance on \ourbench{} evaluated by Gemini 3.1 Pro using category-specific rubrics.
Scores are normalized to $[0,100]$ and averaged within each category.
Overall is the unweighted mean across all six categories.
Bold and underline mark the best and second-best results in each column.
Evaluation dimensions and scoring weights are provided in Appendix~\ref{app:reward_prompt}.
\taskoneab{}, \tasktwoab{}, \taskthreeab{}, \taskfourab{}, \taskfiveab{}, and \tasksixab{} denote
\taskone{}, \tasktwo{}, \taskthree{}, \taskfour{}, \taskfive{}, and \tasksix{}, respectively.
}
\label{tab:per_cat}
\footnotesize
\renewcommand{\arraystretch}{1.0}
\setlength{\tabcolsep}{2.4pt}
\resizebox{0.7\linewidth}{!}{%
\begin{tabular}{lccccccc}
\toprule
Model
& \taskoneab{}
& \tasktwoab{}
& \taskthreeab{}
& \taskfourab{}
& \taskfiveab{}
& \tasksixab{}
& Overall \\
\midrule

\rowcolor{black!7}
\multicolumn{8}{c}{\emph{Open-Source Video Generators}}\\
\rowcolor{white}
CogVideoX-5B~\citep{yang2024cogvideox}
& 38.9 & 48.3 & 37.8 & 41.6 & 50.4 & 30.8 & 41.3 \\
Mochi-1~\citep{genmo2024mochi}
& 41.9 & 49.6 & 37.8 & 48.4 & 55.4 & 28.3 & 43.6 \\
HunyuanVideo-13B~\citep{kong2024hunyuanvideo}
& 37.5 & 29.3 & 17.6 & 31.6 & 31.5 & 31.1 & 29.8 \\
Wan2.1-T2V-14B~\citep{wan2025wan}
& 42.0 & 46.2 & 39.5 & 49.5 & 60.8 & 62.0 & 50.0 \\

\arrayrulecolor{black!25}\midrule\arrayrulecolor{black}
\rowcolor{black!7}
\multicolumn{8}{c}{\emph{Proprietary Video Generators}}\\
\rowcolor{white}
Kling 3.0~\citep{kling2026video3}
& 64.4 & 62.3 & 55.2 & 56.7 & 86.5 & 66.1 & 65.2 \\
Hailuo 2.0 (MiniMax)~\citep{minimax2025hailuo02}
& 63.0 & 58.6 & 52.4 & 55.1 & 86.0 & 67.6 & 63.8 \\
Seedance 1.0 Pro Fast~\citep{gao2025seedance}
& 55.5 & 51.4 & 44.2 & 46.2 & 74.6 & 66.8 & 56.5 \\
Seedance 2.0 Fast~\citep{seedance2026seedance}
& 78.0 & 75.0 & \underline{69.0} & 60.2 & \underline{87.8} & 68.9 & 73.2 \\

\arrayrulecolor{black!25}\midrule\arrayrulecolor{black}
\rowcolor{black!7}
\multicolumn{8}{c}{\emph{VideoGen-Agent}}\\
\rowcolor{ourcolor}
\method{}-Toolset~1
& \underline{80.4}
& \underline{75.1}
& 65.3
& \underline{67.9}
& 83.1
& \underline{81.7}
& \underline{75.6} \\
\rowcolor{ourcolor}
\method{}-Toolset~2
& {\bfseries 92.1}
& {\bfseries 83.2}
& {\bfseries 86.7}
& {\bfseries 69.2}
& {\bfseries 90.7}
& {\bfseries 94.6}
& {\bfseries 86.1} \\
\bottomrule
\end{tabular}}
\end{table}

\noindent\textbf{Generation Tool Upgrades.}
Replacing the generation tools in Toolset~1 with those in Toolset~2 raises the overall score from $75.6$ to $86.1$. The agent policy remains fixed, and the augmentation and verification tools are unchanged. Scores improve in all six categories, with the largest increase on \taskthree{}, from $65.3$ to $86.7$. These results show that the trained agent can benefit from compatible generation tools not encountered during training, without additional policy optimization.

\noindent\textbf{Per-category Analysis.}
\label{subsec:percat}
The teaser on the first page illustrates task-specific tool use across all six categories, showing how \method{} adapts its workflow to different generation requirements through procedural retrieval, visual-reference grounding, physics simulation, verification-guided refinement, and sequential generation; additional qualitative comparisons are provided in Figures~\ref{fig:ab:case_kp}--\ref{fig:ab:case_ms} in the appendix. The category-level results show where tool-augmented generation provides the largest gains. Relative to Seedance 2.0, \method{}-Toolset~2 improves \taskone{} by $14.1$ points, \taskthree{} by $17.7$ points, and \tasksix{} by $25.7$ points. These categories require detailed procedural knowledge, multiple visual references, or explicit temporal structure, which their respective workflows supply through retrieval and sequential generation. Gains on \tasktwo{} and \taskfour{} are $8.2$ and $9.0$ points, respectively, consistent with the use of visual references and simulated motion. The improvement on \taskfive{} is smaller at $2.9$ points, with the strongest standalone baseline already scoring $87.8$. This variation indicates that the benefit of tool augmentation depends on the capability required by the prompt.

\subsection{Ablation Studies}
\label{sec:ablations}

\begin{table}[!t]
\centering
\caption{Ablations on \ourbench{}. Rows (i)--(iv) and (viii) compare vanilla T2V, prompt rewriting, a fixed agentic workflow, SFT, and RL under the Toolset~1 configuration. Rows (v) and (vi) remove individual reward components, and row (vii) reports the single-task variant. Overall is the unweighted mean across all six categories. The highest score in each column is bolded.}
\label{tab:ablation}
\resizebox{1\linewidth}{!}{
\small
\setlength{\tabcolsep}{4.5pt}
\begin{tabular}{@{}llccccccc@{}}
\toprule
& Variant & \taskoneab{} & \tasktwoab{} & \taskthreeab{} & \taskfourab{} & \taskfiveab{} & \tasksixab{} & Overall \\
\midrule
(i) & Vanilla T2V (Seedance 1.0 Pro Fast)
& 55.5 & 51.4 & 44.2 & 46.2 & 74.6 & 66.8 & 56.5 \\
(ii) & \backbone{} + Prompt Rewriting + Toolset 1
& 57.3 & 51.5 & 45.3 & 44.6 & 77.2 & 68.1 & 57.3 \\
(iii) & \backbone{} + Fixed Workflow + Toolset~1
& 63.3 & 59.6 & 56.3 & 56.3 & 75.5 & 73.8 & 64.1 \\
(iv) & \method{}-SFT-Toolset~1
& 72.7 & 68.1 & 58.6 & 60.4 & 79.2 & 76.3 & 69.2 \\
\midrule
(v) & \method{}-RL-Toolset~1 w/o VLM Reward
& 76.4 & 74.3 & 63.9 & 66.5 & 80.1 & 78.7 & 73.3 \\
(vi) & \method{}-RL-Toolset~1 w/o Tool Reward
& 74.3 & 70.2 & 60.2 & 60.9 & 82.7 & 78.8 & 71.2 \\
(vii) & \method{}-RL-Single Task
& {\bfseries 85.5} & 74.5 & {\bfseries 65.9} & 67.3 & {\bfseries 84.4} & 80.3 & {\bfseries 76.3} \\
(viii) & \method{}-RL-Toolset~1 (Full)
& 80.4 & {\bfseries 75.1} & 65.3 & {\bfseries 67.9} & 83.1 & {\bfseries 81.7} & 75.6 \\
\bottomrule
\end{tabular}}
\end{table}

\label{sec:sft_vs_rl}
Table~\ref{tab:ablation} compares training stages, reward components, and single-task versus multitask RL. The fixed-workflow baseline uses the same \backbone{} and Toolset~1 without task-specific training. For \taskthree{}, the backbone is given the required entity count $N$, lists $N$ names, retrieves one reference image per entity, and calls \texttt{video\_gen\_multiple\_reference} for R2V generation. For \taskfive{}, ReAct is capped at three iterations; other tasks execute their prescribed workflow once. Appendix~\ref{app:baseline_protocol} details these task-specific budgets and the single-task training setup. 

Prompt rewriting improves the overall score from $56.5$ to $57.3$, while the fixed workflow reaches $64.1$. SFT further improves performance to $69.2$, and RL adds $6.4$ points to reach $75.6$, with gains across all six categories. The advantage over the fixed workflow supports the usefulness of training under these task-specific baseline budgets.

Removing the VLM reward reduces the overall score to $73.3$, while removing the tool reward lowers it to $71.2$, with both ablations degrading all six categories. Single-task agents undergo SFT and then RL using only their category-specific subsets, with other settings unchanged. They achieve $76.3$ overall, $0.7$ points above the shared multitask agent. Single-task training performs better on \taskone{}, \taskthree{}, and \taskfive{}, while multitask training leads on the remaining categories and supports all six tasks with a single policy at comparable overall performance.

\vspace{-5pt}

\subsection{Generalization to an Unseen Task Combination}
\label{sec:combo}
We evaluate \method{} on \taskcombo{} (\taskcomboab{}), a held-out combination of \taskone{} and \tasktwo{}. We combine 20 specialized actions or processes and 20 named entities from a randomly held-out subset of the original prompt pool, disjoint from SFT, RL, and VABench prompts, and retain 100 prompts through duplicate removal and VABench-style filtering and manual review (Appendix~\ref{app:combo}). No training prompt combines these requirements. The system prompt provides reference workflows and general composition guidance, but no explicit workflow for this particular combination. Without additional training, the RL agent scores 73.2, compared with 57.6 for SFT and 72.4 for the composed fixed workflow. It invokes both retrieval tools in 84\% of trajectories, versus 40\% for SFT and 100\% for the fixed workflow by construction. These results show improved performance after RL on this held-out combination, with a score close to that of the explicitly composed workflow. Table~\ref{tab:combo} reports scores and retrieval-tool invocation rates. The fixed baseline executes Search-Text, Search-Image, and reference-conditioned generation once per prompt with Toolset~1; its generation-call count is not matched to the RL trajectory. Tool invocation rates describe agent behavior, not whether the resulting video satisfies both requirements. Appendix~\ref{app:combo} provides the complete construction and scoring protocol.

\begin{table}[H]
\centering
\caption{Zero-shot results on \taskcombo{} (\taskcomboab{}), an unseen combination of \taskone{} and \tasktwo{} (100 prompts). Tool-use columns report the percentage of trajectories that call each retrieval tool at least once. The fixed workflow invokes both retrieval tools by construction.}
\label{tab:combo}
\small
\setlength{\tabcolsep}{6pt}
\begin{tabular}{@{}lcccc@{}}
\toprule
Model & Search-Text (\%) & Search-Image (\%) & Both (\%) & Score \\
\midrule
\backbone{} + Fixed Workflow & 100 & 100 & 100 & 72.4 \\
\method{}-SFT-Toolset~1 & 58 & 78 & 40 & 57.6 \\
\method{}-RL-Toolset~1 & 88 & 96 & 84 & 73.2 \\
\bottomrule
\end{tabular}
\end{table}
\FloatBarrier

\subsection{Human Evaluation and Judge Validation}
\label{sec:human_eval}

% \noindent\textbf{Human Preference.}
% We conduct a side-by-side evaluation on 100 randomly sampled video pairs from \ourbench{}, comparing \method{}-Toolset~2 with Seedance 2.0. Four raters each select one preferred video per pair, with no tie option. Across 400 judgments, the preference rate for \method{}-Toolset~2 is $84.3\%$ (rounded to one decimal place). In an additional human preference comparison, \method{}-Toolset~1 achieves an $88\%$ preference rate over its standalone text-to-video generator, Seedance 1.0 Pro Fast.

\paragraph{Human Preference.}
We conduct two side-by-side human evaluations under the same
protocol, each using 100 randomly sampled video pairs from
\ourbench{} and four raters. Each rater selects one preferred
video per pair, with no tie option, yielding 400 judgments per
comparison. \method{}-Toolset~2 is preferred over Seedance~2.0
in $84.3\%$ of judgments, while \method{}-Toolset~1 is preferred
over its standalone text-to-video generator, Seedance~1.0 Pro
Fast, in $88\%$ of judgments.

\paragraph{Judge--Human Agreement.}
\label{app:reward_alignment} We assess whether the VLM reward used during RL agrees with human preferences. Gemini~3.1~Pro provides the training reward $R_{\mathrm{vlm}}$ using the category-specific rubrics in Appendix~\ref{app:reward_prompt}. This analysis evaluates the training judge; the main benchmark results use Gemini~3.1~Pro with the same rubrics and scoring weights.

\paragraph{Judge-validation protocol.}
We sample $100$ video pairs per category ($600$ pairs in total), balanced across the strongest standalone baseline, \method{}-SFT-Toolset~1, and \method{}-RL-Toolset~1. Three human raters independently compare each pair without knowing which models generated the videos. They assess the category-specific requirements, including procedural accuracy for \taskone{}, identity preservation for \tasktwo{} and \taskthree{}, and physical plausibility for \taskfour{}. For \taskfive{} and \tasksix{}, they assess subject interactions and temporal phase structure, respectively. Overall prompt faithfulness serves as the tie-breaker. We compare the human majority preference with the judge's preference, initially determined by its category-specific video scores. If the two videos receive equal scores, we re-prompt the VLM with both videos and the same category-specific rubric, explicitly requiring it to select the better video; a tie is not an allowed response. The resulting forced-choice preference is used for the agreement calculation, separately from the scalar video-quality scores reported in the benchmark.

\paragraph{Agreement results.}
Gemini~3.1~Pro agrees with the human majority on $92.0\%$ of the $600$ evaluated pairs. Agreement rates are $72\%$, $98\%$, $100\%$, $94\%$, $88\%$, and $100\%$ across the six task categories, respectively. Agreement is lowest for \taskone{}, suggesting that specialized procedural knowledge remains a challenge for the judge. Across the other five categories, agreement ranges from $88\%$ to $100\%$, supporting its use for evaluating identity, physical dynamics, subject interactions, and temporal order.
% \paragraph{Agreement results.}
% As shown in Table~\ref{tab:judge_human}, Gemini~3.1~Pro agrees with the human majority on $92.0\%$ of the $600$ evaluated pairs. Agreement rates are $72\%$, $98\%$, $100\%$, $94\%$, $88\%$, and $100\%$ across the six task categories, respectively. Agreement is lowest for \taskone{}, suggesting that specialized procedural knowledge remains a challenge for the judge. Across the other five categories, agreement ranges from $88\%$ to $100\%$, supporting its use for evaluating identity, physical dynamics, subject interactions, and temporal order.

% \begin{table}[!htbp]
% \centering
% \caption{Agreement between Gemini~3.1~Pro and human majority preferences on 100 video pairs per category (600 total). Each pair is assessed by three human raters. This study is separate from the four-rater human-preference comparison.}
% \label{tab:judge_human}
% \small
% \begin{tabular}{lrrrrrrr}
% \toprule
%  & PK & SI & MI & PS & CS & MS & Overall \\
% \midrule
% Agreement (\%) & 72.0 & 98.0 & 100.0 & 94.0 & 88.0 & 100.0 & 92.0 \\
% \bottomrule
% \end{tabular}
% \end{table}
% \FloatBarrier

\section{Related Work}
\label{sec:related}

\paragraph{Video Generation and Tool Augmentation}
Video generation has advanced through diffusion-based models~\citep{ho2022video,blattmann2023stable,polyak2024movie,wan2025wan,yang2024cogvideox} and autoregressive approaches that generate visual sequences incrementally~\citep{yu2024magvit2,yan2025loong}. Despite improvements in visual quality, depicting specific entities, physical interactions, and ordered events remains challenging.

To address these limitations, recent systems augment generators with external planning and control. Scene decomposition and editing pipelines organize generation into smaller steps~\citep{automv,huang2026vimaxagenticvideogeneration,mu2026scriptneedagenticframework}, while physics-guided methods use simulation or force-based controls to guide motion~\citep{newtongen,feng2026newton,li2026videococo}. These approaches illustrate how specialized workflows can support requirements that are difficult to satisfy through prompting alone. Our work builds on these capabilities by training a shared agent policy to use different tools across multiple generation tasks.
\paragraph{Agentic Reinforcement Learning with Tool Use}
Reinforcement learning with verifiable rewards improves multi-step reasoning by optimizing models against task outcomes~\citep{deepseek2025r1,shao2024deepseekmath}. Agentic RL extends this approach to trajectories that interleave reasoning with external tool calls. Search-R1~\citep{jin2025searchr1}, R1-Searcher~\citep{song2025r1searcher}, and ReSearch~\citep{chen2025research} train agents to retrieve information during reasoning. DeepResearcher~\citep{zheng2025deepresearcher} and WebDancer~\citep{wu2025webdancer} study extended information-seeking workflows, while ToRL~\citep{li2025torl} incorporates code execution.

Related multimodal research develops visual understanding~\citep{liu2024llava,wang2024qwen2vl}, GUI interaction~\citep{cheng2024seeclick,lin2024showui}, and embodied control~\citep{huang2023voxposer,brohan2023rt2}. These settings primarily concern answering questions or completing actions. Video generation requires evaluating the produced visual content as well as the tool-use decisions that lead to it.
\paragraph{Agentic Visual Generation}
Agentic image generation uses external information to ground visual synthesis, as illustrated by the ``research-then-generate'' approach~\citep{gptimage2}. Gen-Searcher~\citep{feng2026gensearcher} trains a search-augmented image-generation agent through SFT followed by GRPO, using a joint text--image reward. This establishes a connection between learned tool use and visual generation, but focuses on image synthesis with search augmentation.

For video, VISTA~\citep{Long_2026_CVPR} performs test-time refinement through iterative generation, multimodal critique, and prompt revision. \method{} instead trains an agent policy across six video-generation tasks, using workflow guidance to coordinate retrieval, simulation, generation, and verification. Its multitask hybrid reward evaluates tool-call validity and task-appropriate tool use alongside category-specific video quality.

\vspace{-5pt}
\section{Conclusion}
\vspace{-5pt}
We introduced \textbf{\method{}}, a multimodal agent that learns to use external tools across diverse video-generation tasks. SFT followed by multitask RL trains a shared agent policy to coordinate augmentation, generation, and verification tools. We also constructed a category-balanced training dataset and \ourbench{}, a held-out benchmark covering six generation capabilities. Experiments show gains over single video generators, with ablations supporting the contributions of both training stages. Upgrading the generation tools further improves performance without additional agent training, demonstrating complementary benefits from learned tool use and advances in generation models.

\paragraph{Limitations and Future Work}
Our evaluation covers six video-generation tasks and one held-out task combination; broader compositional generalization remains to be studied. Output quality and runtime also depend on the available tools. Future work will explore broader task coverage and more efficient generation and feedback mechanisms.

\clearpage

\clearpage
\bibliographystyle{abbrvnat}
\bibliography{arxiv_v2/main,arxiv_v2/molmov1,arxiv_v2/main_nips}

\clearpage

\clearpage
\appendix
\raggedbottom
\makeatletter
\setlength{\@fptop}{0pt}
\makeatother
\section*{Appendix Overview}
The appendix is organized as follows:
\begin{itemize}[leftmargin=*,itemsep=2pt]
\item Appendix~\ref{app:data_construction}: training-prompt and VABench construction, quality filtering, and the data split.
\item Appendix~\ref{details}: agent rollout, tool implementations, agent system prompt, and the physics simulator.
\item Appendix~\ref{app:rl_details}: reward components and the RL objective.
\item Appendix~\ref{app:evaluation}: scoring criteria, duration handling, physical checks, and judge prompts.
\item Appendix~\ref{result}: qualitative comparisons, the unseen task-combination evaluation protocol, and baseline details.
\item Appendix~\ref{app:robotics}: exploratory robotic-manipulation examples.
\item Appendix~\ref{app:data_system_prompts}: one initial data-generation prompt for each of the six task categories.
\end{itemize}

\section{Training Prompts and VABench}
\label{app:data_construction}
\label{app:vabench_construction}

\subsection{Construction}
\label{app:prompt_construction}
We construct a candidate pool across the six task categories in Section~\ref{sec:task_tools}, using Claude Opus 4.7 to generate prompts in batches of 100 for each category. After each batch, we read all previously generated prompts for that category and check the new candidates for repetition in the task-defining content. For example, for \tasktwo{}, this check focuses on the named characters or entities rather than wording alone. We subsequently apply an additional VLM-based filtering pass using Claude Sonnet 4.6.

Category-specific initial prompts define the target capability and content constraints. Visual-identity prompts use text and image retrieval to ground the named subjects and pair them with simple actions or explicitly explained signature actions. For a signature skill, the prompt itself details the visible movements, props, and effects rather than relying on its name. We also adapt grounded image-generation examples from Gen-Searcher~\citep{feng2026gensearcher} into video descriptions. Appendix~\ref{app:data_system_prompts} lists one initial generation prompt for each of the six tasks.

\subsection{Quality Filtering and Data Split}
\label{app:prompt_quality}
\label{app:prompt_split}
We use Claude Sonnet 4.6 to screen candidates for specificity, factual plausibility, observable motion, and alignment with the intended task. Repetition is assessed through the underlying task content, so superficial changes in wording or camera style do not establish a distinct example. Following candidate filtering, we shortlist 150 prompts per category for manual review. Human reviewers assess whether each prompt isolates the intended capability using the following category-specific criteria:
\begin{itemize}[leftmargin=*,itemsep=1pt,parsep=0pt]
    \item \textbf{\taskone{}:} prioritize processes whose knowledge-dependent steps can be clearly demonstrated visually.
    \item \textbf{\tasktwo{} and \taskthree{}:} use simple actions or provide detailed, self-contained descriptions of any signature skills, so that identity preservation remains the target capability without requiring implicit procedural knowledge. This criterion applies to the actual SFT, RL, and VABench prompts.
    \item \textbf{\taskfour{}:} retain only quantitative collision and falling/dropping prompts with reasonable spatial and motion scales; require both masses and initial velocities for collisions, and release height and initial motion for drops; exclude fluid and other unsupported phenomena.
    \item \textbf{\taskfive{}:} require complex multi-subject spatial relationships, interactions, and temporal changes that test scene composition.
    \item \textbf{\tasksix{}:} first choose two or three shots, then assign an identical duration of at most five seconds to every shot. Require the total duration and each shot's content, expressed through adjacent time intervals or an explicitly equal-length shot sequence, with plausible continuity.
\end{itemize}
This manual review selects 100 prompts per category, yielding the final 600-prompt \ourbench{}, which is excluded from SFT and RL. The training collection contains 24K prompts (4K per category; Table~\ref{tab:workflow}), split into 16K teacher trajectories for SFT and 8K prompts for RL rollouts. The separate 100-prompt \taskcombo{} test (Appendix~\ref{app:combo}) is excluded from the \ourbench{} aggregate.

\section{Agent and Tool Implementation}
\label{details}

\subsection{Agent Rollout Procedure}
\label{app:agent_rollout}

Algorithm~\ref{alg:agent} gives the complete rollout procedure for the multitask video-generation agent described in Section~\ref{sec:method}. The agent iteratively reasons over the interaction history, executes selected tools, retains returned observations, and terminates when the agent emits a valid final answer with all requested segments available, or when the step budget is exhausted. We use a training interaction budget of $T_{\max}=10$ and an evaluation budget of $T_{\max}=20$. Verification outputs are returned as observations; the policy decides whether to regenerate or emit a final answer. No verification-score threshold is hard-coded into the rollout termination rule. The agent decides whether to regenerate or finish; each successful regeneration replaces the previous candidate for that segment, and the latest complete sequence is returned. Evaluation retries for tool errors and incomplete trajectories are specified in Appendix~\ref{app:evaluation_retries}; retrying a failed call is separate from choosing to regenerate a valid candidate.

\begin{algorithm}[!htbp]
\small
\caption{\method{} rollout for multitask video generation.}
\label{alg:agent}
\begin{algorithmic}[1]
\REQUIRE prompt $x$; policy $\pi_\theta$; tools $\mathcal{T}$; step budget $T_{\max}$ (10 for training; 20 for evaluation)
\ENSURE generated video $y$, or $\varnothing$ if no complete sequence is available
\STATE $H_1 \leftarrow [x]$; $\mathcal{V} \leftarrow \varnothing$
\COMMENT{interaction history and current video segments}
\FOR{$t = 1, \ldots, T_{\max}$}
    \STATE Sample action $a_t \sim \pi_\theta(\cdot \mid H_t)$
    \IF{$a_t$ is \texttt{<answer>}}
        \IF{all requested phases have video candidates in $\mathcal{V}$}
            \STATE \textbf{break}
        \ENDIF
        \STATE $o_t \leftarrow$ ``Generate the missing video segments before terminating.''
    \ELSE
        \STATE $(u,\phi) \leftarrow a_t$
        \COMMENT{tool name and arguments}
        \STATE $o_t \leftarrow \textsc{Execute}(u,\phi)$
        \COMMENT{evaluation: at most 3 attempts per call; output or error observation}
        \IF{execution succeeds}
            \IF{$u \in \mathcal{T}_{\mathrm{gen}}$}
                \STATE Update $\mathcal{V}$ with the generated segment
                \COMMENT{replace the corresponding candidate if regenerated}
                \STATE Invalidate verification results and later segments dependent on the replaced segment
            \ELSIF{$u \in \mathcal{T}_{\mathrm{ver}}$}
                \STATE Record verification observations for the evaluated candidate
                \COMMENT{the policy uses this feedback in subsequent decisions}
            \ENDIF
        \ENDIF
    \ENDIF
    \STATE $H_{t+1} \leftarrow H_t \,\Vert\, (a_t,o_t)$
    \COMMENT{retain outputs and errors for subsequent decisions}
\ENDFOR
\STATE $y \leftarrow \textsc{Assemble}(\mathcal{V})$
\COMMENT{return the latest complete sequence in phase order; $\varnothing$ if incomplete}
\STATE \textbf{return} $y$
\end{algorithmic}
\end{algorithm}
\FloatBarrier

\subsection{Concrete Tool Implementations}
\label{app:tool_impl}

Table~\ref{tab:tool_impl} lists the models and services used for each tool in Toolset~1 and Toolset~2. Both toolsets contain augmentation, generation, and verification tools, following the organization in Section~\ref{sec:task_tools}. Toolset~1 is used throughout training. Toolset~2 upgrades the generation tools while retaining the same augmentation and verification tools. We evaluate the same trained agent policy with both toolsets without additional training.

For video verification, we uniformly sample frames and apply Grounding DINO to localize the queried subjects using text labels. The \texttt{grounding\_dino\_video} tool returns detection boxes and their confidence scores. The \texttt{depth\_anything\_video} tool additionally applies Depth Anything to the same frames and computes the median predicted depth within each subject's Grounding DINO bounding box. These per-subject medians are compared within the same frame to assess camera-relative depth ordering. They represent relative depth rather than metric distance; subjects without valid detections receive no depth estimate.

\paragraph{Detection coverage and learned regeneration decisions.}
For each queried subject, \texttt{grounding\_dino\_video} computes
\[
d = \texttt{detection\_ratio}
  = \frac{\text{number of sampled frames containing the subject}}
         {\text{total number of sampled frames}}.
\]
The \texttt{found} flag is true whenever $d>0$, so detection in a single sampled frame is sufficient to mark a subject as found. For a found subject, the returned subject-coverage reward is $r_{\mathrm{coverage}}=0.5+0.5d$, reaching $1$ when the subject is detected in every sampled frame. This returned signal is distinct from a detector's per-box confidence threshold.

During RL, the policy develops an approximate decision boundary around $d\approx0.6$ for accepting a candidate versus regenerating it: for example, detection in roughly six of ten sampled frames can suffice for acceptance, whereas lower coverage tends to trigger regeneration. This is an observed policy tendency, not a hard-coded rule or a threshold explicitly supplied in the system prompt. At $d=0.6$, the coverage reward is $0.8$; the reported $0.6$ refers to detection ratio, not reward magnitude. Relative-depth outputs are separate observations used to assess the requested spatial ordering and are not compared with this coverage threshold.

\FloatBarrier

\subsection{Agent System Prompt}
\label{app:system_prompt}
The following supplied prompt documents the teacher and student tool interfaces, together with example tool-use patterns for common prompt types. These patterns are provided as guidance; no code enforces a particular tool sequence, and the agent selects calls subject to the tool-specific instructions below. The composition guidance and depth-tool interface are those used in the reported experiments. It is distinct from the data-construction prompts in Appendix~\ref{app:data_system_prompts}.
\begin{promptbox}
You are a video generation assistant. You CANNOT produce a video yourself — you must complete the task by invoking the provided tools to ground the prompt in external evidence, generate the video, and verify the result when applicable before submitting the final answer. Do NOT answer the user directly, describe a video in words, or emit `<answer>` before at least one video-generation tool has been called successfully. The available tools and reference workflows for common prompt types are listed below.

[Output Format]

Always start your response with a <think>...</think> block to reason about what to do next:

  <think>your reasoning here</think>
  ...

The think block is typically followed by ONE <tool_call> block to invoke a tool:

  <think>your reasoning here</think>
  <tool_call>
  {"name": "tool_name", "arguments": {...}}
  </tool_call>

Tool-call rules:
- Output ONLY ONE <tool_call> block per turn, then STOP immediately.
- Do NOT write multiple tool calls in one turn.
- Do NOT write any text after </tool_call>.
- Wait for the tool response before deciding the next action.

When you have used the appropriate tools to generate and, when applicable, verify the final video, end the trajectory with a SINGLE final assistant turn containing a brief <think>...</think> block evaluating prompt-video alignment, followed by an <answer>...</answer> stop signal:

  <think>I think the generated video aligns well with the prompt.</think>
  <answer>Video generated. Task complete.</answer>

This final turn:
- Begins with a brief <think>...</think> block evaluating prompt-video alignment.
- Is followed immediately by an <answer>...</answer> tag, with no bridge text.
- Does NOT include narrative text, markdown summaries, or video paths outside the <think> / <answer> tags.


[Candidate Retention]
You decide whether to generate again, continue to the next phase, or finish, based on the prompt and tool observations. For each segment, every successful regeneration replaces its previous candidate; retain the LATEST successfully generated video, without ranking or reverting to older candidates. After replacing an earlier segment, extract its new final frame and regenerate all later dependent segments before finishing. Submit the latest complete sequence in phase order.

[Available Tools]

search
    Search the web for factual knowledge about real processes, phenomena, or events.
    Arguments: {"queries": ["query1", "query2"]}

image_search
    Find reference images for specific subjects, including named people, characters, animals, objects, and landmarks.
    Arguments: {"query": "descriptive text"}

    Returns up to 5 candidate images per call, formatted like:
      --- image search result for [query] ---
      IMG_001: /path/to/image_001.jpg
      IMG_002: /path/to/image_002.jpg
      ...
      --- end of image search result ---

    Each `<image>` tag contains the actual image content. After seeing the candidates, in your NEXT <think> block you MUST explicitly evaluate them and pick ONE IMG_K per entity. Cite the IMG_K identifier and give a brief visual reason for the selection. When calling video_gen_single_reference or video_gen_multiple_reference, pass only the local_path of the SELECTED IMG_K into ref_images, NOT all candidates. IMG numbering is continuous across image_search calls within a trajectory: IMG_001..IMG_005, then IMG_006..IMG_010, and so on.

simulation
    Run a physics simulation for collision or falling/dropping prompts. These are the only physics tasks used in training and VABench. Pick ONE mode through the "mode" field. Returns two MP4 files: a raw rendered video and an optical-flow video. Pass the returned optical-flow path, NOT the raw video, to video_gen_simulation.

    World coordinates (collision / freefall):
      x in [-7, 7]     (origin 0,0 is the CENTER of the 832×480 frame)
      y in [-4.5, 4.5] (y > 0 up, y < 0 down)
      Floor (freefall only) sits at y = -3.8.
      Gravity g = 9.8 units/s².
      Total video = 5 s at 24 fps.

    Velocity is in world units per second, NOT pixels:
      v = 0            still / stationary target.
      |v| 0.3 – 1.5    slow / gentle (crosses 8–40 % of frame in 5 s).
      |v| 1.5 – 3.5    natural speed — PREFERRED. Collision usually in frame.
      |v| 3.5 – 5      fast / forceful. Object may exit frame after impact.
      |v| > 5          too fast — usually exits frame before anything resolves.

    Angular velocity omega (rad/s):
      0: none.
      1–3: visible spin.
      6–10: fast spin.

    mode = "collision" — 2 rigid bodies colliding, without gravity:
      `a` and `b` are independently configurable rigid bodies. For EACH body, choose `shape` independently from "sphere", "cube", "cuboid", or "triangular_prism". Neither label imposes a shape restriction; the two bodies may have the same or different shapes.

      How to assign `a` and `b`:
        - Assign either object to `a` and the other to `b`, and keep the assignment consistent within the call.
        - Specify each body's `shape`, `x`, `y`, `vx`, `vy`, `mass`, `angle`, `omega`, `size`, and `size2` independently, using values appropriate to that object and within the tool's valid ranges.
        - No parameter of `a` determines or constrains the corresponding parameter of `b`.
        - Either body may be moving or stationary, rotating or non-rotating.
        - Neither body is required to be smaller, lighter, or the incoming object.
        - `restitution` is a shared collision parameter, not a per-body parameter.

      Velocities are in the WORLD frame; (`vx`, `vy`, `omega`) are independent for `a` and `b`. Examples:
        - Both moving, head-on:     a.vx=+2, b.vx=-2 (closing speed = 4)
        - Both moving, same way:    a.vx=+3, b.vx=+1 (rear-end / catch-up)
        - Crossing paths:          a.vy=+2, b.vx=-2
        - One moving, one still:   a.vx=+2, b.vx=0
        - Glancing blow + spin:    a.vx=+2, b.vy=+0.3, b.omega=4

      Pick velocities that match the prompt's described motion in world coordinates. Do NOT assume one object must be at rest.

      {"mode":"collision",
       "a":{"shape":"sphere"|"cube"|"cuboid"|"triangular_prism",
            "x":-4,"y":0,"vx":2,"vy":0,"mass":1,
            "angle":0,"omega":0,"size":0.5,"size2":0.5},
       "b":{"shape":"sphere"|"cube"|"cuboid"|"triangular_prism",
            "x":3,"y":0,"vx":-1,"vy":0,"mass":3,
            "angle":0,"omega":1.5,"size":0.5,"size2":0.5},
       "restitution":1.0}

    mode = "freefall" — a single body falls onto the floor, with rotation:
      {"mode":"freefall","shape":"circle"|"rect"|"rectlong"|"triangle",
       "x":0,"y":3.5,"vx":0.5,"vy":0,"angle":0,"omega":6,
       "size":0.5,"size2":0.5}

extract_last_frame
    Save the final frame of an existing video as a PNG for use as the first_frame of a subsequent video_gen_first_frame call.
    Arguments: {"video_path": "/path/to/video.mp4"}
    Returns the saved PNG path.

grounding_dino_video
    Detect objects across sampled video frames to verify subject presence.
    Arguments: {"video": "/path/to/video.mp4", "query": "object1 . object2", "n": 6, "start": 0, "end": 5}
      video: path returned by a previous video-generation call; NEVER invent a path.
      query: object descriptions separated by periods.
      n: total frames sampled evenly from the full video.
      start: first sampled-frame index to check, starting from 0.
      end: last sampled-frame index to check, inclusive.

      Examples:
        Full video: n=6, start=0, end=5.
        First half: n=6, start=0, end=2.
        Second half: n=6, start=3, end=5.
        First third: n=9, start=0, end=2.

    Use grounding_dino_video to verify detectable generic objects. It does not verify specialized processes, named identities, or physical correctness. Do NOT use it to determine whether a named character is the correct character, whether a scientific phenomenon is accurate, or whether simulated motion obeys physics.

    For named entities, detection of a generic class such as "person" does not establish identity preservation.

depth_anything_video
    Compare the relative depth of two text-specified objects in sampled video frames. This tool combines text-guided object localization with Depth Anything depth estimation and returns the median relative-depth value within each localized object region.

    Arguments: {"video": "/path/to/video.mp4", "labels": ["person", "apple"], "n": 6, "start": 0, "end": 5}
      video: path returned by a previous video-generation call; NEVER invent a path.
      labels: exactly two text labels identifying the objects to compare.
      n: total frames sampled evenly from the full video.
      start: first sampled-frame index to evaluate, starting from 0.
      end: last sampled-frame index to evaluate, inclusive.

    Example return format; values below are illustrative:
      {
        "depth_convention": "larger_is_closer",
        "frames": [
          {
            "frame_index": 0,
            "objects": [
              {"label": "person", "status": "ok", "depth_median": 0.72},
              {"label": "apple", "status": "ok", "depth_median": 0.43}
            ]
          }
        ]
      }

    depth_median is the median relative-depth value within the localized object region. Values are not distances in meters. Interpret them according to the returned depth_convention:
      - "larger_is_closer": the object with the larger value is closer to the camera.
      - "smaller_is_closer": the object with the smaller value is closer to the camera.

    Compare the two objects only within the same frame. Do NOT compare absolute depth values across independently estimated frames.

    If an object cannot be localized, the tool returns status="not_found" and depth_median=null. If multiple instances match a label and cannot be distinguished, it returns status="ambiguous" and depth_median=null. Use a more specific label before retrying.

    Use this tool for camera-relative front/behind relationships or relative distance from the camera. It does not determine left/right relationships, exact physical distances, named identity, or an object's intrinsic front/back orientation.

    Do not infer a reliable ordering when either value is null or the two values are too similar to distinguish. Missing or ambiguous results do not establish successful verification.

video_gen_text
    Generate a video from a text prompt only, without reference images or a first frame.
    Arguments: {"prompt": "detailed cinematic description"}

video_gen_single_reference
    Generate a video conditioned on ONE reference image for a specific named entity whose appearance matters.
    Arguments: {"prompt": "detailed cinematic description", "ref_images": "/path/to/one_reference.jpg"}
      ref_images: exactly one image path as a string, not comma-separated.
      Use the selected reference image for the named person, character, landmark, or branded object.

video_gen_multiple_reference
    Generate a video conditioned on reference images for MULTIPLE named entities.
    Arguments: {"prompt": "detailed cinematic description", "ref_images": "/path_A.jpg,/path_B.jpg,/path_C.jpg"}
      ref_images: 2–5 image paths, comma-separated.
      Provide one selected reference image per entity.

video_gen_first_frame
    Generate a video conditioned on a given first frame. The supplied PNG becomes frame 0, and the model animates forward from it.
    Arguments: {"prompt": "detailed cinematic description", "first_frame": "/path/to/frame.png"}
      Use this for later phases of a multi-phase sequence by supplying the previous segment's extracted final frame.

video_gen_simulation
    Generate a video guided by simulation optical flow.
    Arguments: {"prompt": "detailed cinematic description", "src_video": "/path/to/flow.mp4"}
      src_video: the optical-flow video returned by simulation, NOT its raw rendered video.

[Reference Workflows]

The following workflows provide reference patterns for common prompt types. When a task requires multiple capabilities, combine relevant steps from these workflows and integrate their outputs as needed. Adapt tool selection and execution order to the prompt and intermediate observations, while respecting each tool's input constraints. Execute tool calls sequentially, with only one tool call per turn.

- Prompt describes a real-world process, mechanism, or scientific phenomenon, such as "osmosis", "steel forging", or "lightning formation":
  search for factual details → incorporate the retrieved information into the generation prompt → video_gen_text.
  Object detection cannot verify whether the specialized process or phenomenon is depicted correctly.

- Prompt involves ONE named entity whose appearance matters, such as a person, character, landmark, or branded object:
  image_search → visually examine the candidates in the next <think> block → select ONE IMG_K and explain the selection briefly → video_gen_single_reference using the selected image path.
  Generic object detection does not verify the named entity's identity. If the user describes a signature skill, retain its detailed movements, props, and effects in the generation prompt rather than relying on the skill name alone.

- Prompt involves MULTIPLE named entities that must co-appear:
  image_search separately for each entity → after each search, select ONE reference image for that entity → video_gen_multiple_reference with the selected image paths.
  Provide one reference per entity, with 2–5 paths in total. Generic object detection does not verify their identities. Preserve the explicit action description for each entity, including any explained signature skill.

- Prompt involves physical interactions supported by the simulator, restricted to collisions and falling/dropping:
  simulation with parameters matching the prompt → video_gen_simulation with the returned optical-flow path.
  Object detection and relative-depth estimation do not establish physical correctness.

- Prompt involves multiple subjects, including specified camera-relative spatial relationships:
  Generate a candidate using the appropriate generation tool → grounding_dino_video to verify subject presence → if the prompt specifies front/behind relationships or relative distance from the camera, call depth_anything_video with the relevant pair of subject labels.
  Read the "Per-object breakdown" and compare median depth values within the same frame according to the returned depth_convention.
  If subjects must appear together, check their presence in the same frames. If a spatial relationship applies to a particular interval, verify it within that interval.
  If a required subject is missing or the depth ordering clearly contradicts the prompt, revise the generation prompt to emphasize the missing subject or intended spatial arrangement and regenerate. Re-run the relevant verification tools.
  If localization fails, refine the labels and retry. Missing or ambiguous results do not establish successful verification.
  Use the verification observations to decide whether to regenerate or finish within the interaction budget. On finishing, submit the latest successfully generated candidate and note any unresolved verification issue in the final <think> block.

- For a multi-shot prompt, first read the number of shots K and the total duration T. Each requested shot has the same duration d = T / K, with 0 < d <= 5 seconds. Parse adjacent intervals such as 0-4s, 4-8s, and 8-12s, or an explicit request for equal-length shots. Preserve the corresponding description for each shot. Generate separate segments in that order; do not treat T as the length of each individual segment. If backend duration adaptation is required, preserve equal segment lengths of at most 5 seconds and keep the timing specification consistent for generation and evaluation.

- Prompt describes a 2-phase time-ordered sequence:
  Generate two separate video segments with visual continuity, using the final frame of the first segment to condition the second. A reference workflow for generic subjects is:
    1. video_gen_text(phase_1_prompt)
    2. extract_last_frame(video_path=<segment_1_mp4>)
    3. video_gen_first_frame(prompt=phase_2_prompt, first_frame=<segment_1_lastframe>)
    4. grounding_dino_video on segment 1 when its subjects are suitable for detection.
    5. grounding_dino_video on segment 2 when its subjects are suitable for detection.

  Each phase prompt should describe a self-contained continuous moment. Do not include words such as "then", "next", or "first half" within an individual phase prompt. Preserve subject appearance, environment, and visual style across the segment boundary.

  If the prompt also requires external knowledge, identity references, or depth verification, incorporate the relevant steps from the other reference workflows. Use a compatible generation tool for the first segment and continue subsequent segments through extracted final frames.

  Grounding retry guidance:
    - If a required phase-1 subject is not detected, regenerate phase 1 using a revised prompt. Extract its new final frame and regenerate phase 2 from that frame, then repeat applicable verification.
    - If a required phase-2 subject is not detected, regenerate only phase 2 using the same first frame and a revised prompt, then repeat applicable verification.
    - Use the observations to decide whether further regeneration is needed within the interaction budget. On finishing, retain the latest complete sequence with consistent segment dependencies.

- Prompt describes a longer time-ordered sequence of N phases, with N >= 3:
  Generate N separate video segments. Use the appropriate generation tool for the first phase, then chain later phases through first-frame conditioning. A reference workflow for N = 3 is:
    1. video_gen_text(phase_1_prompt)
    2. extract_last_frame(video_path=<segment_1_mp4>)
    3. video_gen_first_frame(prompt=phase_2_prompt, first_frame=<segment_1_lastframe>)
    4. extract_last_frame(video_path=<segment_2_mp4>)
    5. video_gen_first_frame(prompt=phase_3_prompt, first_frame=<segment_2_lastframe>)
    6. Apply suitable verification tools separately to the relevant segments.

  For N >= 4, continue extracting the final frame of segment i and using it as the first frame of segment i+1.
  Preserve subject appearance, environment, and visual style across phases.
  When regenerating a segment changes its final frame, regenerate subsequent dependent segments to preserve continuity.
  Incorporate retrieval, reference conditioning, simulation, or depth verification when required by the prompt and supported by the selected tools.
\end{promptbox}

\subsection{Custom Function: Rigid-Body Collision Simulator}
\label{app:custom_fn}

The \texttt{Code} tool in the \emph{Augmentation} tier of
Table~\ref{tab:tool_impl} is a small Python physics library
(\texttt{collision\_sim} and \texttt{freefall\_sim}) that the
agent invokes for \emph{\taskfour{}} prompts. Actual training and VABench use only collision and falling/dropping tasks; fluid simulation is outside the reported setup. Given a \texttt{<simulation>}
JSON spec produced by the policy --- mode, body shapes, masses, initial
positions, velocities, angular velocities, restitution, and an optional
target post-collision velocity --- the simulator renders a deterministic
$832{\times}480$, 5\,s, 24\,fps motion-only video that is then fed to a
motion flow-conditioned generator (Wan~2.1-VACE) as a kinematic guide.

Below we list the core collision routines of \texttt{collision\_sim.py}: the
\texttt{RigidBody} state container, a circle-vs-convex-polygon detector, the
impulse-based response resolver, and the main forward-Euler loop. The
omitted parts are bookkeeping (frame sampling and OpenCV rendering) plus
shape-specific size defaults. The listing provides the core routines; it omits the supporting modules and rendering code.

\begin{lstlisting}[style=pystyle,caption={Core of \texttt{collision\_sim.py}: planar rigid-body motion of spheres, cubes, cuboids, and triangular prisms, cross-section-based contact detection, impulse-based response, and time stepping.},label={lst:collision}]
class RigidBody:
    def __init__(self, mass, shape, size, position, velocity,
                 angle, omega, color_bgr, size2=None):
        self.mass, self.shape, self.size = mass, shape, size
        self.size2 = size2 if size2 is not None else size
        self.pos = np.array(position, dtype=np.float64)
        self.vel = np.array(velocity, dtype=np.float64)
        self.angle, self.omega = float(angle), float(omega)

        # Uniform solid bodies; motion is restricted to a plane.
        # Inertia is about the centroidal axis normal to that plane.
        # size: sphere radius, cube/cuboid half-width, or
        # equilateral triangular cross-section circumradius.
        # size2: cuboid half-height.
        if shape == 'sphere':
            self.I = (2/5) * mass * size**2
        elif shape == 'cube':
            self.I = (1/6) * mass * (2*size)**2
        elif shape == 'cuboid':
            self.I = (1/12) * mass * (
                (2*size)**2 + (2*self.size2)**2
            )
        elif shape == 'triangular_prism':
            self.I = (1/12) * mass * (size*np.sqrt(3))**2
        else:
            raise ValueError(f"Unsupported shape: {shape}")


# ----- Contact detection using planar cross-sections -----
# Sphere: circular cross-section.
# Cube/cuboid/triangular prism: convex polygonal cross-section.
def detect_collision(sphere, obj):
    corners = obj.corners_world()
    p = sphere.pos

    if _point_in_convex_polygon(p, corners):
        # Deep penetration: find the nearest exit edge.
        # Determine vertex winding to orient normals outward.
        signed_area2 = sum(
            corners[i][0] * corners[(i+1) % len(corners)][1]
            - corners[(i+1) % len(corners)][0] * corners[i][1]
            for i in range(len(corners))
        )

        best_dist = float('inf')
        best_n = None
        for i in range(len(corners)):
            a = corners[i]
            b = corners[(i+1) % len(corners)]
            edge = b - a
            length = np.linalg.norm(edge)
            if length < 1e-12:
                continue

            outward = np.array([edge[1], -edge[0]]) / length
            if signed_area2 < 0:
                outward = -outward

            # Nonnegative distance from the interior point to edge.
            dist = np.dot(a - p, outward)
            if dist < best_dist:
                best_dist, best_n = dist, outward

        if best_n is None:
            raise ValueError("Degenerate polygonal cross-section")

        best_dist = max(best_dist, 0.0)
        contact = p + best_n * best_dist
        return True, contact, best_n, sphere.size + best_dist

    # Center outside the polygon: find the closest boundary point.
    min_dist, closest = float('inf'), None
    for i in range(len(corners)):
        c = _closest_on_segment(
            p, corners[i], corners[(i+1) % len(corners)]
        )
        d = np.linalg.norm(p - c)
        if d < min_dist:
            min_dist, closest = d, c

    if min_dist >= sphere.size:
        return False, None, None, 0.0

    diff = p - closest
    normal = diff / max(np.linalg.norm(diff), 1e-10)
    return True, closest, normal, sphere.size - min_dist


# ----- Impulse-based collision response -----
def resolve_collision_impulse(a, b, contact, normal, e=1.0):
    """Apply collision impulse; normal points from b to a."""
    r_a, r_b = contact - a.pos, contact - b.pos
    v_a = a.vel + a.omega * np.array([-r_a[1], r_a[0]])
    v_b = b.vel + b.omega * np.array([-r_b[1], r_b[0]])
    v_rel_n = np.dot(v_a - v_b, normal)
    if v_rel_n > 0:  # Separating; no impulse needed.
        return

    ra_x = r_a[0] * normal[1] - r_a[1] * normal[0]
    rb_x = r_b[0] * normal[1] - r_b[1] * normal[0]
    denom = (
        1/a.mass + 1/b.mass
        + ra_x**2 / a.I + rb_x**2 / b.I
    )
    j = -(1 + e) * v_rel_n / denom
    imp = j * normal

    a.vel += imp / a.mass
    b.vel -= imp / b.mass
    a.omega += (r_a[0]*imp[1] - r_a[1]*imp[0]) / a.I
    b.omega -= (r_b[0]*imp[1] - r_b[1]*imp[0]) / b.I


# ----- Impulse update followed by planar integration -----
def simulate(args):
    body_a, body_b = _make_bodies(args)
    e_eff = args.restitution
    hist_a, hist_b = [], []
    n_steps = int(TOTAL_TIME / DT)
    sample = max(1, round((1.0 / FPS) / DT))

    for step in range(n_steps):
        if step % sample == 0:
            hist_a.append(_snapshot(body_a))
            hist_b.append(_snapshot(body_b))

        hit, contact, normal, pen = detect_any(body_a, body_b)
        if hit:
            # Apply impulse at the detected contact geometry.
            resolve_collision_impulse(
                body_a, body_b, contact, normal, e=e_eff
            )

            # Correct overlap using inverse-mass weighting.
            inv_a, inv_b = 1/body_a.mass, 1/body_b.mass
            correction = normal * (pen + 2e-3)
            body_a.pos += correction * inv_a / (inv_a + inv_b)
            body_b.pos -= correction * inv_b / (inv_a + inv_b)

        # Integrate using post-impact velocities.
        body_a.pos += body_a.vel * DT
        body_a.angle += body_a.omega * DT
        body_b.pos += body_b.vel * DT
        body_b.angle += body_b.omega * DT

    return hist_a, hist_b, body_a, body_b
\end{lstlisting}

The routine dispatches \texttt{detect\_any} to \texttt{detect\_circle\_circle},
\texttt{detect\_collision} (circle--polygon), or \texttt{detect\_polygon\_polygon}
(SAT) depending on the two body shapes. Restitution $e$ is either taken from
the spec or, when the policy supplies a target post-collision velocity for
one body, derived once at first contact from momentum conservation along
the contact normal and held fixed for subsequent collisions. The same
\texttt{RigidBody} integrator is reused with body-specific external forces
in \texttt{freefall\_sim} with gravity for falling/dropping tasks.

\clearpage
\section{Reward and RL Implementation Details}
\label{app:rl_details}

This section provides implementation details for the agentic RL training procedure introduced in Section~\ref{sec:grpo}, including the hybrid reward in Eq.~\eqref{eq:reward}, task advantage normalization, the token-level policy objective, and tool-failure handling.

\subsection{Format Reward}

The format reward evaluates whether the agent follows the required interaction protocol. Let $\mathcal{U}(\tau)$ denote the set of agent-emitted actions in trajectory $\tau$. We compute
\begin{equation}
R_{\mathrm{format}}(\tau)
=
\frac{1}{|\mathcal{U}(\tau)|}
\sum_{a_t\in\mathcal{U}(\tau)}
\mathbb{I}_{\mathrm{format}}(a_t),
\label{eq:app_format_reward}
\end{equation}
where $\mathbb{I}_{\mathrm{format}}(a_t)=1$ if action $a_t$ satisfies the required output format and $0$ otherwise.

For a tool-call action, the format check requires that:

\begin{itemize}
    \item the response contains a valid \texttt{<think>} block followed by exactly one \texttt{<tool\_call>} block;
    \item the content of the tool-call block can be parsed as JSON;
    \item the specified tool belongs to the available tool set; and
    \item all required arguments are present and have valid types.
\end{itemize}

For a final-answer action, the response must contain a valid \texttt{<think>} block followed by the required termination tag. Malformed actions receive a format score of zero for the corresponding turn.

\subsection{Tool-Use Reward}

The tool-use reward evaluates whether the agent selects tools appropriate for the task and uses the outputs returned by those tools. We define
\begin{equation}
R_{\mathrm{tool}}(\tau,c)
=
\frac{
\sum_{m=1}^{M(\tau,c)}q_m(\tau,c)
}{
M(\tau,c)
},
\label{eq:app_tool_reward}
\end{equation}
where $M(\tau,c)$ is the number of applicable tool-use checks for trajectory $\tau$ from category $c$, and $q_m(\tau,c)\in\{0,1\}$ indicates whether the $m$-th check is satisfied. When no check is applicable, we set $R_{\mathrm{tool}}(\tau,c)=0$.

Table~\ref{tab:tool_reward_rules} summarizes the category-specific checks. These checks evaluate the functional use of tool outputs rather than merely counting tool calls.

\begin{table}[H]
\centering
\caption{Category-specific checks for the tool-use reward, following the typical workflows in Table~\ref{tab:workflow}. Each check scores one if satisfied and zero otherwise.}
\label{tab:tool_reward_rules}
\resizebox{\linewidth}{!}{
\begin{tabular}{@{}lll@{}}
\toprule
Category & Expected tool use & Output-utilization check \\
\midrule
\taskone{}
& Search-Text $\rightarrow$ T2V
& Retrieved procedural evidence is incorporated into the T2V prompt \\
\tasktwo{}
& Search-Image $\rightarrow$ R2V or I2V
& A selected retrieved image is passed to R2V or I2V \\
\taskthree{}
& Search-Image$^{\times N}$ $\rightarrow$ R2V
& Retrieved references for the named subjects are jointly passed to R2V \\
\taskfour{}
& Code $\rightarrow$ M2V
& The returned motion or optical-flow condition is passed to M2V \\
\taskfive{}
& T2V $\rightarrow$ (Obj Det., Depth Est.) $\rightarrow$ T2V
& Verification feedback informs acceptance or regeneration when needed \\
\tasksix{}
& T2V $\rightarrow$ (ELF $\rightarrow$ I2V)$^{\times(K-1)}$
& Each extracted final frame conditions the next segment through I2V \\
\bottomrule
\end{tabular}
}
\end{table}

For retrieval-based tasks, the utilization check is performed against the evidence or reference identifier returned by the retrieval tool. For simulation and multi-phase generation, it is performed by matching the returned artifact path with the conditioning input of the subsequent generation call. For verification-based refinement, the check determines whether the verification result is followed by an appropriate acceptance or regeneration action.

\subsection{Category-Specific VLM Reward}
\label{app:vlm_reward}

The VLM reward evaluates the final generated video using a category-specific rubric. The judge receives:

\begin{enumerate}
    \item the original user prompt;
    \item the final generated video (for \tasksix{}, equal-duration shot clips paired with their corresponding shot prompts); and
    \item category-dependent reference information, when applicable.
\end{enumerate}

For each category $c$, we obtain scores $\{s_d\}_{d\in\mathcal{D}_c}$ over a set of evaluation dimensions $\mathcal{D}_c$. The VLM judge provides all scores except \texttt{motion\_correctness} for Physics Simulation, which is a binary score computed by code using a $10\%$ relative-error tolerance (Appendix~\ref{app:physics_checks}) and included with weight $0.20$. Each raw dimension score is mapped to $[0,1]$ and combined using category-specific weights:
\begin{equation}
R_{\mathrm{vlm}}(\tau,x,c)
=
\frac{
\sum_{d\in\mathcal{D}_c}
w_{c,d}\,
\operatorname{Norm}_{c,d}(s_d)
}{
\sum_{d\in\mathcal{D}_c}w_{c,d}
}.
\label{eq:app_vlm_reward}
\end{equation}

Here, $w_{c,d}$ is the weight assigned to dimension $d$, and $\operatorname{Norm}_{c,d}$ maps its raw score to $[0,1]$. Integer scores on a $1$--$10$ scale are divided by $10$, while dimensions scored in $\{0,0.5,1\}$ and the binary motion score in $\{0,1\}$ are used directly. Malformed judge responses or responses missing a required VLM-scored dimension are assigned a reward of zero.

The task-specific reference information and auxiliary assessment procedures are described in Appendix~\ref{app:evaluation_references}. The complete judge prompts, evaluation dimensions, raw score domains, and aggregation weights for all six categories are provided in Appendix~\ref{app:reward_prompt}.

\subsection{Hybrid Reward}

The final trajectory reward is
\begin{equation}
R(\tau,x,c)
=
0.1R_{\mathrm{format}}(\tau)
+
0.5R_{\mathrm{vlm}}(\tau,x,c)
+
0.4R_{\mathrm{tool}}(\tau,c).
\label{eq:app_hybrid_reward}
\end{equation}

All components lie in $[0,1]$, so the resulting hybrid reward also lies in $[0,1]$.

\subsection{Token-Level Group-Relative Advantage}

For prompt $x_{c,s}$, let
\[
\left\{
\tau_{c,s,g}
\right\}_{g=1}^{G_{c,s}}
\]
denote the group of sampled trajectories. Rollouts discarded because of external tool failures are resampled until $G_{c,s}=6$ usable trajectories are available for the prompt. The group mean and standard deviation below are computed after replenishment. We first compute a standard group-relative advantage:
\begin{equation}
\hat{A}_{c,s,g}
=
\frac{
R_{c,s,g}-\mu_{c,s}^{\mathrm{grp}}
}{
\sigma_{c,s}^{\mathrm{grp}}+\epsilon
},
\label{eq:app_group_advantage}
\end{equation}
where
\begin{equation}
R_{c,s,g}
=
R(\tau_{c,s,g},x_{c,s},c),
\end{equation}
\begin{equation}
\mu_{c,s}^{\mathrm{grp}}
=
\frac{1}{G_{c,s}}
\sum_{g'=1}^{G_{c,s}}R_{c,s,g'},
\end{equation}
and
\begin{equation}
\sigma_{c,s}^{\mathrm{grp}}
=
\operatorname{std}
\left(
\left\{
R_{c,s,g'}
\right\}_{g'=1}^{G_{c,s}}
\right).
\end{equation}

The same group-relative advantage $\hat{A}_{c,s,g}$ is initially assigned to every agent-emitted token in trajectory $\tau_{c,s,g}$.

\subsection{Task-Level Advantage Normalization}

Following AgentRL~\citep{agenticrl}, we further normalize token-level advantages separately for each task category. Let a high-level action at interaction step $t$ contain tokens
\[
a_{c,s,g,t}
=
\left\{
y_{c,s,g,t,k}
\right\}_{k=1}^{L_{c,s,g,t}}.
\]
For every agent-emitted token, we initially assign
\begin{equation}
\hat{A}_{c,s,g,t,k}
=
\hat{A}_{c,s,g}.
\end{equation}

We collect the advantages of all agent-emitted tokens from category $c$ in the current training batch:
\begin{equation}
\mathcal{A}_{c}^{\mathrm{tok}}
=
\left\{
\hat{A}_{c,s,g,t,k}
\;\middle|\;
y_{c,s,g,t,k}
\text{ is agent-emitted and occurs in the current batch}
\right\}.
\label{eq:app_task_advantage_set}
\end{equation}

The task-specific statistics are
\begin{equation}
\mu_c^{\mathrm{task}}
=
\operatorname{mean}
\left(
\mathcal{A}_{c}^{\mathrm{tok}}
\right),
\qquad
\sigma_c^{\mathrm{task}}
=
\operatorname{std}
\left(
\mathcal{A}_{c}^{\mathrm{tok}}
\right).
\end{equation}

The final advantage used for policy optimization is
\begin{equation}
\widetilde{A}_{c,s,g,t,k}
=
\frac{
\hat{A}_{c,s,g,t,k}
-
\mu_c^{\mathrm{task}}
}{
\sigma_c^{\mathrm{task}}+\epsilon
}.
\label{eq:app_task_normalized_advantage}
\end{equation}

Thus, group-relative normalization compares trajectories sampled for the same prompt, whereas task advantage normalization equalizes the advantage distributions of different task categories within the current training batch.

\subsection{Token-Level Policy Objective}

For an agent-emitted token $y_{c,s,g,t,k}$ with preceding interaction history $H_{c,s,g,t,k}$, the importance ratio is
\begin{equation}
\rho_{c,s,g,t,k}(\theta)
=
\frac{
\pi_\theta
\left(
y_{c,s,g,t,k}
\mid
H_{c,s,g,t,k}
\right)
}{
\pi_{\theta_{\mathrm{old}}}
\left(
y_{c,s,g,t,k}
\mid
H_{c,s,g,t,k}
\right)
},
\label{eq:app_token_ratio}
\end{equation}
where $\pi_{\theta_{\mathrm{old}}}$ is the behavior policy used to generate the rollout.

Let $\mathcal{M}_{c,s,g}$ denote the set of agent-emitted token positions in trajectory $\tau_{c,s,g}$. The clipped policy loss is
\begin{align}
\mathcal{L}_{\mathrm{policy}}(\theta)
=
-&
\frac{1}{
\sum_{c,s,g}|\mathcal{M}_{c,s,g}|
}
\sum_{c,s,g}
\sum_{(t,k)\in\mathcal{M}_{c,s,g}}
\min
\left(
\rho_{c,s,g,t,k}(\theta)
\widetilde{A}_{c,s,g,t,k},
\right.
\nonumber\\[-1mm]
&
\hspace{25mm}
\left.
\operatorname{clip}
\left(
\rho_{c,s,g,t,k}(\theta),
1-\epsilon_{\mathrm{lo}},
1+\epsilon_{\mathrm{hi}}
\right)
\widetilde{A}_{c,s,g,t,k}
\right).
\label{eq:app_policy_loss}
\end{align}

For an optional direct actor KL penalty, the optimization objective is
\begin{equation}
\mathcal{L}_{\mathrm{GRPO}}(\theta)
=
\mathcal{L}_{\mathrm{policy}}(\theta)
+
\beta_{\mathrm{KL}}\mathcal{L}_{\mathrm{KL}},
\end{equation}
where
\begin{equation}
\mathcal{L}_{\mathrm{KL}}
=
\mathbb{E}_{c,s,g,(t,k)\in\mathcal{M}_{c,s,g}}
\left[
D_{\mathrm{KL}}
\left(
\pi_\theta(\cdot\mid H_{c,s,g,t,k})
\,\Vert\,
\pi_{\mathrm{ref}}(\cdot\mid H_{c,s,g,t,k})
\right)
\right].
\label{eq:app_kl_loss}
\end{equation}

The direct actor KL-loss coefficient is $\beta_{\mathrm{KL}}=0$, so the direct objective is $\mathcal{L}_{\mathrm{policy}}$. The reference policy is fixed to the initial SFT checkpoint,
\[
\pi_{\mathrm{ref}}=\pi_{\theta_0},
\]
whereas $\pi_{\theta_{\mathrm{old}}}$ is the behavior policy used for rollout collection and is updated during RL optimization. We use
\[
\epsilon_{\mathrm{lo}}=0.2,
\qquad
\epsilon_{\mathrm{hi}}=0.28,
\]
for asymmetric clipping.

\subsection{Training Configuration}
\label{app:rl_config}
Table~\ref{tab:rl_config} summarizes the RL launch settings. The training and evaluation interaction budgets are 10 and 20, respectively. The rollout group size of six follows the reported experiment setting and the replenishment procedure described above. Observation tokens are retained as context and masked from the policy loss.

\begin{table}[H]
\centering
\caption{RL configuration. Length limits are in tokens. The direct actor KL-loss coefficient and the separately configured KL-controller coefficient are distinct settings.}
\label{tab:rl_config}
\small
\begin{tabular}{@{}ll@{}}
\toprule
Setting & Value \\
\midrule
Learning rate / warmup steps & $10^{-6}$ / 10 \\
Epochs & 1 \\
Prompt batch / PPO mini-batch & 10 / 10 \\
Usable rollouts per prompt & 6 (replenished after tool failures) \\
Category sampling & Stratified \\
Temperature / top-$p$ / top-$k$ & 1.0 / 1.0 / disabled \\
Prompt / response length limits & 4096 / 4096 \\
Action / observation length limits & 2048 / 3000 \\
Training / evaluation interaction limits & 10 / 20 \\
PPO / log-probability micro-batch per GPU & 1 / 1 \\
Clipping $(\epsilon_{\mathrm{lo}},\epsilon_{\mathrm{hi}})$ & $(0.2,0.28)$ \\
Direct actor KL-loss coefficient & 0 \\
KL-controller coefficient & $10^{-3}$ \\
Entropy coefficient & 0 \\
Policy sharding / rollout engine & FSDP / asynchronous vLLM \\
Tensor / sequence parallel size & 1 / 1 \\
vLLM GPU-memory utilization & 0.55 \\
Gradient checkpointing & Enabled \\
Parameter / optimizer offload & Disabled / disabled \\
\bottomrule
\end{tabular}
\end{table}
The launch script sets the direct actor KL-loss coefficient to zero and separately configures the KL controller to $10^{-3}$. The controller setting is not a nonzero direct KL-loss term in Eq.~\eqref{eq:grpo}.

\clearpage
\section{Evaluation Protocol and Judge Prompts}
\label{app:reward_prompt}
\label{app:evaluation}

\subsection{Scoring Criteria and Aggregation}
\label{app:scoring}

We use Gemini~3.1~Pro to score generated videos against category-specific rubrics. The judge receives the user prompt, optional textual or image references, and the generated video, then returns per-dimension scores in JSON format. For \tasksix{}, its video input is instead an ordered list of equal-duration clips, pre-segmented according to the prompt and paired with the corresponding shot descriptions (Appendix~\ref{app:duration_handling}). All categories share a common prompt structure, with different evaluation criteria, scoring scales, and category-specific instructions.

For VLM-scored dimensions, we use two scoring scales. \texttt{discrete\_3} assigns scores in $\{0,0.5,1\}$ to criteria such as aesthetics and action correctness. \texttt{int\_1\_10} assigns integer scores from $1$ to $10$ to factual correctness and identity preservation, providing finer distinctions within these criteria. Integer scores are divided by $10$, while discrete scores are used directly. The code-based \texttt{motion\_correctness} criterion instead uses a binary score in $\{0,1\}$, with the $10\%$ relative-error tolerance described in Appendix~\ref{app:physics_checks}.

Table~\ref{tab:reward_dims} lists the criteria and weights, which sum to one within each category. For \taskfour{}, code-based quantitative checks supplement the VLM scores with weight $0.20$. Using normalized criterion scores $s_d$, we compute
\[
R_{\mathrm{vlm}}=\sum_d w_d s_d.
\]
During RL, judge responses that cannot be parsed as valid JSON or omit a required score receive a reward of zero. During benchmark evaluation, an invalid judge response triggers a retry of the judge call on the same video and references, with at most three attempts in total; a response that remains invalid receives zero. Generation/execution failures are handled separately, as described below.

\begin{table}[H]
\centering
\caption{Category-specific scoring criteria and weights for $R_{\mathrm{vlm}}$. VLM-scored criteria use the field names in the judge prompts; \taskfour{} additionally includes code-based quantitative checks with weight $0.20$. Weights sum to one within each category, and the reward is the weighted sum of normalized scores. Integer scores are divided by $10$, while scores in $\{0,0.5,1\}$ and the binary motion score are used directly. For \taskthree{}, \texttt{identity\_match} assesses both identity preservation and whether all named subjects appear together in at least one frame.}
\label{tab:reward_dims}
\small
\setlength{\tabcolsep}{5pt}
\begin{tabular}{@{}llcc@{}}
\toprule
Category & Evaluation criterion & Raw score domain & Weight \\
\midrule
\multirow{2}{*}{\taskone{}}
  & \texttt{factual\_correctness}
  & $\{1,\ldots,10\}$ & $0.90$ \\
  & \texttt{aesthetics}
  & $\{0,0.5,1\}$ & $0.10$ \\
\midrule
\multirow{3}{*}{\tasktwo{}}
  & \texttt{identity\_match}
  & $\{1,\ldots,10\}$ & $0.70$ \\
  & \texttt{action\_correctness}
  & $\{0,0.5,1\}$ & $0.20$ \\
  & \texttt{aesthetics}
  & $\{0,0.5,1\}$ & $0.10$ \\
\midrule
\multirow{3}{*}{\taskthree{}}
  & \texttt{identity\_match}
  & $\{1,\ldots,10\}$ & $0.70$ \\
  & \texttt{action\_correctness}
  & $\{0,0.5,1\}$ & $0.20$ \\
  & \texttt{aesthetics}
  & $\{0,0.5,1\}$ & $0.10$ \\
\midrule
\multirow{3}{*}{\taskfour{}}
  & \texttt{physical\_realism}
  & $\{0,0.5,1\}$ & $0.70$ \\
  & \texttt{motion\_correctness (by code)}
  & $\{0,1\}$ & $0.20$ \\
  & \texttt{aesthetics}
  & $\{0,0.5,1\}$ & $0.10$ \\
\midrule
\multirow{3}{*}{\taskfive{}}
  & \texttt{subjects\_correctness}
  & $\{0,0.5,1\}$ & $0.50$ \\
  & \texttt{interaction\_logic}
  & $\{0,0.5,1\}$ & $0.40$ \\
  & \texttt{aesthetics}
  & $\{0,0.5,1\}$ & $0.10$ \\
\midrule
\multirow{3}{*}{\tasksix{}}
  & \texttt{phase\_transition}
  & $\{0,0.5,1\}$ & $0.50$ \\
  & \texttt{intraphase\_correctness}
  & $\{0,0.5,1\}$ & $0.40$ \\
  & \texttt{aesthetics}
  & $\{0,0.5,1\}$ & $0.10$ \\
\bottomrule
\end{tabular}
\end{table}

\FloatBarrier
\subsection{Reference Information and Task-Specific Assessment}
\label{app:evaluation_references}

The category-specific scoring procedure uses the following reference information and auxiliary checks. For benchmark evaluation, each prompt has fixed textual or image references, where applicable, shared across all compared methods and both toolsets. These references are collected independently of each evaluated agent's retrieval outputs using Claude Web Search. The agents' Search-Text and Search-Image tools use Bing Search in both Toolset~1 and Toolset~2; their retrieval outputs do not define the evaluation ground truth.

\paragraph{Procedural knowledge (\taskone{}).}
We use Claude Web Search to retrieve key factual information about the target technique, process, or phenomenon. The retrieved content is manually checked for relevance to the requested process and for whether its characteristic steps or outcomes can be demonstrated visually. These checked knowledge items provide the factual reference for assessing the generated video.

\paragraph{Single- and multi-entity identity (\tasktwo{} and \taskthree{}).}
We use independently collected image references obtained through Claude Web Search for identity assessment, retaining the top three image results for each named character or entity separately (three per entity, not three per multi-entity prompt). The reference set for each evaluation prompt is fixed across methods. The judge compares the subjects depicted in the generated video with these reference images using the category-specific identity rubric.

\paragraph{Physical dynamics (\taskfour{}).}
Code-based checks use SAM~3 object tracks to assess image-plane velocity and acceleration. They compare mass-weighted velocities before and after collisions under a fixed-camera assumption and assess approximately constant, nonzero acceleration for free fall. These checks supplement the VLM's assessment of physical plausibility, as detailed in Appendix~\ref{app:physics_checks} and Table~\ref{tab:reward_dims}.

\paragraph{Compositional scenes (\taskfive{}).}
The VLM directly evaluates the video against the task prompt and the category-specific rubric, without additional external references. It checks the requested subjects, their interactions, and spatial relationships.

\paragraph{Multi-shot temporal structure (\tasksix{}).}
We divide the final video into the number of equal-duration shots specified by the prompt, using its prescribed boundaries. The judge receives the resulting ordered video clips together with their corresponding shot prompts. It evaluates clip $k$ against shot prompt $k$, then checks the requested ordering and continuity across adjacent clips. Content appearing only in another clip does not satisfy the requirements of shot $k$. The same preprocessing and rubric apply to standalone generators and agents, during both RL reward evaluation and benchmark evaluation.

\subsection{Duration Handling for \tasksix{}}
\label{app:duration_handling}
Each \tasksix{} prompt first specifies a shot count $K\in\{2,3\}$ and a common shot duration $d$, with $0<d\leq5\,\mathrm{s}$, giving a total duration $T=Kd$. The prompt either gives consecutive intervals or explicitly requests $K$ equal-length shots with an ordered description of each. For example, a 12-second, three-shot prompt assigns its events to $0$--$4\,\mathrm{s}$, $4$--$8\,\mathrm{s}$, and $8$--$12\,\mathrm{s}$. These intervals are adjacent: they contain no gaps or overlaps.

For evaluation, we split the final video at $t_k=kd$, $k=0,\ldots,K$, before sending it to the VLM. The evaluator receives the ordered pairs $(x_k,v_k)$, where $x_k$ is the description assigned to shot $k$ (including relevant shared subject and setting constraints), and $v_k$ is the clip from $[(k-1)d,kd)$; the last interval includes the final endpoint. A prompt phrased as ``three equal-length shots'' is processed identically to one giving explicit timestamps. The judge does not infer new boundaries from the generated content or move a correctly depicted event into a different shot. It checks every clip against its paired description and uses the category-level dimensions and weights in Table~\ref{tab:reward_dims}. Any backend duration adaptation must retain equal shot lengths of at most five seconds and provide consistent updated timing in both the generation request and the shot prompts used for evaluation.

\subsection{Execution Failures and Retry Limits}
\label{app:evaluation_retries}
Benchmark retries target the failed operation rather than restarting successful earlier stages: an external tool or generation call is attempted at most three times in total (the initial attempt and at most two retries), using the same request. These technical retries are distinct from an agent's decision to revise a prompt and generate a new candidate. If the interaction budget is exhausted without a complete video, the evaluation trajectory is retried on the same prompt, again with at most three trajectory attempts in total. A complete latest candidate is retained when available. A generation case still lacking a complete video after this limit is flagged as unresolved, not silently removed or assigned a video-quality score. No generation cases remained unresolved after the permitted retries in the reported benchmark runs; all 600 prompts were included in the benchmark means. Judge transport or parsing failures are retried at the judge-call level on the same completed video, with the three-attempt limit above; they do not trigger video regeneration.

\subsection{Physical Plausibility and Quantitative Checks}
\label{app:physics_checks}
The actual \taskfour{} SFT, RL, and VABench data contain only quantitative collision and falling/dropping prompts; qualitative, non-numerical prompts and fluid tasks are not included. Collision prompts specify both masses and initial velocities; drop prompts specify release height and initial motion. For these two task types, the VLM assesses the plausibility of the depicted physical interactions, including contact, rebound, motion direction, and induced rotation. The separate code-based \texttt{motion\_correctness} component uses SAM~3 to track objects in the generated video and estimates their velocities and accelerations in image coordinates. Because camera geometry and the conversion from image displacement to physical distance are not calibrated, these measurements test image-plane motion consistency rather than recover exact three-dimensional velocities or accelerations.

\paragraph{Collision and momentum consistency.}
Under a fixed-camera assumption, we estimate each object's pre-collision velocity from the first $0.5\,\mathrm{s}$ of the clip and its post-collision velocity from the final $0.5\,\mathrm{s}$. For two colliding objects, we use the masses $m_1$ and $m_2$ specified in the prompt, expressed in consistent units, and compare
\[
m_1\mathbf{v}_{1,\mathrm{pre}}+m_2\mathbf{v}_{2,\mathrm{pre}}
\quad\text{and}\quad
m_1\mathbf{v}_{1,\mathrm{post}}+m_2\mathbf{v}_{2,\mathrm{post}},
\]
where each $\mathbf{v}$ is an image-plane velocity vector. This is a projected momentum-consistency check whose interpretation assumes that the initial and final windows represent motion before and after the collision. Camera motion, perspective effects, and changes in object depth can affect the comparison; it is not a calibrated test of three-dimensional momentum conservation.

\paragraph{Free-fall consistency.}
For free-fall videos, we estimate image-plane velocity from each pair of consecutive tracked frames and acceleration from the difference between consecutive velocity estimates. For uniformly spaced frame times with interval $\Delta t$ and tracked positions $\mathbf{p}_i$,
\[
\mathbf{v}_i=\frac{\mathbf{p}_{i+1}-\mathbf{p}_i}{\Delta t},
\qquad
\mathbf{a}_i=\frac{\mathbf{v}_{i+1}-\mathbf{v}_i}{\Delta t}.
\]
Thus a velocity estimate uses two frames, while an acceleration estimate uses two adjacent velocity estimates (at least three frames). During the falling interval, the acceleration must be approximately constant and nonzero; static objects and constant-velocity motion do not pass. This checks temporal consistency of image-plane acceleration, rather than requiring its magnitude to equal gravitational acceleration in physical units.

\paragraph{Binary scoring with a relative-error tolerance.}
The applicable motion check uses a $10\%$ relative-error tolerance. For collisions, \texttt{motion\_correctness} is $1$ when the relative discrepancy between pre- and post-collision image-plane momentum is at most $0.10$, and $0$ otherwise. For free fall, it is $1$ when the measured acceleration is nonzero and satisfies the constant-acceleration check within $10\%$ relative error, and $0$ otherwise. Thus the code-based score lies in $\{0,1\}$ and is used directly, without division by $10$.

The code-based component contributes $0.20$ to the total score, as reported in Table~\ref{tab:reward_dims}; the VLM's physical-plausibility assessment remains the primary criterion. The simulation coordinates described in Appendix~\ref{app:system_prompt} are used to construct generation conditions, not as a substitute for tracking the final generated video.

\subsection{Category-Specific Judge Prompts}
\label{app:judge_prompts}

\subsubsection{\taskone{}}
\label{app:judge_1}
\begin{promptbox}[Reward Prompt --- \taskone{}]
You are a strict expert evaluator for AI-generated videos.

You receive:
  (a) USER PROMPT - what the video should depict.
  (b) (optional) REFERENCE - factual description / reference images.
  (c) VIDEO - the generated video clip.

Your job is to score the video on the following dimensions:

  - "factual_correctness" (weight 0.90, scale int_1_10): Does the video factually depict the SPECIFIC named process / action / phenomenon as described in the reference? Treat REFERENCE as ground truth. Distil 3-5 specific KNOWLEDGE ITEMS from the reference (stance, hand shape, sub-motion sequence, body orientation, specific phase, ...), then check the video against each item. Give a score from 1 to 10. Anchor scale (use the FULL range - intermediate values like 4, 7, 9 are encouraged, not just the extremes): 10 = every reference specific clearly and correctly shown, no contradictions. 9 = all specifics correct, one minor detail approximate. 8 = most specifics correct, one missing or approximate, no contradictions. 7 = right subject + clearly in the right family, ~half of the named specifics visible. 6 = right subject + right family of action, but the characteristic specifics of the named term are mostly absent (e.g. basketball drive that doesn't crisply show Euro step). 5 = right subject but action is generic and does not commit to the named posture / mechanism. 4 = right subject family but action is wrong sub-type, or one or two specifics barely visible while rest are wrong. 3 = mostly wrong action though subject is recognizable. 2 = wrong action entirely; subject only loosely related. 1 = wrong subject entirely (e.g. ribbon dance instead of Tai Chi), unidentifiable action, video nearly static, or depiction directly contradicts the reference.
  - "aesthetics" (weight 0.10, scale discrete_3): Visual quality only: lighting, composition, color, sharpness - independent of factual correctness.

REFERENCE may include factual details retrieved from the web - TREAT AS GROUND TRUTH for factual_correctness.

DISCRETE 3-LEVEL - dims marked `<0 | 0.5 | 1>` use EXACTLY one of {0, 0.5, 1}:
  1   - fully satisfies the dimension's requirement
  0.5 - mostly correct, minor issues / partial mismatch
  0   - fails the key requirement of this dimension

INTEGER 1-10 - dims marked `<integer 1-10>` use an integer from 1 to 10:
  Use the FULL range. Avoid defaulting to the middle. Intermediate values (4, 6, 7, 9) MUST be used when the video is partially correct - do NOT round to 5 or 10. Follow the anchors listed in that dim's definition above.

Procedure (must follow):
  1. From the prompt, list the TOP HARD CONSTRAINTS (2-5 items).
  2. Score each dimension against those constraints + the video.
  3. Do NOT assume correctness if a detail is not clearly visible - score lower.

Respond with EXACTLY ONE JSON object on its own line, no preamble, no markdown, no commentary outside the JSON:

{
  "rationale": "<5-10 evidence-based sentences. Start by listing the hard constraints extracted from the prompt, then walk through each dimension.>",
  "factual_correctness": <integer 1-10>,
  "aesthetics": <0 | 0.5 | 1>
}
\end{promptbox}

\subsubsection{\tasktwo{}}
\label{app:judge_2}
\begin{promptbox}[Reward Prompt --- \tasktwo{}]
You are a strict expert evaluator for AI-generated videos.

You receive:
  (a) USER PROMPT - what the video should depict.
  (b) (optional) REFERENCE - factual description / reference images.
  (c) VIDEO - the generated video clip.

Your job is to score the video on the following dimensions:

  - "identity_match" (weight 0.70, scale int_1_10): Does the subject in the video LOOK like the SPECIFIC named person / character / landmark from the prompt + reference image(s)? Distil 3-5 distinctive visual features from the reference (face shape, hair, glasses, signature outfit, distinctive accessory, body type, ...) then check the video against those features. Give an integer 1-10. Anchor scale (use the FULL range - intermediate values (4, 6, 7, 9) are encouraged): 10 = every distinctive feature clearly and correctly matches the reference; the subject is unmistakably the named entity. 9 = clearly the right person, one minor feature off. 8 = clearly the right person, ~one specific detail approximate or missing. 7 = strong resemblance, more than half of distinctive features match. 6 = some resemblance, about half of distinctive features match. 5 = generic-looking stand-in with the right rough ethnicity / build, but most distinguishing features missing. 4 = generic stand-in, only one feature aligns. 3 = visibly approximate but features are off. 2 = visibly the wrong person / wrong subject. 1 = subject absent, unidentifiable, or video near static.
  - "action_correctness" (weight 0.20, scale discrete_3): Does the subject's pose / activity match the prompt's description of what they're doing? Score 1 = action clearly visible and correct. Score 0.5 = action approximately right but missing key specifics. Score 0 = different action entirely.
  - "aesthetics" (weight 0.10, scale discrete_3): Visual quality.

If the prompt includes a signature skill, assess action_correctness against its explicit description of movements, props, and effects; do not require additional unstated skill knowledge.

REFERENCE IMAGE(S) show the real appearance of the named subject. identity_match is the dominant signal - examine face, hair, glasses, outfit carefully.

DISCRETE 3-LEVEL - dims marked `<0 | 0.5 | 1>` use EXACTLY one of {0, 0.5, 1}:
  1   - fully satisfies the dimension's requirement
  0.5 - mostly correct, minor issues / partial mismatch
  0   - fails the key requirement of this dimension

INTEGER 1-10 - dims marked `<integer 1-10>` use an integer from 1 to 10:
  Use the FULL range. Avoid defaulting to the middle. Intermediate values (4, 6, 7, 9) MUST be used when the video is partially correct - do NOT round to 5 or 10. Follow the anchors listed in that dim's definition above.

Procedure (must follow):
  1. From the prompt, list the TOP HARD CONSTRAINTS (2-5 items).
  2. Score each dimension against those constraints + the video.
  3. Do NOT assume correctness if a detail is not clearly visible - score lower.

Respond with EXACTLY ONE JSON object on its own line, no preamble, no markdown, no commentary outside the JSON:

{
  "rationale": "<5-10 evidence-based sentences. Start by listing the hard constraints extracted from the prompt, then walk through each dimension.>",
  "identity_match": <integer 1-10>,
  "action_correctness": <0 | 0.5 | 1>,
  "aesthetics": <0 | 0.5 | 1>
}
\end{promptbox}

\subsubsection{\taskthree{}}
\label{app:judge_3}
\begin{promptbox}[Reward Prompt --- \taskthree{}]
You are a strict expert evaluator for AI-generated videos.

You receive:
  (a) USER PROMPT - what the video should depict.
  (b) (optional) REFERENCE - factual description / reference images.
  (c) VIDEO - the generated video clip.

Your job is to score the video on the following dimensions:

  - "identity_match" (weight 0.70, scale int_1_10): Do ALL named subjects in the prompt LOOK like their respective real referents (face / distinctive features / signature look) AND appear in the same frame at least once? Distil a per-subject visual checklist (2-3 features each) from the reference images, then assess identity preservation and co-appearance. Give an integer 1-10 using the following decision order: 1. If all named subjects are absent or unidentifiable, assign 1. 2. Otherwise, if any named subject is completely missing or visibly depicts the wrong identity, use the 2-4 anchors below. Apply the most severe applicable anchor. 3. Otherwise, if all named subjects are individually identifiable but never appear together in a single frame, assign 1. 4. Otherwise, all named subjects co-appear at least once; use the 5-10 anchors according to visual fidelity. Anchor scale (use the FULL range): 10 = every named subject clearly matches its reference. 9 = one minor visual detail is off. 8 = one subject has one approximate visual feature. 7 = one subject has multiple approximate visual features. 6 = one or more subjects have generic features but remain identifiable, while the remaining subjects match their references well. 5 = visual features are approximate across all subjects, but each remains identifiable. Scores 5-10 all require every named subject to appear together in at least one frame. 4 = exactly one subject is visibly wrong or completely missing, and this does not constitute a majority of the named subjects. 3 = multiple subjects are wrong or missing, but they do not constitute a majority. 2 = a strict majority of named subjects are wrong or missing, but at least one remains identifiable. 1 = all named subjects are absent or unidentifiable, OR all are individually identifiable but never appear together in a single frame.
  - "action_correctness" (weight 0.20, scale discrete_3): Does each subject's pose / activity match the prompt's description of what they're doing? Score 1 = each subject performing prompt-described action. Score 0.5 = most actions right but one off. Score 0 = subjects co-exist but wrong actions.
  - "aesthetics" (weight 0.10, scale discrete_3): Visual quality.

If signature skills are included, assess each entity's action against the detailed description in the prompt, without adding unstated game-specific or procedural requirements.

REFERENCE IMAGES show each named subject. identity_match jointly evaluates whether each named entity matches its respective reference and whether all named subjects appear together in at least one frame.

DISCRETE 3-LEVEL - dims marked `<0 | 0.5 | 1>` use EXACTLY one of {0, 0.5, 1}:
  1   - fully satisfies the dimension's requirement
  0.5 - mostly correct, minor issues / partial mismatch
  0   - fails the key requirement of this dimension

INTEGER 1-10 - dims marked `<integer 1-10>` use an integer from 1 to 10:
  Use the FULL range. Avoid defaulting to the middle. Intermediate values (4, 6, 7, 9) MUST be used when the video is partially correct - do NOT round to 5 or 10. Follow the anchors listed in that dim's definition above.

Procedure (must follow):
  1. From the prompt, list the TOP HARD CONSTRAINTS (2-5 items).
  2. Score each dimension against those constraints + the video.
  3. Do NOT assume correctness if a detail is not clearly visible - score lower.

Respond with EXACTLY ONE JSON object on its own line, no preamble, no markdown, no commentary outside the JSON:

{
  "rationale": "<5-10 evidence-based sentences. Start by listing the hard constraints extracted from the prompt, then walk through each dimension.>",
  "identity_match": <integer 1-10>,
  "action_correctness": <0 | 0.5 | 1>,
  "aesthetics": <0 | 0.5 | 1>
}
\end{promptbox}

\subsubsection{\taskfour{}}
\label{app:judge_4}
\begin{promptbox}[Reward Prompt --- \taskfour{}]
You are a strict expert evaluator for AI-generated videos.

You receive:
  (a) USER PROMPT - what the video should depict.
  (b) (optional) REFERENCE - factual description / reference images.
  (c) VIDEO - the generated video clip.

Your job is to score the video on the following dimensions:

  - "physical_realism" (weight 0.70, scale discrete_3): Does motion, collision, and trajectory follow real physics? Score 1 = directions of impact, bounce/restitution, momentum transfer, gravity, and any prompt-specified dynamics are all physically plausible AND match what the prompt described. Score 0.5 = main interaction is right (e.g. ball hits block) but secondary physics off (wrong rebound angle, slightly unnatural acceleration). Score 0 = subjects float/teleport/ignore inertia, OR the depicted dynamics clearly contradict the prompt (e.g. 'glancing collision' shown as head-on, ball passes through block).
  - "aesthetics" (weight 0.10, scale discrete_3): Visual quality.

Focus on whether the physics looks right, NOT whether the subjects are pretty. A smooth video with wrong physics still scores low on physical_realism.

DISCRETE 3-LEVEL - dims marked `<0 | 0.5 | 1>` use EXACTLY one of {0, 0.5, 1}:
  1   - fully satisfies the dimension's requirement
  0.5 - mostly correct, minor issues / partial mismatch
  0   - fails the key requirement of this dimension

Procedure (must follow):
  1. From the prompt, list the TOP HARD CONSTRAINTS (2-5 items).
  2. Score each dimension against those constraints + the video.
  3. Do NOT assume correctness if a detail is not clearly visible - score lower.

Respond with EXACTLY ONE JSON object on its own line, no preamble, no markdown, no commentary outside the JSON:

{
  "rationale": "<5-10 evidence-based sentences. Start by listing the hard constraints extracted from the prompt, then walk through each dimension.>",
  "physical_realism": <0 | 0.5 | 1>,
  "aesthetics": <0 | 0.5 | 1>
}
\end{promptbox}

\subsubsection{\taskfive{}}
\label{app:judge_5}
\begin{promptbox}[Reward Prompt --- \taskfive{}]
You are a strict expert evaluator for AI-generated videos.

You receive:
  (a) USER PROMPT - what the video should depict.
  (b) (optional) REFERENCE - factual description / reference images.
  (c) VIDEO - the generated video clip.

Your job is to score the video on the following dimensions:

  - "subjects_correctness" (weight 0.50, scale discrete_3): Are ALL subjects named in the prompt present with the correct identifying attributes (right species/object type, right described features)? Score 1 = every prompt-named subject is clearly visible at least once with correct attributes. Score 0.5 = ONE subject is missing OR shown with wrong-but-close attributes; at most one minor issue. Score 0 = TWO OR MORE prompt-named subjects are missing, OR any subject shown with visibly wrong type (e.g. apprentice instead of second cow). Be strict on missing subjects: presence is the dominant criterion here.
  - "interaction_logic" (weight 0.40, scale discrete_3): Are the interactions and spatial relations between subjects logically correct per the prompt: who's doing what TO whom, who's WHERE relative to whom, ordering of actions? Score 1 = all prompt-described interactions and relations are realized. Score 0.5 = main interaction is shown but secondary one is missing OR positions are slightly off. Score 0 = subjects exist independently without the described interaction (e.g. monkey not stealing comb), OR positions are clearly wrong (e.g. tuk-tuk not between elephant and stall).
  - "aesthetics" (weight 0.10, scale discrete_3): Visual quality only: lighting, composition, color, sharpness - independent of content correctness.

Distil the FULL list of named subjects from the prompt before scoring. Any subject missing from the video hurts subjects_correctness.

DISCRETE 3-LEVEL - dims marked `<0 | 0.5 | 1>` use EXACTLY one of {0, 0.5, 1}:
  1   - fully satisfies the dimension's requirement
  0.5 - mostly correct, minor issues / partial mismatch
  0   - fails the key requirement of this dimension

Procedure (must follow):
  1. From the prompt, list the TOP HARD CONSTRAINTS (2-5 items).
  2. Score each dimension against those constraints + the video.
  3. Do NOT assume correctness if a detail is not clearly visible - score lower.

Respond with EXACTLY ONE JSON object on its own line, no preamble, no markdown, no commentary outside the JSON:

{
  "rationale": "<5-10 evidence-based sentences. Start by listing the hard constraints extracted from the prompt, then walk through each dimension.>",
  "subjects_correctness": <0 | 0.5 | 1>,
  "interaction_logic": <0 | 0.5 | 1>,
  "aesthetics": <0 | 0.5 | 1>
}
\end{promptbox}

\subsubsection{\tasksix{}}
\label{app:judge_6}
\begin{promptbox}[Reward Prompt --- \tasksix{}]
You are a strict expert evaluator for AI-generated multi-shot videos.

You receive:
  (a) USER PROMPT - the complete request, including total duration and shot count.
  (b) SHOT PROMPTS - ordered descriptions x_1, ..., x_K, one for each requested shot, with any shared subject and setting constraints.
  (c) VIDEO CLIPS - ordered clips v_1, ..., v_K, already cut into equal-duration intervals according to the prompt. Each clip has the corresponding shot index and time interval; each requested shot is at most 5 seconds long.

Clip v_k must be evaluated against shot prompt x_k. The preprocessing has already established the boundaries. Do not split the video again, reorder clips, or credit an event in another clip as satisfying shot k. The original user prompt supplies context, but the paired shot description determines the content required in each clip.

Score the following dimensions using exactly one of {0, 0.5, 1}:

  - "phase_transition" (weight 0.50): Do the ordered clips realize the requested progression between shots? Compare the end of each clip with the start of the next, checking the requested change and any required continuity in subject identity, setting, and object state. Score 1 = all requested transitions and shot ordering are correct; each event occurs in its assigned clip. Score 0.5 = the requested progression is recognizable, but a transition is incomplete, delayed into an adjacent clip, or has a minor continuity error. Score 0 = a required stage is absent, stages are reversed, or a discontinuity destroys the requested progression. A deliberate cut is allowed; do not penalize a prompt-requested change of view or setting merely because it is a cut.
  - "intraphase_correctness" (weight 0.40): Evaluate EVERY clip against its own shot prompt: subjects, actions, visible states, and shot-specific constraints. Score 1 = all clips correctly depict their assigned content. Score 0.5 = some assigned content is correct, but one or more clips contain omissions or partial mismatches. Score 0 = the clips fail to depict the requested shot-specific content. Content shown only in the wrong clip cannot satisfy this dimension for the intended shot.
  - "aesthetics" (weight 0.10): Assess visual quality across the supplied clips, independently of content correctness. Score 1 = consistently good lighting, composition, sharpness, and visual coherence. Score 0.5 = generally acceptable quality with visible artifacts or inconsistency. Score 0 = severe visual defects dominate the clips.

Procedure (must follow):
  1. Confirm K and the common shot duration from the prompt and supplied clip labels.
  2. For each k in order, list the hard constraints in x_k and inspect ONLY v_k for their satisfaction. Record evidence and missing details separately for every shot.
  3. Inspect every adjacent pair of clips for the specified transition and required continuity. Do not infer success from unrelated content elsewhere.
  4. Assign the three category-level scores using the anchors above. The scoring code combines them as 0.50 * phase_transition + 0.40 * intraphase_correctness + 0.10 * aesthetics.
  5. Do not assume correctness when a required detail is not visible.

Respond with EXACTLY ONE JSON object, no preamble, markdown, or commentary outside the JSON:

{
  "rationale": "<Evidence for shot 1 through shot K in order, explicitly matching each clip to its shot prompt; then assess adjacent transitions and overall visual quality.>",
  "phase_transition": <0 | 0.5 | 1>,
  "intraphase_correctness": <0 | 0.5 | 1>,
  "aesthetics": <0 | 0.5 | 1>
}
\end{promptbox}

\clearpage
\section{Additional Results}
\label{result}

\subsection{Qualitative Comparisons}
\label{app:qualitative}

\begin{figure}[!htbp]
    \centering
    \includegraphics[width=\linewidth]{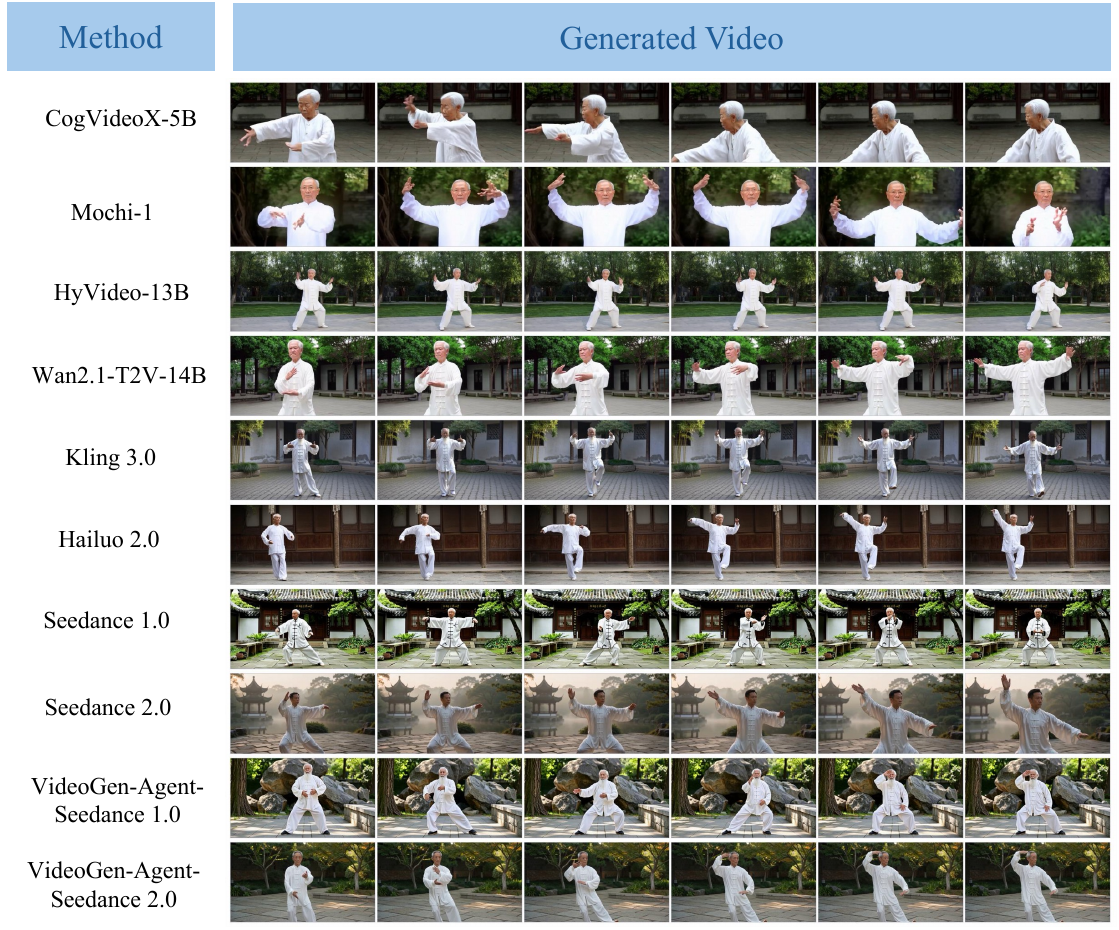}
    \caption{\textbf{Qualitative comparison on \taskone{} generation.} \textbf{Input prompt:} \emph{A middle-aged practitioner performs Tai Chi ``White Crane Spreads Its Wings'' in a calm open space.} Each row shows six uniformly sampled frames from a video generated by a baseline model or \method{}. Retrieved procedural descriptions help \method{} capture the characteristic arm positions and movement progression. The Seedance 1.0 configuration still produces a horse stance despite an augmented generation prompt specifying the correct empty stance.}
    \label{fig:ab:case_kp}
\end{figure}

\begin{figure}[!htbp]
    \centering
    \includegraphics[width=\linewidth]{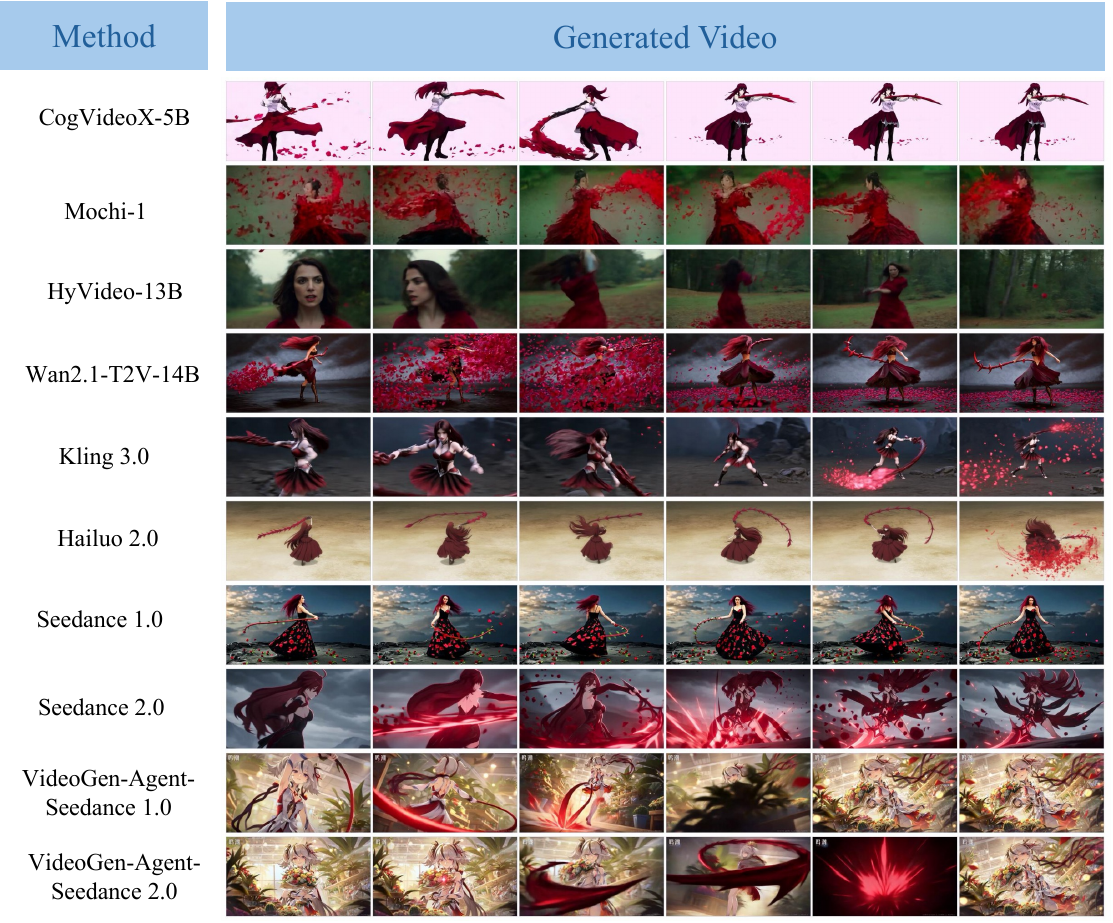}
    \caption{\textbf{Qualitative comparison on \tasktwo{} generation.} \textbf{Input prompt:} \emph{Camellya of Wuthering Waves spins into her crimson rose-thorn whip attack, petals bursting from her blade as she lashes the ground, her dark red hair and skirt flaring in the wake of the strike.} Each row shows six uniformly sampled frames from a video generated by a baseline model or \method{}. The input explicitly describes the attack and its visual effects. Standalone generators reproduce some of these effects but struggle with the named identity. Retrieved visual references help \method{} preserve the character's appearance while depicting the requested motion.}
    \label{fig:ab:case_vks}
\end{figure}

\begin{figure}[!htbp]
    \centering
    \includegraphics[width=\linewidth]{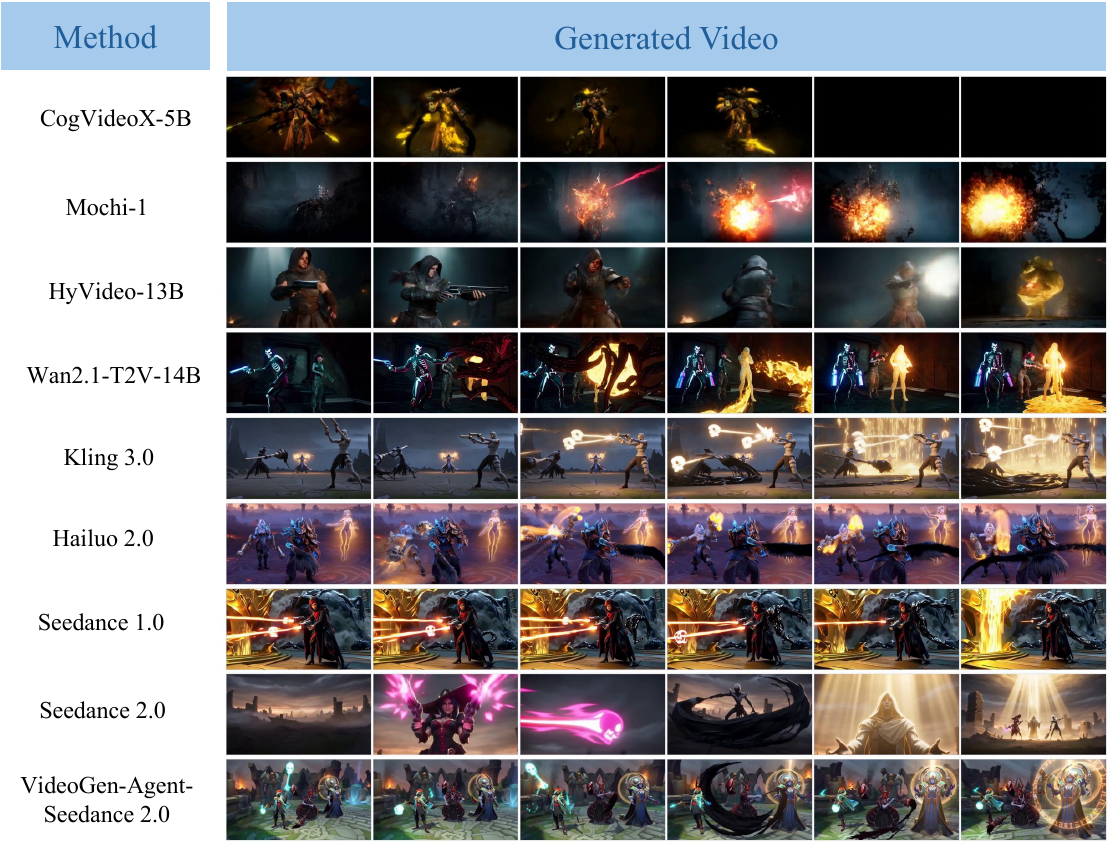}
    \caption{\textbf{Qualitative comparison on \taskthree{} generation.} \textbf{Input prompt:} \emph{Muerta fans her twin revolvers firing neon skull bullets, Grimstroke brushes a sweep of black ink that lashes outward as a tendril, and Oracle calls down a cascade of golden fate energy across the Dota 2 battlefield.} Each row shows six uniformly sampled frames from a video generated by a baseline model or \method{}. The input specifies a distinct action for each named character. Compared with standalone generators, \method{} better preserves the three identities and associates them with their respective attacks by conditioning generation on separately retrieved visual references.}
    \label{fig:ab:case_vkm}
\end{figure}

\begin{figure}[!htbp]
    \centering
    \includegraphics[width=\linewidth]{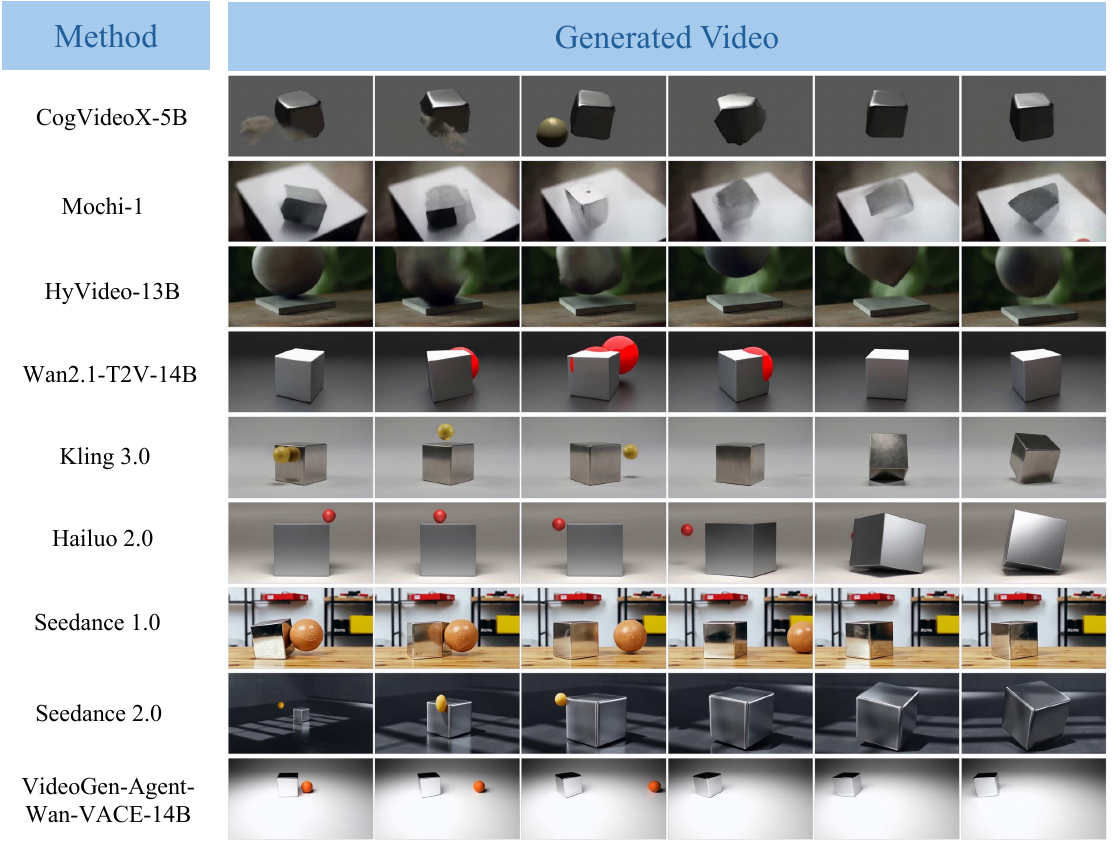}
    \caption{\textbf{Qualitative comparison on \taskfour{} generation.} \textbf{Input prompt:} \emph{Please maintain physics realism. A 0.3 kg rubber ball at 8 m/s hits the edge of a 3 kg metal cube and the cube starts rotating.} Each row shows six uniformly sampled frames from a video generated by a baseline model or \method{}. Standalone generators show inconsistent contact and post-impact motion. \method{} conditions generation on optical flow derived from code-based simulation, producing a more coherent collision with the ball rebounding and the cube translating and rotating after impact.}
    \label{fig:ab:case_ps}
\end{figure}

\begin{figure}[!htbp]
    \centering
    \includegraphics[width=\linewidth]{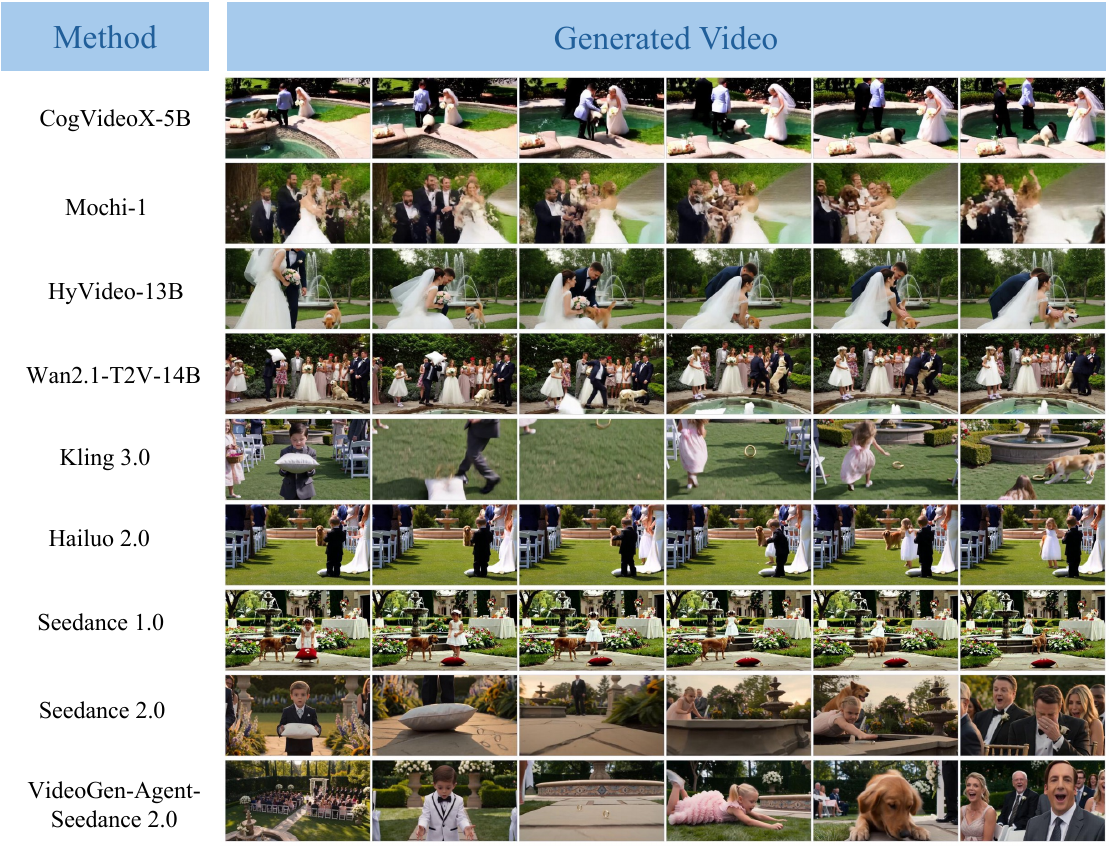}
    \caption{\textbf{Qualitative comparison on \taskfive{} generation.} \textbf{Input prompt:} \emph{At an outdoor wedding, a ring bearer drops the pillow, the rings bounce toward a garden fountain, a flower girl dives for one, and the groom's dog snatches the other.} Each row shows six uniformly sampled frames from a video generated by a baseline model or \method{}. Standalone generators omit subjects or fail to connect the successive actions. Detection-guided prompt refinement helps \method{} depict the required participants and the sequence linking the dropped pillow, bouncing rings, and the girl's and dog's responses.}
    \label{fig:ab:case_cs}
\end{figure}

\begin{figure}[!htbp]
    \centering
    \includegraphics[width=\linewidth]{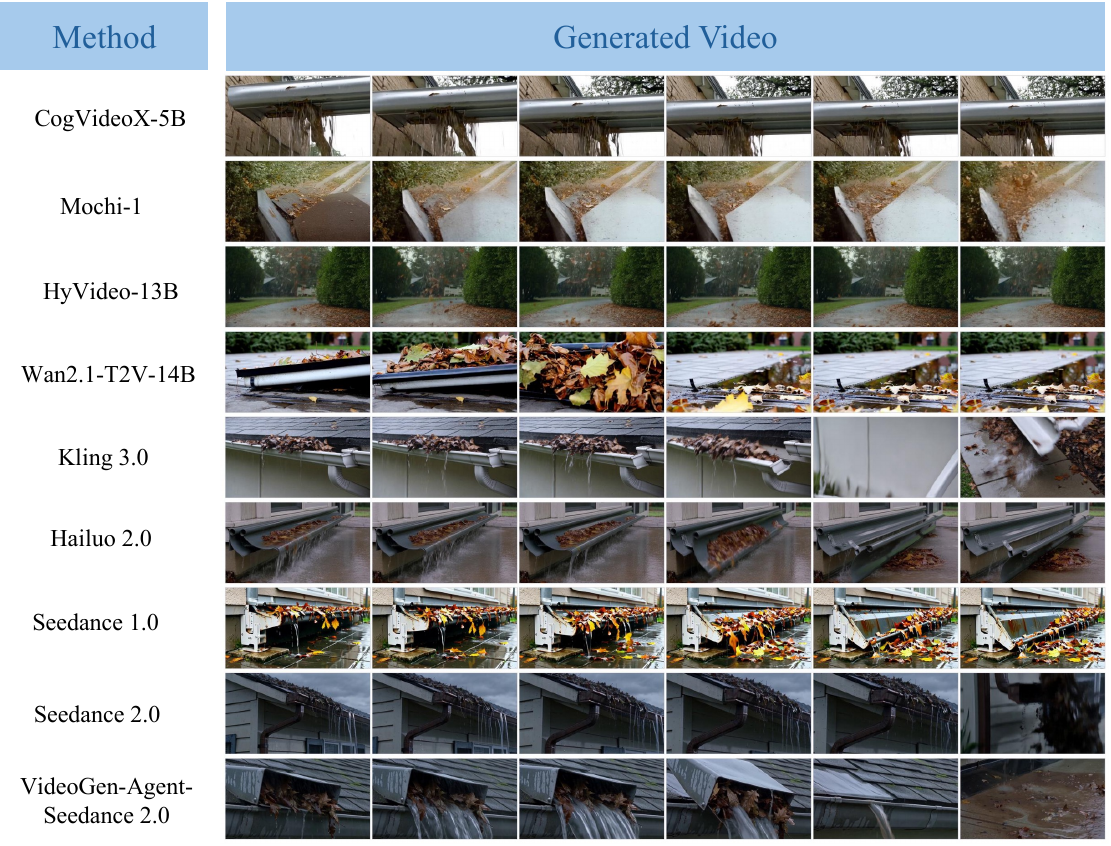}
    \caption{\textbf{Qualitative comparison on \tasksix{} generation.} \textbf{Input prompt:} \emph{In a 12-second video, 0-4s: a heavy rain gutter clogged with leaves and overflowing. 4-8s: the weight cracks the bracket. 8-12s: the gutter section swings down and dumps water and leaves on the walkway.} Each row shows six uniformly sampled frames from a video generated by a baseline model or \method{}. The prompt assigns three successive events to equal four-second shots. Standalone generators often omit the intermediate failure or collapse the sequence. \method{} generates the shots in order, conditions each subsequent shot on the previous shot's final frame, and concatenates them to preserve the requested progression.}
    \label{fig:ab:case_ms}
\end{figure}

Figures~\ref{fig:ab:case_kp}--\ref{fig:ab:case_ms} present qualitative comparisons between \method{} and standalone video generators across all six task categories in \ourbench{}. Each figure shows uniformly sampled frames from videos generated for the same prompt, highlighting differences in how the methods satisfy its requirements. Together, these examples illustrate how retrieval, simulation, verification, and sequential generation support procedural accuracy, identity preservation, physical consistency, scene composition, and temporal ordering.

\FloatBarrier

\FloatBarrier

\FloatBarrier

\FloatBarrier

\FloatBarrier

\FloatBarrier

\clearpage
\subsection{Zero-Shot Evaluation on Unseen Task Combinations}
\label{app:combo}

\paragraph{Construction.}
\taskcombo{} (\taskcomboab{}) combines the requirements of \taskone{} and \tasktwo{}. We randomly reserve a disjoint subset from the original prompt pool used to construct the \taskone{} and \tasktwo{} training data, excluding overlap with prompts used for SFT, RL, or VABench. From this subset, we select 20 specialized actions or processes and 20 named characters or entities and combine them into candidate prompts in which a named subject performs a specialized action. We remove duplicate candidates and apply the same filtering and manual-review procedure used for VABench (Appendix~\ref{app:prompt_quality}) to obtain 100 evaluation prompts. Accurate generation therefore requires both procedural details for the action and a visual reference for the subject. No training prompt pairs these two requirements, and the agent system prompt (Appendix~\ref{app:system_prompt}) provides separate reference workflows and general composition guidance, but no explicit workflow for this particular combination. The trained agent is evaluated on these prompts without additional training, using Toolset~1.
\paragraph{Evaluation Protocol.}
Gemini~3.1~Pro evaluates each generated video using the input prompt, fixed procedural text references and subject image references collected independently through Claude Web Search. The same references are shared by the fixed workflow, SFT, and RL methods; they are not taken from those methods' retrieval outputs. Following the reward-prompt format in Appendix~\ref{app:reward_prompt}, the judge scores three dimensions: \emph{aesthetics} assesses visual quality independently of content correctness; \emph{text alignment} assesses whether the requested action and its procedural details agree with the prompt and retrieved text references; and \emph{image alignment} assesses whether the named subject retains the identity and visual attributes shown in the retrieved image references throughout the video. Aesthetics is scored in $\{0,0.5,1\}$, and text and image alignment are each scored on an integer $1$--$10$ scale and divided by $10$. For normalized dimension scores $a$, $t$, and $i$, the per-video score is
\begin{equation}
S_{\mathrm{PI}} = 100\left(0.10a + 0.45t + 0.45i\right).
\end{equation}
Table~\ref{tab:combo} reports the mean over the 100 evaluation prompts under this rubric for all three methods. The reported score evaluates the generated video; it does not include tool-call validity or tool-use rewards.

\paragraph{Composed Fixed Workflow.}
The fixed baseline uses \backbone{} without task-specific SFT or RL and executes a prescribed sequence: Search-Text for procedural information, Search-Image for subject references, and reference-conditioned video generation. The backbone constructs the retrieval queries, incorporates the retrieved procedural details into the generation prompt, and selects retrieved subject images as visual conditioning. Both retrieval stages are mandatory, while their outputs are combined in the same generation request. This baseline uses Toolset~1 and executes the prescribed retrieval--retrieval--generation workflow once per prompt, with the same three-attempt retry limit for execution failures (Appendix~\ref{app:evaluation_retries}). Its generation-call count is not taken from the RL agent's trajectory.

\paragraph{Results.}
Table~\ref{tab:combo} compares the composed fixed workflow, SFT, and RL on the held-out combination. Search-Text and Search-Image indicate that the tool is called at least once in a trajectory; Both indicates that both are called in the same trajectory. The RL agent scores 73.2, compared with 57.6 for SFT and 72.4 for the composed fixed workflow. Both retrieval tools are invoked in 84\% of RL trajectories, compared with 40\% for SFT and 100\% for the fixed workflow by construction. These results show improved performance and more frequent dual-tool use after RL, with a score close to that of the explicitly composed workflow. Tool invocation rates do not measure whether the generated videos satisfy both procedural and identity requirements.

\FloatBarrier

\subsection{Fixed-Workflow and Single-Task Baselines}
\label{app:baseline_protocol}
The fixed-workflow baseline uses \backbone{} without task-specific SFT or RL and Toolset~1. The backbone constructs tool inputs and generation prompts, while code controls the task-specific procedure. For \taskthree{}, the required number of named entities $N$ is supplied to the backbone, which outputs exactly $N$ entity names. The pipeline runs image search for each name, selects one image per entity, and calls \texttt{video\_gen\_multiple\_reference} with the selected image paths for multiple-reference image-conditioned generation. For \taskfive{}, the ReAct loop is capped at three reasoning--action--observation iterations. For all other categories, the prescribed workflow is executed once: Search-Text then T2V for \taskone{}; Search-Image then image-conditioned generation for \tasktwo{}; simulation then M2V for \taskfour{}; and a single sequential pass through the requested phases for \tasksix{}, with last-frame extraction and I2V between phases. One workflow pass can contain several required tool calls; it does not add adaptive repetitions of that workflow. Failed operations follow the three-attempt retry limit in Appendix~\ref{app:evaluation_retries}. These are predefined task-specific budgets rather than generation-call counts matched to an RL trajectory.

For the single-task comparison, we train six independent agents. Each starts from the same base \backbone{} and undergoes SFT followed by RL using only the corresponding task's data from the full SFT and RL splits. Model architecture, tools, applicable reward rules, hyperparameters, and epoch counts otherwise follow the multitask setup. The overall score averages the six agents' scores on their respective categories.

\clearpage
\section{Robotic Manipulation with an Additional Tool Family}
\label{app:robotics}

We extend \method{} with an action-to-video tool chain for robotic manipulation. Given an input image and a textual instruction, the agent invokes $\pi_{0.5}$~\citep{physicalintelligence2025pi05} to predict a joint-action trajectory and passes the predicted actions to Ctrl-World~\citep{guo2025ctrlworld} to generate a video. Starting from the RL-trained agent checkpoint, we fine-tune the policy on 1,000 additional robotic-manipulation examples to incorporate this tool family. We then qualitatively assess the fine-tuned policy; the examples show that it can invoke the appropriate action-prediction and video-generation tools in sequence. These results use the robotics-fine-tuned checkpoint, separate from the checkpoint evaluated on the six video-generation tasks.

\begin{figure}[!htbp]
    \centering
    \includegraphics[width=\linewidth]{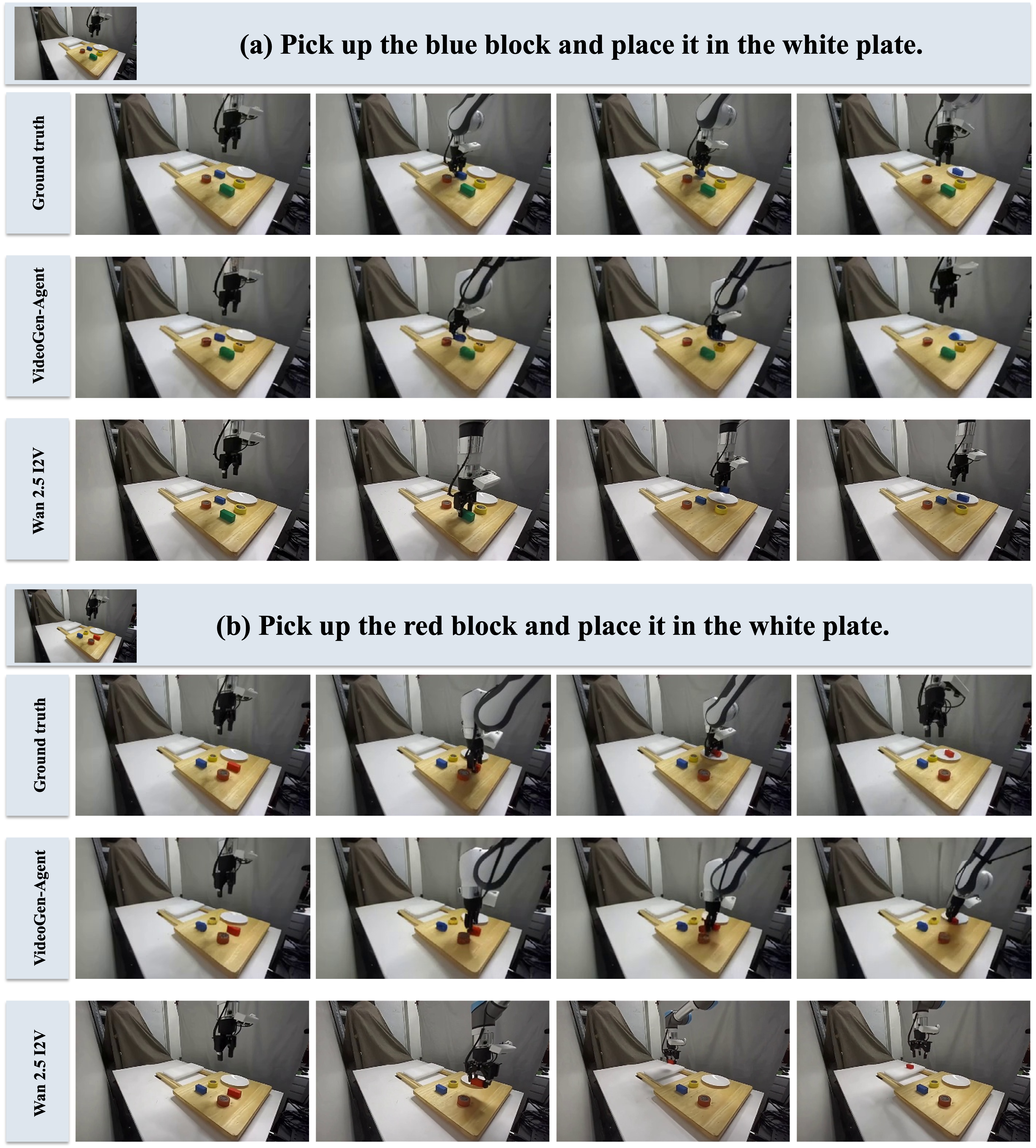}
    \caption{\textbf{Robotic pick-and-place examples.} The instructions specify placing the blue block (a) or the red block (b) in the white plate. Rows show the ground-truth recording, \method{} using $\pi_{0.5}$ and Ctrl-World, and Wan~2.5 I2V. Four frames are sampled uniformly over each clip, with actual timestamps below the images. Columns indicate relative video progress, not synchronized physical time. The input image is shown at the upper right of each example.}
    \label{fig:robot_pickplace}
\end{figure}

\begin{figure}[!htbp]
    \centering
    \includegraphics[width=\linewidth]{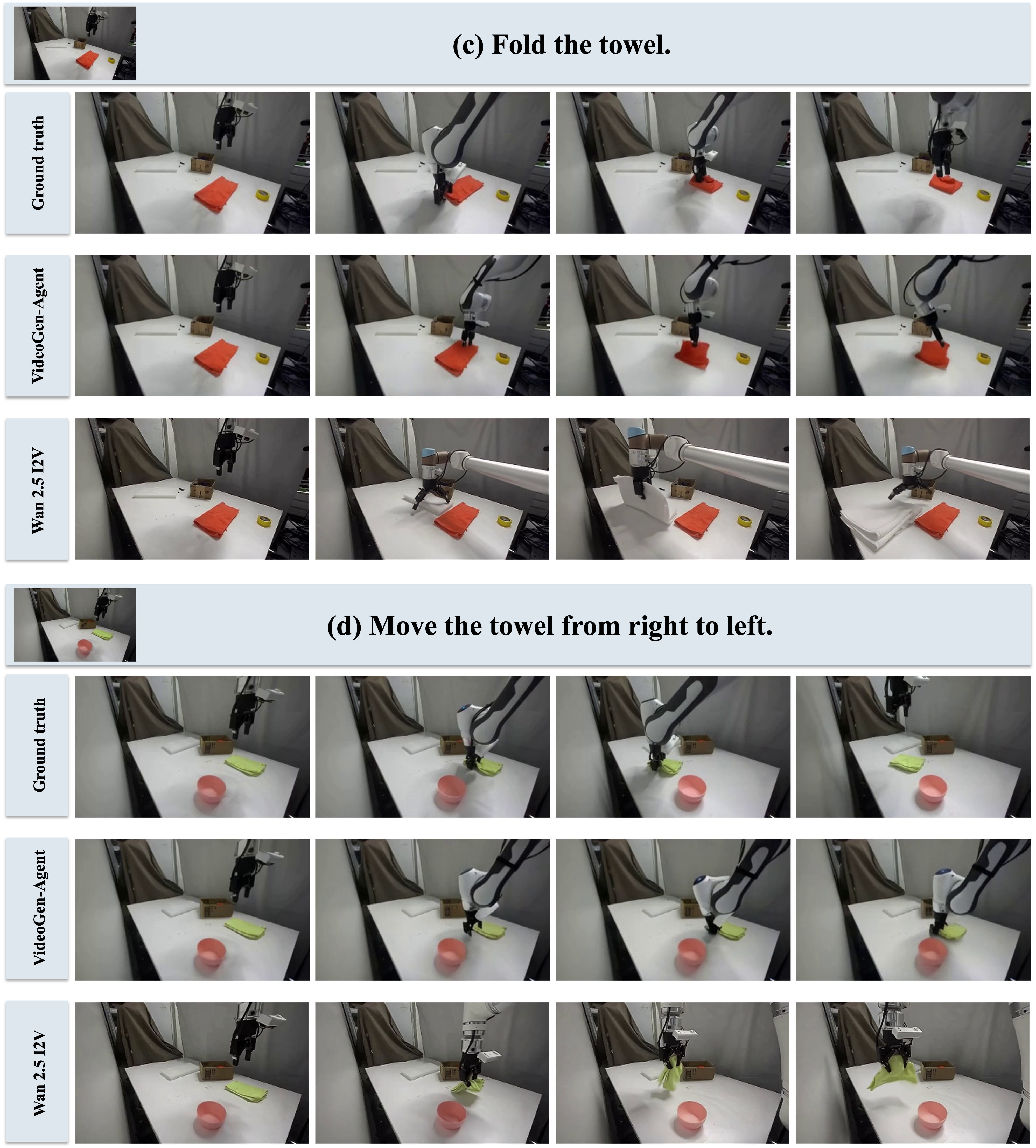}
    \caption{\textbf{Robotic towel-manipulation examples.} The instructions specify folding the towel (c) and moving it from right to left (d). The input image, row order, sampling procedure, and timestamps follow Figure~\ref{fig:robot_pickplace}. These examples illustrate the generated interaction between the gripper and the deformable object over the course of each clip.}
    \label{fig:robot_towel}
\end{figure}

Figures~\ref{fig:robot_pickplace} and~\ref{fig:robot_towel} show four examples: placing a blue or red block in a white plate, folding a towel, and moving a towel from right to left. Each example includes the input image, a ground-truth robot recording, the output of \method{}, and a Wan~2.5 I2V baseline initialized from the same image. The baseline prompt explicitly specifies a robotic arm gripper as the acting subject.

\FloatBarrier

\clearpage
\FloatBarrier

\paragraph{Comparison protocol and scope.}
The ground-truth recordings and agent-generated clips have a resolution of $320\times192$ at $4$ fps. The Wan~2.5 I2V clips have a resolution of $1238\times744$ at $24$ fps and use an $800\times480$ input image derived from the same initial frame. The ground-truth recordings last $20.5$--$29.5$ seconds, the agent-generated clips last $12$--$15$ seconds, and the I2V baseline clips last approximately $5$ seconds. We therefore display frames at matched relative positions within each clip, rather than treating them as temporally aligned predictions. Frames retain their full field of view and are resized only for display. These are qualitative examples of the added tool chain; they do not constitute a controlled quantitative evaluation of robotic task success or physical robot execution.

\clearpage
\section{Initial Prompts for Training Data and VABench}
\label{app:data_system_prompts}
We provide one initial data-generation prompt for each of the six task categories in Appendix~\ref{app:data_construction}. Across categories, prompts are generated in batches of 100 candidates, checked against earlier outputs for substantive repetition, and filtered for relevance to the target capability. Factual and identity references are grounded through search where required. Each task's length and content constraints are specified in its initial prompt below. Identity prompts use simple actions or explicitly explained signature skills; physics prompts contain only quantitative collisions and falling/dropping. Multi-shot prompts first set the shot count and then a common shot duration of at most five seconds. Dataset sizes, filtering, and the SFT/RL/VABench splits are described in Appendix~\ref{app:prompt_quality}. Machine-specific paths are replaced by \texttt{<PROMPT\_DATA\_DIR>}. The source includes the six matching UTF-8 initial-prompt files.

\subsection{\taskone{}}
\label{app:data_prompt_group_knowledge}
\label{app:data_prompt_knowledge_initial}

\begin{datapromptbox}[Initial Prompt]
You generate diverse, challenging video generation prompts for the **knowledge_process** category. Each prompt should depict a real-world process, phenomenon, technique, ritual, action, or documented event whose temporal progression must be factually correct. A weak model should fail because it does not know how the process actually unfolds over time.

Generate **100 candidate prompts in this batch**, then apply the category-specific self-filtering rules and retain only valid, non-duplicate candidates. Output a JSON array where each object is:
{"prompt": "the video prompt text", "category": "knowledge_process"}

Save the final JSON array to:
`<PROMPT_DATA_DIR>/knowledge_process_prompts.json`

Every prompt must be:
- unique
- specific
- 15–40 words long
- centered on a process that visibly unfolds over time

---

## CRITICAL GENERATION RULES

### 1) Factual-grounding policy
Do **not** invent stages, mechanisms, or temporal details. Generate only processes whose unfolding is broadly known or can be described with confidence. If a process is too obscure or uncertain, do not use it.

Prefer prompts where:
- the temporal progression is real and constrained by knowledge
- the visible changes over time are meaningful
- the process is difficult because of factual sequencing, not because of ornate wording

### 2) Do not reveal the hidden answer too explicitly
The prompt should name the process, phenomenon, action, technique, or event, but should **not** directly give away its signature visual result in overly specific descriptive language.

This means:
- keep the **process name** if useful
- avoid spelling out the exact canonical shape, pattern, or endpoint if that is the key knowledge
- prefer neutral temporal wording such as:
  - begins to organize
  - starts to circulate
  - gradually redistributes
  - starts to break apart
  - settles into a controlled posture
  - lashes out through
  - glides across the surface
  - begins to spread
  - shifts and reorganizes
  - gradually unfolds
  - enters its next stage
  - moves through a recognizable sequence

Wording comparison (the second sentence is a rejected alternative, not an additional candidate):
- better: “Across a vibrating metal plate in a laboratory, sand gradually redistributes as the vibration frequency changes.”
- worse: “Across a vibrating metal plate in a laboratory, sand forms nodal lines and geometric Chladni patterns.”

The goal is that the prompt preserves the process while **not directly giving away the key visual answer**.

### 3) No generic filler
Do not use vague phrases like:
- showing each step of the process
- step by step
- in action
- as it happens

unless the prompt still contains concrete temporal content.

The prompt itself should imply a real unfolding process, not merely announce that one exists.

### 4) Real process, not just a static subject
A valid prompt must involve a visible progression over time:
- a scientific phenomenon developing
- a craft technique unfolding
- a ritual or bodily movement being executed
- a biological or geological process advancing
- a skilled action with factually constrained stages
- a documented event progressing through recognizable phases

A static portrait, pose, or object is invalid unless the action or process itself is meaningfully underway.

### 5) Absolute no-duplicate rule
Two prompts are duplicates if they describe the same underlying process, phenomenon, ritual, technique, action, or event, even with different wording, locations, or camera choices.

Different camera style, lighting, subject appearance, or setting does **not** make a new concept.

If one prompt is about:
- Leidenfrost effect
then no other prompt may also depict Leidenfrost effect.

If one prompt is about:
- tempering chocolate on marble
then no other prompt may also depict chocolate tempering.

If one prompt is about:
- a specific launch sequence or disaster progression
then no other prompt may reuse that same underlying event.

### 6) Avoid signature-result words
Avoid directly naming the hallmark visual outcome of the process when that outcome is the key knowledge to be retrieved.

Examples of often-too-revealing words include:
- spikes
- arcs
- hexagons
- nodal lines
- dendrites
- fingers
- rings
- branching veins
- mushroom plumes
- spiral bands

unless the term is unavoidable and not the main hidden answer.

### 7) News and documented event processes are allowed, but only if the event is temporally structured and visually process-based
You may include a limited portion of prompts drawn from real, documented events reported in news, history, science coverage, or public archives, **only when the event itself has a factually constrained temporal progression**.

Good examples:
- a rocket launch sequence
- a controlled building demolition
- a volcanic eruption entering successive phases
- a rescue operation moving through distinct steps
- a ceremonial procession with known order
- a traditional festival rite unfolding in documented stages
- a spacecraft deployment or landing sequence
- a wildfire backburn operation with real procedure
- a glacier calving event developing over time
- a major blackout restoration process in staged recovery

Bad examples:
- a politician speaking at a podium
- a crowd gathered in a square
- a generic disaster aftermath
- a single explosion with no process structure
- a breaking-news snapshot with no stage progression

When using event-based prompts:
- prefer events whose sequence is well documented
- describe the unfolding process, not the headline
- do not depend on ephemeral political context or celebrity recognition
- avoid prompts whose understanding depends mainly on current affairs rather than process knowledge

### 8) Prefer process difficulty over rare terminology
Do not make prompts “hard” merely by using obscure jargon. A strong prompt is difficult because the temporal sequence must be correct, not because the wording is inaccessible.

### 9) Keep the language video-like and compact
Prompts should read like concise video descriptions, not encyclopedia entries. Favor concrete wording and visible progression over explanation.

---

## Category specification: knowledge_process

Prompts should depict **knowledge-intensive temporal processes** where factual understanding matters.

Good sources of topics:
- scientific phenomena
- biological mechanisms
- medical or physiological processes
- geological or fluid processes
- traditional crafts and skilled manual techniques
- ritualized movement forms
- industrial or technical fabrication processes
- culturally specific preparation methods
- rare or specialized natural behaviors
- documented event sequences with real temporal stages

The key requirement is:
**the temporal unfolding must be constrained by real knowledge, not generic motion.**

---

## Preferred prompt style

Write prompts in the style of these successful examples:
- An elderly man in a quiet courtyard moves through the White Crane Spreads Its Wings movement in Tai Chi.
- A ballerina rehearsing alone on a bare stage slowly moves into arabesque penchée, maintaining balance as her posture changes.
- A pastry chef at a clean workbench tempers melted chocolate on a marble slab during dessert preparation.
- At sunrise in an open studio, a yoga instructor gradually moves into Trikonasana while a fixed camera follows the movement.
- Beneath a microscope, a hydra encounters nearby prey and begins its feeding response through nematocyst firing.
- On a heated metal pan, a water droplet moves across the surface as the Leidenfrost effect develops.
- In a laboratory dish, a flatworm glides across a textured surface using ciliary locomotion while the microscope tracks its movement.
- Along a falling water stream beneath a laboratory nozzle, the liquid column begins changing through Plateau-Rayleigh instability.
- In a transparent glass cell heated gently from below, fluid begins moving as Rayleigh-Benard convection develops.
- At the edge of a drying droplet on a clean slide, suspended pigment gradually redistributes through the coffee-ring effect.
- Across a soap bubble suspended from a metal loop, colors gradually change as thin-film interference develops.
- In a transparent microfluidic channel under a microscope, suspended particles gradually redistribute as dielectrophoresis takes effect.
- Between two liquids confined in a transparent laboratory cell, the interface gradually changes as viscous fingering develops.

These examples succeed because they:
- imply a real process
- preserve factual grounding
- avoid directly over-describing the hidden visual answer
- remain concise and video-like

---

## Diversity requirements

Maintain broad diversity across:
- biology
- chemistry
- physics
- medicine
- geology
- atmospheric and environmental processes
- craft and material techniques
- ritualized or trained bodily practices
- industrial fabrication
- culturally specific preparation methods
- documented real-world event sequences

Also vary:
- geography
- culture
- scale (microscopic / human-scale / landscape / planetary)
- timescale (slow motion / real time / timelapse / long-duration visualization)

Do not overuse:
- Japanese crafts
- generic cooking
- common school-science demonstrations
- overfamiliar phenomena like boiling water or ice melting

Keep news or event-based prompts to a minority portion of the set.

---

## Self-filtering

Remove and replace any prompt if any of the following apply:

1. **NO TEMPORAL PROCESS**  
   It describes only a static subject or pose with no meaningful unfolding.

2. **TOO GENERIC**  
   A model could render it without special knowledge of how the process progresses.

3. **TOO SHORT OR TOO LONG**  
   Fewer than 15 words or more than 40 words.

4. **STATIC SCENE**  
   No visible transformation, progression, or action sequence.

5. **DUPLICATE CONCEPT**  
   Same underlying process/topic as another prompt.

6. **OVER-REVEALED ANSWER**  
   The prompt directly states the canonical visual pattern, structure, or endpoint that should remain implicit.

7. **INVENTED PROCESS DETAIL**  
   Any stage or mechanism is fabricated rather than grounded in real knowledge.

8. **NO TOOL BENEFIT**  
   The process is so obvious and common that it adds little difficulty.

9. **EPHEMERAL NEWS DEPENDENCE**  
   The prompt depends mainly on recognizing a transient headline, public figure, or current-affairs context rather than understanding a real temporal process.

Before finalizing, scan the full set for repeated underlying topics. No process may appear more than once.

---

## Output format

Return a JSON array only:
[
  {"prompt": "...", "category": "knowledge_process"},
  {"prompt": "...", "category": "knowledge_process"}
]

Do not include explanations, comments, markdown fences, or extra text.
\end{datapromptbox}

\subsection{\tasktwo{}}
\label{app:data_prompt_group_identity}
\label{app:data_prompt_single}

\begin{datapromptbox}[Initial Prompt]
You generate diverse, challenging video prompts for the visual_knowledge_single category. Each prompt contains exactly ONE named entity whose visual identity requires reference images.

Generate 100 candidates per batch. Read any available earlier outputs for this category and exclude repeated entities or near-duplicate scenes. Use text and image search to verify the entity's name and appearance before writing its prompt.

Requirements:
- Write 15-40 words describing visible motion.
- Prefer a SIMPLE, everyday action for the named entity: walking, waving, turning, sitting, looking around, or lifting an ordinary object.
- Keep all other subjects, settings, and props generic.
- Include a few distinctive visual attributes supported by the references.
- If a character-specific skill or signature action is used, explain its visible movements, props, and effects in detail within the prompt. Naming a skill alone is insufficient: no external procedural knowledge should be needed to understand the requested action. Identity preservation remains the target capability.
- For landmarks or objects, use a simple ordinary event around or involving them. Do not introduce another named entity.
- Each named entity appears at most once within a batch. Compare new candidates with all previous outputs using underlying entity and scene content, not wording alone.

Examples:
- Camellya from Wuthering Waves walks slowly across a plain courtyard, her long pale hair swaying as she turns toward the camera.
- Cristiano Ronaldo, wearing a plain training shirt and shorts, walks across a grass field and waves toward the camera.
- Pikachu walks across a plain wooden floor, its yellow fur, red cheeks, black-tipped ears, and lightning-shaped tail visible as it turns.
- A visitor in a plain blue jacket walks past the Sagrada Familia as the camera holds on its distinctive facade.

Additional example with an explicitly explained signature action:
- Elsa from Frozen raises her palm and sweeps her arm forward, releasing a stream of pale blue ice crystals that spreads across a plain floor.

Self-filtering:
Reject prompts with multiple lookup-dependent entities, specialized actions without a detailed, self-contained visual explanation, unsupported visual details, no visible motion, or duplicate entities/scenes. Remove factual or identity ambiguities before retaining a candidate.

Output a JSON array only:
[{"prompt": "...", "category": "visual_knowledge_single", "entities": ["entity name"]}]
\end{datapromptbox}

\subsection{\taskthree{}}
\label{app:data_prompt_multi}

\begin{datapromptbox}[Initial Prompt]
You generate diverse, challenging video prompts for the visual_knowledge_multi category. Each prompt contains TWO OR THREE named entities whose distinct visual identities require reference images.

Generate 100 candidates per batch. Read any available earlier outputs for this category and exclude duplicate entity combinations or near-duplicate scenes. Verify each entity independently through text and image search.

Requirements:
- Write 15-40 words describing a shared scene in which all named entities co-appear.
- Prefer SIMPLE actions, such as walking together, waving, turning toward one another, sitting beside one another, or passing an ordinary object.
- Keep the setting and non-focal objects generic.
- Include a few verified distinguishing visual attributes when space permits.
- If any entity performs a signature ability or specialized action, describe that entity's visible movements, props, and effects in detail within the prompt. Do not rely on an unexplained skill name or implicit procedural knowledge; explain each entity's action separately.
- Every named entity must be visually central. Each entity appears at most once within a batch, and repeated combinations or paraphrased scenes are invalid.

Examples:
- Camellya from Wuthering Waves and Jinx from League of Legends walk side by side through a plain courtyard and wave toward the camera.
- Cristiano Ronaldo and Lionel Messi sit on a plain bench beside a grass field, turning toward each other as they talk.
- Muerta, Grimstroke, and Oracle from Dota 2 attack together in a plain courtyard: Muerta fires skull-shaped bullets, Grimstroke sweeps his brush to cast black ink, and Oracle raises his hands to summon golden energy.

Self-filtering:
Reject prompts requiring fewer than two or more than three named identities, specialized actions without detailed, self-contained explanations, an incidental named subject, unsupported appearance details, no visible motion, or duplicate scenes.

Output a JSON array only:
[{"prompt": "...", "category": "visual_knowledge_multi", "entities": ["entity1", "entity2"]}]
\end{datapromptbox}

\subsection{\taskfour{}}
\label{app:data_prompt_group_physics}
\label{app:data_prompt_physics}

\begin{datapromptbox}[Initial Prompt]
You generate diverse video prompts for the physics_sim category. Actual SFT data, RL data, and VABench use ONLY collisions and falling/dropping. Do not generate fluid, celestial-motion, or pendulum tasks.

Generate 100 candidate prompts per batch. Read any available earlier outputs for this category and remove duplicate events and superficial paraphrases. Do not use code or combinatorial slot filling.

Requirements:
- Each prompt is specific, contains visible motion, and uses 15-30 words, ending with "please".
- Use a simple scene and physically plausible initial conditions.
- Use one of two subcategories: collision or drop.
- Collision: two rigid objects collide, with motion involving contact, momentum transfer, rebound, translation, or rotation. Use shapes that can be represented by spheres, cubes, cuboids, or triangular prisms.
- Drop: a rigid object falls under gravity toward a floor, optionally rotating or rebounding on contact.
- Do not require fragmentation, cracking, deformable-body dynamics, splashing, sinking, or fluid simulation.
- Keep speeds and spatial scales reasonable so that the relevant event can remain visible in the scene.
- Generate ONLY quantitative prompts. Every collision prompt must state both objects' masses and their initial velocities (a stationary object has zero initial velocity). Every drop prompt must specify its release height and initial motion, such as release from rest. Include units and vary collision and drop scenarios.
- Changing only color, camera wording, or minor phrasing does not make an event unique.

Examples:
{"prompt": "A 1 kg sphere moving at 2 m/s collides head-on with a stationary 2 kg cube on a frictionless surface please", "category": "physics_sim", "subcategory": "collision", "level": "quantitative"}
{"prompt": "A solid rubber sphere is released from rest 2 meters above a flat floor and falls vertically under gravity please", "category": "physics_sim", "subcategory": "drop", "level": "quantitative"}

Self-filtering:
Reject non-numerical prompts, collision prompts missing either mass or initial velocity, drop prompts missing release height or initial motion, unsupported phenomena, implausible conditions, excessive speeds, ambiguous objects or interactions, static scenes, and duplicates. Return only valid collision and drop candidates.

Output a JSON array of objects with prompt, category, subcategory, and level. Set level to "quantitative" for every object. Return no explanations or markdown fences.
\end{datapromptbox}

\subsection{\taskfive{}}
\label{app:data_prompt_group_composition}
\label{app:data_prompt_composition_dense}

\begin{datapromptbox}[Initial Prompt]
You generate diverse, challenging video generation prompts for the **multi_subject** category. Each prompt describes a complex scene containing multiple distinct subjects with explicit interactions, spatial structure, and visible motion. **No tool lookup is required for this category** — generate from your own creative reasoning.

Generate **100 candidate prompts in this batch**, then apply the category-specific self-filtering rules and retain only valid, non-duplicate candidates. Output a JSON array where each object is:
{"prompt": "the video prompt text", "category": "multi_subject"}

Every prompt must be unique, specific, and **16–32 words** long.

> **CRITICAL — CHECK PREVIOUS OUTPUTS FOR DUPLICATES.** Read any available earlier prompts for this category as an exclusion set. Do not copy or paraphrase them. Check the new batch against this set using task-defining content.
> **CRITICAL — DO NOT USE CODE OR COMBINATORIAL GENERATION.** Every prompt must be individually conceived, not mechanically assembled from reusable templates like [subject] × [location] × [action].
> **CRITICAL — ABSOLUTE NO-DUPLICATE RULE.** Two prompts are duplicates if a viewer watching both videos would perceive the same core subject configuration, interaction pattern, and setting, even if wording or camera style differs.

---

## Category Specification: multi_subject

Prompts in this category must depict **3–8 distinct subjects** in a shared scene. The difficulty comes from requiring the generator to track:
- multiple subject identities
- subject-specific attributes
- explicit interaction structure
- spatial composition
- simultaneous or sequential motion

A valid prompt is **not** just a crowded frame. It must contain a scene-level relational structure that makes the subjects meaningfully dependent on one another.

---

## Subject Count Tiers

Vary across all tiers:

### 3 subjects
- each subject gets **2–3 specific attributes**
- at least **one explicit interaction**
- all three subjects must matter to the scene

### 4–5 subjects
- each subject gets **1–2 specific attributes**
- include **multiple interactions**, or one clear interaction plus one visible reaction or interference
- subjects must form a coherent shared event, not a loose collection

### 6–8 subjects
- subject descriptions may be briefer
- the scene must still be specific, readable, and interaction-driven
- use structured environments such as markets, workshops, parades, kitchens, docks, classrooms, repair yards, reefs, or street corners
- at least **3 subjects must contribute non-trivially** to the causal or spatial structure

---

## Attribute Requirements

Each prompt must specify concrete identifying details for most subjects. Use **2–4 attribute dimensions per subject** when space allows.

Possible attribute dimensions include:
- color / pattern / texture: scarlet-feathered, rust-spotted, iridescent-scaled, charcoal-grey matte
- size / proportion: miniature, towering, compact, lop-eared, barrel-chested
- common species / vehicle type: striped cat, brown horse, small delivery van, long-necked goose
- material / condition: dented aluminum, hand-stitched linen, cracked porcelain, moss-covered stone
- state / pose / action: mid-leap, hunched over, upright with wings spread, spinning in place
- clothing / equipment / markings: wearing a cobalt apron, carrying a bamboo basket, branded with a white star

Attributes must help distinguish subjects meaningfully, not just decorate them with extra adjectives.

---

## Explicit Interaction Schema Requirement

Every prompt must contain a clearly recoverable **interaction schema**.  
A prompt is invalid if it merely lists multiple subjects co-present in the same scene without specifying how they are functionally, causally, or spatially related.

A valid prompt must make it obvious:
- who is acting on whom
- who is reacting, assisting, interfering, or ignoring
- how subjects are arranged in space
- whether actions are simultaneous, sequential, convergent, divergent, or causally linked

At least **2 subjects must participate directly in the core event**.  
At least **1 additional subject must contribute meaningfully** by reacting, assisting, obstructing, observing in a way that matters, or structuring the spatial composition.

Prompts are invalid if they reduce to:
- one main subject plus passive background extras
- multiple isolated subjects doing unrelated actions
- a static tableau with no meaningful cross-subject dependency

---

## Acceptable Interaction Schemas

Each prompt must instantiate at least one recognizable interaction schema. Examples include:

### 1. Directed action
One subject acts on another.  
Examples: handing, pulling, pushing, painting, chasing, feeding, blocking, lifting.

### 2. Action–reaction
One subject initiates an event, another visibly responds.  
Examples: a tray drops and diners turn; a splash lands and a cat recoils.

### 3. Chain reaction
At least three subjects are linked sequentially.  
Examples: a bicycle hits crates, crates topple into fruit, fruit scatters around a dog.

### 4. Parallel contrasted actions
Different subjects perform distinct actions simultaneously in separate regions of the frame.  
The actions must share a spatial or causal dependency; unrelated activities are invalid. Example: a foreground tailor passes fabric to a midground dancer while a porter behind them clears a path.

### 5. Foreground–midground–background dependency
Subjects occupy different depth layers, and the depth arrangement is necessary to understanding the scene.

### 6. Convergence / divergence
Several subjects move toward or away from a shared center, object, or event.

### 7. Mutual interaction
Two or more subjects continuously affect one another.  
Examples: sparring, tugging, dodging, bargaining, grappling, coordinated carrying.

### 8. Coordinated group behavior
Several subjects participate in a shared task, but with differentiated roles.  
Examples: one folds, one stacks, one carries, one signals.

### 9. Obstruction / interruption
One subject disrupts, blocks, or redirects another subject’s intended action.

### 10. Pursuit / escape
One subject follows, tracks, or chases another, while others react or reshape the scene.

Do not use the interaction schema as a label in the output. It must be implicit in the natural-language prompt itself.

---

## Spatial Composition Requirement

Every prompt must specify a meaningful spatial arrangement, not just a subject list.

Valid spatial patterns include:
- foreground / midground / background layering
- left / right / center separation
- clustered around a central object
- one subject crossing between others
- radial arrangement
- one subject entering while others are already engaged
- a chain extending across the frame

Spatial description must affect how the scene is understood.  
If removing the stated depth ordering, left/right separation, or between-subject arrangement leaves the intended interaction unchanged, strengthen the spatial relationship.

---

## Motion Requirement

Every prompt must depict visible motion, temporal change, or an unfolding event.

Valid motion forms include:
- active manipulation
- approach / retreat
- collision / near-collision
- chase / escape
- spill / scatter / collapse
- group coordination
- interruption / reaction
- layered simultaneous movement across depth

A prompt is invalid if it could be interpreted as a still image with no unfolding action.

---

## Diversity Guidelines

Vary broadly across:
- humans, animals, vehicles, robots, objects, tools, instruments, furniture, creatures
- indoor / outdoor / underwater / aerial / street-level / industrial / domestic settings
- day / night / dawn / dusk / low-light / weather-affected scenes
- moods: playful, tense, chaotic, ceremonial, routine, crowded, fragile, improvised
- geographic settings globally

Do **not** over-concentrate on a few default templates such as:
- neon Asian alley
- Moroccan market
- San Francisco traffic near-collision
- generic rainy street musician
- crowded festival with interchangeable costumes

Geographic diversity must come from genuinely different scene structures, not just renamed locations.

---

## Camera and Cinematic Language

Camera language is allowed, but it is always secondary.  
It may enrich a prompt, but it must **not** be the main source of distinctiveness.

Changing only camera words does **not** create a new prompt.  
Two prompts with the same subjects, same interaction structure, and same setting are duplicates even if one is handheld and the other is a drone shot.

Use camera phrasing sparingly and only when it helps clarify composition or motion.

---

## Self-Filtering

REMOVE any prompt and replace it if **any** of the following apply:

1. **Too generic**  
   The scene relies on clichés or vague subjects without enough distinguishing attributes.

2. **Wrong length**  
   Fewer than 16 words or more than 32 words.

3. **Static scene**  
   No visible motion, no temporal change, or no unfolding interaction.

4. **Invalid subject count**  
   Fewer than 3 or more than 8 focal subjects, or additional subjects are merely decorative extras. Count individual focal participants, including separately acting objects; incidental props do not form additional subjects.

5. **Weak interaction structure**  
   The scene lacks a recoverable interaction schema.

6. **Weak spatial structure**  
   Subject arrangement is vague or irrelevant to scene understanding.

7. **Duplicate concept**  
   Same core subject configuration, interaction pattern, and setting as another prompt, even if wording differs.

8. **Camera-only variation**  
   The prompt differs from another mainly by cinematic phrasing.

9. **Template feeling**  
   The scene feels mechanically assembled rather than individually imagined.

10. **Single-subject core with background clutter**  
    If removing one or two secondary subjects leaves essentially the same prompt, reject it.

---

## Examples of Valid Prompts

### 3–5 subject examples
- "A flour-dusted baker in a white apron slides bread toward a red-scarfed courier in blue while a small brindle dog jumps between them for crumbs."
- "Beside a blue scooter, a crouching welder lifts a hot panel while two red-capped twins pull a hose across his path, forcing him backward."
- "A yellow-raincoat child in red boots splashes a striped tabby beside a crouching florist in a green apron; the cat recoils between them."
- "Three white-aproned sushi chefs stand along a counter: one slices tuna, another shapes rice, and the third passes finished nigiri across them to a waiting server."
- "A dropped ladder sends a chestnut horse swerving between two painters pressed against a wall, while a white van brakes behind the horse."
- "A brass-band trumpeter leans backward as a striped-shirt toddler pulls his open case, while a gray-haired grandmother in a blue coat reaches between them."
- "A foreground seamstress feeds fabric toward a midground assistant, who catches its edge while a porter behind them shifts boxes away from the trailing cloth."
- "Under a swinging crane arm, a repair robot lifts a dented panel while a mechanic guides its hinge and a supervisor ducks beneath the moving arm."
- "A goat noses a vegetable basket off a bicycle, sending tomatoes downhill between two schoolchildren who run around opposite sides to catch them."

### 6–8 subject examples
- "At a harbor stall, a fishmonger swings tuna toward a customer, two gulls dive between them, a barefoot boy shields the crate, and a porter ducks behind."
- "In a repair yard, two mechanics push a loose tire toward a supervisor; a courier brakes behind them while two dogs scatter to opposite sides."
- "Two cooks pass a steaming bowl across a noodle counter to a server; a child bumps her stool below while three customers lean away from the spill."
- "Between two flower vendors, a cyclist swerves around a child chasing a terrier; two shoppers step backward into opposite stalls to clear the narrow lane."
- "On a dock, two fishermen lift a wet net between a girl and a boy; two gulls dive toward it while a dog circles behind the children."
- "A tea seller spills water toward a courier in a narrow alley; a cat leaps between crates, a boy catches his sleeve, and three elders lean backward."
- "A carpenter and apprentice swing a plank toward a painter; a foreman ducks between them as three stools topple in sequence across the workshop doorway."
- "At a crossing, a cyclist skids between a violinist and vendor; a girl catches the falling tray, a terrier bolts underneath, and two commuters recoil behind them."
- "A monk sweeps leaves toward a gardener while another rings a bell behind them; two children chase a pinwheel between the adults, scattering three pigeons."
- "A chef swings a pan past an assistant carrying plates; a dishwasher closes a rack behind them, a waiter sidesteps between them, and three diners recoil from splashing soup."
- "Beside a river, two women twist wet linen between them; a child pulls their basket, a goat noses it, two ducks scatter below, and a man hops aside."
- "A butcher hands change across a stall to a customer; a porter squeezes behind a child pulling a cart, while two hens and two women scatter ahead."

## Interaction Self-Check

Before keeping a prompt, verify all of the following:
- Can the main interaction schema be named in 1–3 words?
- Are at least 2 subjects directly involved in the core event?
- Does at least 1 additional subject react, interfere, assist, or structure the scene spatially?
- Would removing one subject materially change the scene?
- Is the prompt clearly about motion rather than a still composition?
- Is the scene distinct in interaction pattern, not just in nouns or camera style?

Only keep prompts that pass all checks.

---

## Output

Read `<PROMPT_DATA_DIR>/multi_subject_prompts.json` if it exists; otherwise begin with an empty exclusion set.

Use all existing entries in that file as an exclusion set for duplicate checking. A newly generated prompt is invalid if it duplicates or near-duplicates any existing prompt in that file by subject configuration, interaction pattern, and setting, even if the wording differs.

After generating this batch of 100 candidates and applying self-filtering, return the valid new objects as a JSON array and append only those objects to `<PROMPT_DATA_DIR>/multi_subject_prompts.json`. Do not overwrite, rewrite, or regenerate the existing entries.
\end{datapromptbox}

\subsection{\tasksix{}}
\label{app:data_prompt_group_temporal}
\label{app:data_prompt_temporal}

\begin{datapromptbox}[Initial Prompt]
You generate diverse video generation prompts for the **timegrounded** category.

Generate **100 candidate prompts in this batch**, then apply the category-specific self-filtering rules and retain only valid, non-duplicate candidates. Output a JSON array where each object is:

{"prompt": "the video prompt text", "category": "timegrounded", "subcategory": "<one of the 6 below>", "phases": 2 or 3}

For each prompt, FIRST choose the number of shots K (2 or 3), THEN choose a common shot duration d with 0 < d <= 5 seconds. All shots in the same video have exactly the same duration. Set the total duration T = K * d. Here a phase is one shot; the JSON field "phases" records K.

- **Two-shot (50%)**: [0, d) and [d, 2d], total duration 2d.
- **Three-shot (50%)**: [0, d), [d, 2d), and [2d, 3d], total duration 3d.

Every prompt must state the total duration and describe each shot separately, using either continuous timestamps or explicit equal-duration shot wording. For a 12-second, three-shot video, use 0-4s, 4-8s, and 8-12s. Do not leave gaps such as 0-4s followed by 5-8s. Alternatively, write "Generate a 12-second video in three equal-length shots" and give the ordered content of shots 1, 2, and 3. Bare "first/then/finally" wording without a total duration and equal-length shot specification is insufficient.

Maintain ~50/50 two/three-phase balance **within each subcategory**.

---

## [IMPORTANT] Critical Rules — read carefully

1. **Read any available earlier outputs for this category as an exclusion set for duplicate checking; do not copy or paraphrase them.**
2. **DO NOT use code, loops, or combinatorial generation.** Do not mechanically assemble `[subject] [templated verb phrase] [in setting]` with only the subject swapped. Conceive each scene individually.
3. **Vary descriptive phrasing.** Limited reuse is acceptable, but no single descriptive pattern should dominate the batch. Reusing the required timing notation is allowed.
4. **DO NOT mix mismatched subjects and settings.** Every (subject, action, setting) triple must be physically and culturally plausible. Read each prompt back to yourself and check: "would this sentence appear in a real video description?"
5. **ABSOLUTE NO-DUPLICATES.** Two prompts are duplicates if a viewer watching both videos would see the same core subject undergoing the same transition in essentially the same setting — regardless of wording. **Same transition pattern + same subject family = duplicate even if wording differs.**
6. **No tool lookup needed.** Use only common, everyday subjects (no rare species, no fictional characters, no obscure landmarks). The challenge is the temporal structure, not entity identification.
7. **Vary template phrasing.** Do not use the same descriptive opener for more than ~15% of prompts; the mandatory timing specification is exempt.

---

## [EXCLUDE] Anti-pattern examples (DO NOT generate prompts like these)

- [REJECT] "Begins with a toy ball is whole and still on a wooden desk, transitions to a toy ball is cracked or dented on a wooden desk, ends with a toy ball is broken apart on a wooden desk."
- [REJECT] "The scene starts as a cook starts walking slowly in a bathroom, becomes a cook breaks into a run in a bathroom, and finishes as a cook exits the frame leaving the space empty in a bathroom."
- [REJECT] "A 5-second video: first half a window sits closed near a puddle, second half a window is open and scattered near a puddle."
- [REJECT] "The sequence begins with a person holds folded laundry on a shelf, shifts into a person hangs the laundry on a line on a shelf..."
- [REJECT] "A dog is missing from in a small kitchen, becomes a dog appears and pauses in the middle of in a small kitchen, finishes as a dog dashes away from in a small kitchen."

[PASS] The right way: imagine an actual short video clip you could film in real life. Write a fresh sentence describing it, with concrete details (time of day, color, exact location, action verb), within the shot count and common duration already chosen. Never start from a template and fill slots.

---

## [CATEGORIES] Subcategories (6) — distribute candidates across the six subcategories

### 1. `state_process_change` — the subject itself physically changes over time
The object or material changes state, form, condition, or visible composition.
Examples include: intact → broken, dry → wet, frozen → melted, full → empty, sealed → opened, a folded cloth being opened, a capped bottle being uncapped, a sponge being squeezed.

Use this category when the **main visual story is the subject itself transforming**.

Examples:
- "In an 8-second video, 0-4s: a softened ice cream scoop tilts on a cone; 4-8s: it slides off onto the porch steps."
- "Create three equal-length shots in a 9-second video: first, a hand presses a wet sponge; next, water streams into a bowl; finally, the released sponge expands."
- "A 6-second video has two equal-length shots: first, a hand twists open an orange soda bottle at a picnic table; second, foam spills over the rim."

### 2. `motion_change` — stillness becomes movement, or movement comes to rest
A subject starts moving, stops moving, tips, falls, rolls, springs up, settles down, or changes from active to still.

Use this category when the **core change is motion status**, not entering the frame and not a large causal chain.

Examples:
- "In a 6-second video, 0-3s: a chess piece teeters at the board edge; 3-6s: it tips off and bounces on the floor."
- "Film three equal-length shots over 9 seconds: initially, a tabby sleeps on a windowsill; next, its ears twitch; finally, it jumps upright toward the glass."
- "For an 8-second video, show two equal-length shots: a skateboard rolls across concrete in the first, then wobbles to a stop against a curb in the second."

### 3. `frame_crossing` — a subject enters or leaves the frame
A space changes from empty to occupied or occupied to empty because the main subject visibly crosses into or out of the shot.

Use this category when the **main temporal event is crossing the frame boundary**.

Examples:
- "Over 8 seconds, 0-4s: an empty hospital corridor remains still; 4-8s: a nurse pushes a gurney into view from the right."
- "Use three equal-length shots totaling 12 seconds: first, a bicycle stands beside a wall; next, a man enters and mounts it; last, he rides out of frame."
- "Across a 9-second video, 0-3s: an empty playground at dusk; 3-6s: two children run into view; 6-9s: they disappear behind the swings."

### 4. `human_task` — a person completes an everyday action with visible stages
Preparation → execution → result. Cooking gestures, dressing, cleaning, writing, packing, tying shoes, pouring drinks, brushing hair, sports motions, simple craft actions.

Use this category when the **main story is a person carrying out a task**, even if objects are involved.

Examples:
- "Make a 12-second video with three equal-length shots: first, a barista aligns a filled portafilter; next, she locks it into the machine; finally, she presses the brew button."
- "Within an 8-second video, 0-4s: a runner tightens already tied shoelaces; 4-8s: he stands and jogs down the sidewalk."
- "Divide 12 seconds into three equal-length shots: a woman folds a blanket, smooths its corners, and places it in a basket, preserving the same sofa setting."

### 5. `environment_shift` — ambient conditions change around a mostly stable scene
Light, weather, atmosphere, fog, steam, shadows, rain, wind, sunrise, cloud cover, or water level changes while the main subject or setting remains largely the same.

Use this category when the **environment changes more than the subject does**.

Examples:
- "During an 8-second video, 0-4s: dry cobblestones lie under gray light; 4-8s: raindrops begin striking and darkening the same alley."
- "Record three equal-length shots in 12 seconds: first, a ridge lies under moving clouds; next, sunlight reaches its crest; finally, light spreads down the slope."
- "An 8-second video contains two equal-length shots: first, laundry hangs still in a backyard; second, a gust lifts the shirts and bends the nearby branches."

### 6. `causal_event` — one action or trigger causes a visible consequence
A push triggers dominoes, a ball knocks over blocks, a child startles pigeons, a phone begins ringing, a balloon pops, a dropped spoon splashes soup, a door slam makes papers flutter.

Use this category when the **main story is a cause-and-effect event**, including sudden disruptions and multi-object interactions.

Example:
- "In a 9-second video, 0-3s: a hand nudges a domino; 3-6s: the row topples toward a bell; 6-9s: the final domino strikes the bell. A fixed camera holds the tabletop."


---

## [KEY] Tie-break rules — use these whenever a prompt could fit more than one class

1. If the main subject **crosses into or out of the frame**, label it `frame_crossing`.
2. If a **person performing a task** is the visual focus, label it `human_task`, even if objects change during the task.
3. If the key change is **the subject transforming physically**, label it `state_process_change`.
4. If the key change is **still <-> moving / moving <-> still**, label it `motion_change`.
5. If the scene stays mostly fixed but **weather, light, or ambient conditions** change, label it `environment_shift`.
6. If the clip is defined by **a trigger producing a consequence**, especially involving multiple objects or a sudden disruption, label it `causal_event`.

When unsure between `state_process_change` and `causal_event`:
- choose `state_process_change` if the emphasis is on the **transformation itself**
- choose `causal_event` if the emphasis is on the **trigger and resulting reaction**

When unsure between `human_task` and `causal_event`:
- choose `human_task` if the person’s staged action is the focus
- choose `causal_event` if the causal reaction is the focus

---

## [FORMAT] Format templates — VARY phrasing across all of these

Two-shot formats:
- State "a T-second video in two equal-length shots", then describe shot 1 and shot 2.
- State T and use adjacent intervals 0-d seconds and d-T seconds, with T = 2d.

Three-shot formats:
- State "a T-second video in three equal-length shots", then describe shots 1, 2, and 3 in order.
- State T and use adjacent intervals 0-d seconds, d-2d seconds, and 2d-T seconds, with T = 3d.

In both formats, d must be positive and no greater than 5 seconds. Always replace T and d with actual numbers. Timing notation may repeat; vary the scenes and descriptive wording instead of compromising precise timing.

No single descriptive opener may exceed ~15% of the batch; required timing notation is exempt.

---

## [REJECT] Self-filtering — REMOVE and replace any prompt if ANY apply

1. Slot-fill template: prompt looks like `[subject] [templated phrase] [in setting]` and merely swaps subjects or settings without changing the underlying event. Reusing required timestamp syntax alone is not a violation.
2. Implausible (subject, action, setting) combination — read it aloud: would a real videographer film this? If not, reject.
3. Broken grammar or visible copy-paste artifacts.
4. Two/three phases not visually distinct enough — a viewer of just one segment must be able to tell which segment they saw.
5. Only camera angle changes between phases — that is not a real transition.
6. Same subject + same transition family already used.
7. Fewer than 12 words or more than 45 words, counting whitespace-separated tokens in the full prompt including timing information.
8. Wrong subcategory.
9. Three-phase prompt where middle segment is trivial or redundant.
10. Descriptive opener over-concentration, excluding mandatory timing notation.
11. Missing total duration, unspecified shot count, unequal shot lengths, any shot longer than 5 seconds, or gaps/overlaps in timestamps.
12. A shot requires implausibly fast changes for its allocated duration, or its content is not specified separately from the other shots.

---

## [PASS] Generation procedure

For each new prompt:
1. FIRST choose K = 2 or K = 3 shots, respecting the approximate balance.
2. THEN choose one common duration d <= 5 seconds and compute T = K * d.
3. Pick an underrepresented subcategory and imagine a plausible everyday event that fits those intervals.
4. Describe each shot's subjects, actions, and visible state separately, preserving continuity where the event requires it.
5. State the total duration and either continuous timestamps or explicit equal-length shot wording with ordered shot descriptions.
6. Apply the subcategory tie-break rules.
7. Check all 12 self-filter rules and every timing calculation.
8. Check against available earlier outputs and the anti-pattern examples; remove repeated underlying events.

Do this for each of the 100 candidates in the batch. No shortcuts. No slot-filling. No loops.

---

## Output

Read `<PROMPT_DATA_DIR>/timegrounded.json` if it exists; otherwise begin with an empty exclusion set.

Use all existing entries in that file as an exclusion set for duplicate checking. A newly generated prompt is invalid if it duplicates or near-duplicates any existing prompt in that file by subject configuration, interaction pattern, and setting, even if the wording differs.

After generating this batch of 100 candidates and applying self-filtering, return the valid new objects as a JSON array and append only those objects to `<PROMPT_DATA_DIR>/timegrounded.json`. Do not overwrite, rewrite, or regenerate the existing entries.
\end{datapromptbox}

\end{document}